\documentclass{article}

\usepackage[final]{neurips_2021}

\usepackage{amsmath,amsfonts,bm}

\newcommand{\f}{\mathbf{f}}
\newcommand{\x}{\bm{x}}

\newcommand{\bu}{\mathbf{u}}
\newcommand{\bv}{\mathbf{v}}
\newcommand{\K}{\mathbf{K}}
\newcommand{\X}{\mathbf{X}}
\newcommand{\Y}{\mathbf{Y}}
\newcommand{\Z}{\mathbf{Z}}
\newcommand{\F}{\mathbf{F}}
\newcommand{\BH}{\mathbf{H}}

\def\eqref#1{equation~\ref{#1}}

\def\1{\bm{1}}

\def\rva{{\mathbf{a}}}

\def\rmA{{\mathbf{A}}}

\def\rmE{{\mathbf{E}}}

\def\rmU{{\mathbf{U}}}

\def\rmW{{\mathbf{W}}}

\def\vmu{{\bm{\mu}}}
\def\vtheta{{\bm{\theta}}}

\def\mA{{\bm{A}}}

\def\mD{{\bm{D}}}

\def\mI{{\bm{I}}}

\def\mL{{\bm{L}}}

\def\mZ{{\bm{Z}}}

\DeclareMathAlphabet{\mathsfit}{\encodingdefault}{\sfdefault}{m}{sl}
\SetMathAlphabet{\mathsfit}{bold}{\encodingdefault}{\sfdefault}{bx}{n}

\usepackage[utf8]{inputenc} 
\usepackage[T1]{fontenc}    
\usepackage{hyperref}       
\hypersetup{colorlinks,linkcolor={MidnightBlue},citecolor={MidnightBlue},urlcolor={BurntOrange}}
\usepackage{url}            
\usepackage{booktabs}       
\usepackage{amsfonts}       
\usepackage{nicefrac}       
\usepackage{microtype}      
\usepackage[dvipsnames]{xcolor}         
\usepackage{graphicx,subfigure}
\usepackage{caption}
\usepackage{wrapfig}
\usepackage{enumitem}
\usepackage{cancel}
\usepackage[nameinlink,capitalize]{cleveref}
\crefname{equation}{Eq.}{eqs.}
\usepackage{tabularx}

\usepackage[frozencache=true,cachedir=.]{minted}

\usepackage[acronym,smallcaps,nowarn]{glossaries}
\newacronym{bnn}{BNN}{Bayesian Neural Network}

\newacronym{ece}{ECE}{expected calibration error}

\newacronym{fcg}{FCG}{Full-covariance Gaussian}
\newacronym{ffg}{FFG}{Fully-factorized Gaussian}

\newacronym{gp}{GP}{Gaussian process}

\newacronym{nll}{NLL}{negative log likelihood}

\newacronym{ood}{OOD}{out-of-distribution}

\newacronym{svgp}{SVGP}{variationally sparse \acrshort{gp}}

\newacronym{vi}{VI}{variational inference}
\makeatletter
\newcommand*{\glsplainhyperlink}[2]{%
  \colorlet{currenttext}{.}
  \colorlet{currentlink}{\@linkcolor}
  \hypersetup{linkcolor=currenttext}
  \hyperlink{#1}{#2}%
  \hypersetup{linkcolor=currentlink}
}
\let\@glslink\glsplainhyperlink
\makeatother

\title{Sparse Uncertainty Representation in Deep Learning with Inducing Weights}

\author{
  Hippolyt Ritter$^1$\thanks{Work done at Microsoft Research Cambridge. },
  Martin Kukla$^2$,
  Cheng Zhang$^2$ \&
  Yingzhen Li$^{3*}$ \\
  $^1$University College London $^2$Microsoft Research Cambridge, UK   $^3$Imperial College London\\
  \texttt{j.ritter@cs.ucl.ac.uk,  \{Martin.Kukla,Cheng.Zhang\}@microsoft.com,}\\ \texttt{yingzhen.li@imperial.ac.uk}
}

\newcommand{\params}{\vtheta}
\newcommand{\auxvars}{\rva}
\newcommand{\mEps}{\rmE}
\newcommand{\weight}{\rmW}
\newcommand{\allweights}{\weight_{1:L}}
\newcommand{\inducing}{\rmU}
\newcommand{\allinducing}{\inducing_{1:L}}

\newcommand{\given}{\vert}
\newcommand{\data}{\mathcal{D}}
\newcommand{\varparams}{\bm{\phi}}
\newcommand{\eye}{\mI}
\newcommand{\pmat}[1]{\begin{pmatrix}#1\end{pmatrix}}
\newcommand{\covparam}{\mZ}
\newcommand{\precparam}{\mD}
\newcommand{\fullcov}{\bm{\Sigma}}
\newcommand{\inducingcov}{\bm{\Psi}}
\newcommand{\inducingprec}{\inducingcov^{-1}}
\newcommand{\inducingchol}{\mL}
\newcommand{\whitenedchol}{\tilde{\inducingchol}}
\newcommand{\whitenedfullcov}{\tilde{\fullcov}}
\newcommand{\kron}{\otimes}

\DeclareMathOperator{\gauss}{\mathcal{N}}
\DeclareMathOperator{\matgauss}{\mathcal{MN}}
\DeclareMathOperator{\avg}{\mathbb{E}}
\DeclareMathOperator{\kl}{\mathbb{KL}}
\DeclareMathOperator{\vect}{vec}
\DeclareMathOperator{\trace}{tr}

\usepackage{xcolor}

\usepackage{catchfile}

\begin{document}

\maketitle

\begin{abstract}
\acrlongpl{bnn} and deep ensembles represent two modern paradigms of uncertainty quantification in deep learning.
Yet these approaches struggle to scale mainly due to memory inefficiency, requiring parameter storage several times that of their deterministic counterparts.
To address this, we augment each weight matrix with a small inducing weight matrix, projecting the uncertainty quantification into a lower dimensional space.
We further extend Matheron's conditional Gaussian sampling rule to enable fast weight sampling, which enables our inference method to maintain reasonable run-time as compared with ensembles.
Importantly, our approach achieves competitive performance to the state-of-the-art in prediction and uncertainty estimation tasks with fully connected neural networks and ResNets, while reducing the parameter size to $\leq 24.3\%$ of that of a \emph{single} neural network.
\end{abstract}

\section{Introduction}

Deep learning models are becoming deeper and wider than ever before. From image recognition models such as ResNet-101 \citep{he2016deep} and DenseNet \citep{huang2017densely} to BERT \citep{xu-etal-2019-bert} and GPT-3 \citep{brown2020language} for language modelling, deep neural networks have found consistent success in fitting large-scale data. As these models are increasingly deployed in real-world applications, calibrated uncertainty estimates for their predictions become crucial, especially in safety-critical areas such as healthcare. In this regard, \acrfullpl{bnn} \citep{mackay1995bayesian,blundell2015weight,gal2016dropout,zhang2019cyclical} and deep ensembles \citep{lakshminarayanan2017simple} represent two popular paradigms for estimating uncertainty, which have shown promising results in applications such as (medical) image processing \citep{kendall2017uncertainties,tanno2017bayesian} and out-of-distribution detection \citep{ovadia2019can}.

Though progress has been made, one major obstacle to scaling up \acrshortpl{bnn} and deep ensembles is their high storage cost. Both approaches require the parameter counts to be several times higher than their deterministic counterparts. Although recent efforts have improved memory efficiency \citep{louizos2017multiplicative,swiatkowski2020k,wen2019batchensemble,dusenberry2020efficient}, these still use more parameters than a deterministic neural network. This is particularly problematic in hardware-constrained edge devices, when on-device storage is required due to privacy regulations.

Meanwhile, an \emph{infinitely wide} \acrshort{bnn} becomes a \acrfull{gp} that is known for good uncertainty estimates \citep{neal1995bayesian,matthews2018gaussian,lee2018deep}. But perhaps surprisingly, this infinitely wide \acrshort{bnn} is ``parameter efficient'', as its ``parameters'' are effectively the datapoints, which have a considerably smaller memory footprint than explicitly storing the network weights. In addition, sparse posterior approximations store a smaller number of \emph{inducing points} instead \citep{snelson2006sparse,titsias2009variational}, making sparse \acrshortpl{gp} even more memory efficient. 


Can we bring the advantages of sparse approximations in \acrshortpl{gp} --- which are infinitely-wide neural networks --- to finite width deep learning models?
We provide an affirmative answer regarding memory efficiency, by proposing an uncertainty quantification framework based on \emph{sparse uncertainty representations}. We present our approach in \acrshort{bnn} context, but the proposed approach is also applicable to deep ensembles. In detail, our contributions are as follows:

\begin{itemize}
\setlength\itemsep{0em}
    \item We introduce  \emph{inducing weights} --- an auxiliary variable method with lower dimensional counterparts to the actual weight matrices --- for variational inference in \acrshortpl{bnn}, as well as a memory efficient parameterisation and an extension to ensemble methods (\cref{sec:inducing}).
    \item We extend Matheron's rule to facilitate efficient posterior sampling (\cref{sec:conditional_sampling_matheron}).
    \item We provide an in-depth computation complexity analysis (\cref{sec:computation}), showing the significant advantage in terms of parameter efficiency.
    \item We show the connection to sparse (deep) \acrshortpl{gp}, in that inducing weights can be viewed as \emph{projected noisy inducing outputs} in pre-activation output space (\cref{sec:function_space_view}).
    \item We apply the proposed approach to \acrshortpl{bnn} and deep ensembles. Experiments in classification, model robustness and out-of-distribution detection tasks show that our inducing weight approaches achieve competitive performance to their counterparts in the original weight space on modern deep architectures for image classification, while reducing the parameter count to $\leq 24.3\%$ of that of a single network.
\end{itemize}


We open-source our proposed inducing weight approach, together with baseline methods reported in the experiments, as a PyTorch \citep{paszke2017automatic} wrapper named \texttt{bayesianize}:  \url{https://github.com/microsoft/bayesianize}. As demonstrated in \cref{app:open_source_code}, our software makes the conversion of a deterministic neural network to a Bayesian one with a few lines of code:
\vspace{-0.5em}
\begin{minted}{python}
import bnn # our pytorch wrapper package
net = torchvision.models.resnet18() # construct a deterministic ResNet18
bnn.bayesianize_(net, inference="inducing") # convert it into a Bayesian one
\end{minted}

\section{Inducing variables for variational inference}
\label{sec:background}

Our work is built on variational inference and inducing variables for posterior approximations.
Given observations $\data = \{\X, \Y \}$ with $\X = [\x_1, ..., \x_N]$, $\Y = [\bm{y}_1, ..., \bm{y}_N]$, we would like to fit a neural network $p(\bm{y} | \bm{x}, \allweights)$ with weights $\allweights$ to the data. \acrshortpl{bnn} posit a prior distribution $p(\allweights)$ over the weights, and construct an approximate posterior $q(\allweights)$ to the exact posterior $p(\allweights \given \data) \propto p(\data \given \allweights) p(\allweights)$, where $p(\data | \allweights) = p(\Y | \X, \allweights) = \prod_{n=1}^N p(\bm{y}_n | \bm{x}_n, \allweights)$.
%

\paragraph{Variational inference}
Variational inference \citep{hinton1993keeping,jordan:vi1999,zhang2018advances} constructs an approximation $q(\params)$ to the posterior $p(\params | \data) \propto p(\params)p(\data|\params)$ by maximising a variational lower-bound:
\begin{equation}
    \log p(\data)
    \geq \mathcal{L}(q(\params)) := \avg_{q(\params)}\left[ \log p(\data\given\params) \right] - \kl\left[ q(\params) \vert\vert p(\params) \right].
\label{eq:variational_lowerbound}
\end{equation}
For \acrshortpl{bnn}, $\params = \{ \allweights \}$, and a simple choice of $q$ is a \acrfull{ffg}: $q(\allweights) = \prod_{l=1}^L \prod_{i=1}^{d_{out}^l}\prod_{j=1}^{d_{in}^l} \gauss(m_l^{(i,j)}, v_l^{(i,j)})$, with $m_l^{(i,j)}, v_l^{(i,j)}$ the mean and variance of $\weight_l^{(i,j)}$ and  $d_{in}^l, d_{out}^l$ the respective number of inputs and outputs to layer $l$.
The variational parameters are then $\varparams = \{ m_l^{(i,j)}, v_l^{(i,j)} \}_{l=1}^L$.
Gradients of $\mathcal{L}$ w.r.t. $\varparams$ can be estimated with mini-batches of data \citep{hoffman2013stochastic} and with Monte Carlo sampling from the $q$ distribution \citep{titsias2014doubly, kingma2014auto}. By setting $q$ to an \acrshort{bnn}, a variational \acrshort{bnn} can be trained with similar computational requirements as a deterministic network \citep{blundell2015weight}.

\paragraph{Improved posterior approximation with inducing variables}
%
%
Auxiliary variable approaches \citep{agakov2004auxiliary,salimans:mcmcvi2015,ranganath:hvm2016}  construct the $q(\params)$ distribution with an auxiliary variable $\auxvars$: $q(\params) = \int q(\params | \auxvars)q(\auxvars) d\auxvars$, with the hope that a potentially richer mixture distribution $q(\params)$ can achieve better approximations. As then $q(\params)$ becomes intractable, an auxiliary variational lower-bound is used to optimise $q(\params, \auxvars)$ (see \cref{app:auxiliary_variational_objective_derivations}):
\begin{equation}
    \log{} p(\data) \geq \mathcal{L}(q(\params, \auxvars)) = \mathbb{E}_{q(\params, \auxvars)}[\log p(\data | \params)] + \mathbb{E}_{q(\params, \auxvars)} \left[ \log \frac{p(\params) r(\auxvars | \params)}{q(\params | \auxvars)q(\auxvars)} \right].
\label{eq:auxiliary_variational_lowerbound}
\end{equation}

Here $r(\auxvars|\params)$ is an auxiliary distribution that needs to be specified, where existing approaches often use a ``reverse model'' for $r(\auxvars|\params)$.
Instead, we define $r(\auxvars | \params)$ in a generative manner: $r(\auxvars|\params)$ is the ``posterior'' of the following ``generative model'', whose ``evidence'' is exactly the prior of $\params$:
\begin{equation}
    r(\auxvars|\params) = \tilde{p}(\auxvars|\params) \propto \tilde{p}(\auxvars)\tilde{p}(\params| \auxvars), \quad \text{such that } \tilde{p}(\params) := \int \tilde{p}(\auxvars)\tilde{p}(\params| \auxvars) d\auxvars = p(\params).\\
\label{eq:marginalisation_constraint}
\end{equation}
Plugging \cref{eq:marginalisation_constraint} into \cref{eq:auxiliary_variational_lowerbound}:
\begin{equation}
    \mathcal{L}(q(\params, \auxvars)) = \mathbb{E}_{q(\params)}[\log p(\data | \params)]
    - \mathbb{E}_{q(\auxvars)} \left[ \kl[q(\params | \auxvars) || \tilde{p}(\params | \auxvars)] \right] - \kl[q(\auxvars) || \tilde{p}(\auxvars)].
\label{eq:inducing_elbo_general}
\end{equation}
This approach yields an efficient approximate inference algorithm, translating the complexity of inference in $\params$ to $\auxvars$.
If $\text{dim}(\auxvars) < \text{dim}(\params) $ and $q(\params, \auxvars) =q(\params | \auxvars) q(\auxvars) $ has the following properties: 

\begin{itemize}[leftmargin=1cm,topsep=0em]
\setlength\itemsep{0em}
    \item[1.] A ``pseudo prior'' $\tilde{p}(\auxvars) \tilde{p}(\params | \auxvars)$ is defined such that $\int \tilde{p}(\auxvars)\tilde{p}(\params| \auxvars) d\auxvars = p(\params)$;
    \item[2.] The conditionals $q(\params | \auxvars)$ and $\tilde{p}(\params| \auxvars)$ are in the same parametric family, so can share parameters;
    \item[3.] Both sampling $\params \sim q(\params)$ and computing $\kl[q(\params | \auxvars) || \tilde{p}(\params | \auxvars)]$ can be done efficiently;
    \item[4.] The designs of $q(\auxvars)$ and $\tilde{p}(\auxvars)$ can potentially provide extra advantages (in time and space complexities and/or optimisation easiness).
\end{itemize}
We call $\auxvars$ the \emph{inducing variable} of $\params$, which is inspired by \acrfull{svgp} with inducing points \citep{snelson2006sparse,titsias2009variational}.
Indeed \acrshort{svgp} is a special case (see \cref{app:gp_intro}): $\params = \f$, $\auxvars = \bu$, the \acrshort{gp} prior is $p(\f | \X) = \mathcal{GP}(\bm{0},  \K_{\X\X})$, $p(\bu) = \mathcal{GP}(\bm{0}, \K_{\Z\Z})$, $\tilde{p}(\f, \bu) = p(\bu)p(\f|\X, \bu)$, $q(\f | \bu) = p(\f | \X, \bu)$, $ q(\f, \bu) = p(\f |\X, \bu) q(\bu)$, and $\Z$ are the optimisable inducing inputs.
The variational lower-bound is
$
    \mathcal{L}(q(\f, \bu)) = \mathbb{E}_{q(\f)}[\log p(\Y | \f)] - \kl[q(\bu) || p(\bu)],
$
and the variational parameters are $\varparams = \{ \Z, \text{distribution parameters of } q(\bu) \}$. 
\acrshort{svgp} satisfies the marginalisation constraint \cref{eq:marginalisation_constraint} by definition, and it has $\kl[q(\f | \bu) || \tilde{p}(\f |\bu)] = 0$. Also by using small $M=\text{dim}(\bu)$ and exploiting the $q$ distribution design, \acrshort{svgp} reduces run-time from $\mathcal{O}(N^3)$ to $\mathcal{O}(NM^2 + M^3)$ where $N$ is the number of inputs in $\X$, meanwhile it also makes storing a full Gaussian $q(\bu)$ affordable. Lastly, $\bu$ can be whitened, leading to the ``pseudo prior'' $\tilde{p}(\f, \bv) = p(\f |\X, \bu = \K_{\Z\Z}^{1/2}\bv) \tilde{p}(\bv), \tilde{p}(\bv) = \mathcal{N}(\bv; \bm{0}, \mathbf{I})$ which could bring potential benefits in optimisation.

We emphasise that the introduction of ``pseudo prior'' does \emph{not} change the \emph{probabilistic model} as long as the marginalisation constraint \cref{eq:marginalisation_constraint} is satisfied. In the rest of the paper we assume the constraint \cref{eq:marginalisation_constraint} holds and write $p(\params, \auxvars) := \tilde{p}(\params, \auxvars)$. 
It might seem unclear how to design such $\tilde{p}(\params, \auxvars)$ for an arbitrary probabilistic model, however, for a Gaussian prior on $\params$ the rules for computing conditional Gaussian distributions can be used to construct $\tilde{p}$. In \cref{sec:method} we exploit these rules to develop an efficient approximate inference method for Bayesian neural networks with inducing weights.

\section{Sparse uncertainty representation with inducing weights}
\label{sec:method}

\subsection{Inducing weights for neural network parameters}
\label{sec:inducing}

Following the above design principles, we introduce to each network layer $l$ a \emph{smaller} inducing weight matrix $\inducing_l$ to assist approximate posterior inference in $\weight_l$. Therefore in our context, $\params = \allweights$ and $\auxvars = \inducing_{1:L}$. In the rest of the paper, we assume a factorised prior across layers $p(\allweights) = \prod_{l}p(\weight_l)$, and drop the $l$ indices when the context is clear to ease notation.

\paragraph{Augmenting network layers with inducing weights}
Suppose the weight $\weight \in \mathbb{R}^{d_{out} \times d_{in}}$ has a Gaussian prior $p(\weight) = p(\vect(\weight)) = \gauss(0, \sigma^2 \eye)$ where $\vect(\weight)$ concatenates the columns of the weight matrix into a vector.
A first attempt to augment $p(\vect(\weight))$ with an inducing weight variable $\inducing \in \mathbb{R}^{M_{out} \times M_{in}}$ may be to construct a multivariate Gaussian $p(\vect(\weight), \vect(\inducing))$, such that $\int p(\vect(\weight), \vect(\inducing)) d\inducing = \gauss(0, \sigma^2 \eye)$. This means for the joint covariance matrix of $(\vect(\weight), \vect(\inducing))$, it requires the block corresponding to the covariance of $\vect(\weight)$ to match the prior covariance $\sigma^2 \eye$. We are then free to parameterise the rest of the entries in the joint covariance matrix, as long as this full matrix remains positive definite. Now the conditional distribution $p(\weight\given\inducing)$ is a function of these parameters, and the conditional sampling from $p(\weight\given\inducing)$ is further discussed in \cref{app:inducing_weight_vector_normal}.
Unfortunately, as $\text{dim}(\vect(\weight))$ is typically large (e.g.~of the order of $10^7$), using a full covariance Gaussian for $p(\vect(\weight), \vect(\inducing))$ becomes computationally intractable. 

We address this issue with matrix normal distributions \citep{gupta2018matrix}. The prior $p(\vect(\weight)) = \gauss(\bm{0}, \sigma^2 \eye)$ has an equivalent matrix normal distribution form as $p(\weight) = \matgauss(0, \sigma_r^2 \eye, \sigma_c^2 \eye)$, with $\sigma_r, \sigma_c > 0$ the row and column standard deviations satisfying $\sigma = \sigma_r\sigma_c$.
Now we introduce the inducing variable $\inducing$ in matrix space, as well as two auxiliary variables $\inducing_r \in \mathbb{R}^{M_{out} \times d_{in}}$, $\inducing_c \in \mathbb{R}^{d_{out} \times M_{in}}$, so that the full augmented prior is:
%
\begin{align}
    \pmat{\weight & \inducing_c\\\inducing_r & \inducing} &\sim p(\weight, \inducing_c, \inducing_r, \inducing) := \matgauss(0, \fullcov_r, \fullcov_c), \label{eq:inducing_full_joint}
\end{align}
\begin{align*}
    \text{with}\quad
    &\inducingchol_r = \pmat{\sigma_r \eye & 0\\\covparam_r & \precparam_r}
    \quad\text{s.t.}\quad
    \fullcov_r = \inducingchol_r\inducingchol_r^\top = \pmat{\sigma_r^2\eye & \sigma_r \covparam_r^\top\\\sigma_r \covparam_r & \covparam_r\covparam_r^\top + \precparam_r^2}
    \\
    \;\text{and}\quad
    &\inducingchol_c = \pmat{\sigma_c \eye & 0\\\covparam_c & \precparam_c}
    \quad\text{s.t.}\quad
    \fullcov_c = \inducingchol_c\inducingchol_c^\top = \pmat{\sigma_c^2\eye & \sigma_c\covparam_c^\top\\\sigma_c\covparam_c & \covparam_c\covparam_c^\top + \precparam_c^2}.
\end{align*}
See \cref{fig:visualisation_matheron}(a) for a visualisation of the augmentation. Matrix normal distributions have similar marginalisation and conditioning rules as multivariate Gaussian distributions, for which we provide further examples in \cref{app:inducing_weight_matrix_normal}. Therefore the marginalisation constraint \cref{eq:marginalisation_constraint} is satisfied for any $\covparam_c \in \mathbb{R}^{M_{in} \times d_{in}}, \covparam_r \in \mathbb{R}^{M_{out} \times d_{out}}$ and diagonal matrices $\precparam_c, \precparam_r$.  
For the inducing weight $\inducing$ we have $p(\inducing) = \matgauss(0, \inducingcov_r, \inducingcov_c)$ with $\inducingcov_r = \covparam_r\covparam_r^\top + \precparam_r^2$ and $\inducingcov_c = \covparam_c\covparam_c^\top + \precparam_c^2$. In the experiments we use \emph{whitened} inducing weights which transforms $\inducing$ so that $p(\inducing) = \matgauss(0, \eye, \eye)$ (\cref{app:hierarchical_inducing}), but for clarity we continue with the above formulas in the main text.

The matrix normal parameterisation introduces two additional variables $\inducing_r, \inducing_c$ without providing additional expressiveness.
Hence it is desirable to integrate them out, leading to a joint multivariate normal with Khatri-Rao product structure for the covariance:
\begin{equation}
    p(\vect(\weight), \vect(\inducing)) = \gauss \left(0, \pmat{\sigma_c^2 \eye \kron \sigma_r^2 \eye & \sigma_c\covparam_c^\top \kron \sigma_r\covparam_r^\top\\\sigma_c\covparam_c \kron \sigma_r\covparam_r & \inducingcov_c \kron \inducingcov_r} \right).
    \label{eq:inducing_joint_khatri_rao}
\end{equation}
As the dominating memory complexity here is $\mathcal{O}(d_{out}M_{out} + d_{in}M_{in})$ which comes from storing $\covparam_r$ and $\covparam_c$, we see that the matrix normal parameterisation of the augmented prior is memory efficient.

\paragraph{Posterior approximation in the joint space}
We construct a factorised posterior approximation across the layers: $q(\allweights, \allinducing) = \prod_l q(\weight_l | \inducing_l) q(\inducing_l)$. Below we discuss options for $q(\weight | \inducing)$. 

The simplest option is $q(\weight\given\inducing) = p(\vect(\weight)\given\vect(\inducing)) = \gauss(\vmu_{\weight\given\inducing}, \fullcov_{\weight\given\inducing})$, similar to sparse GPs.
A slightly more flexible variant adds a rescaling term $\lambda^2$ to the covariance matrix, which allows efficient KL computation (\cref{app:kl_divergence_derivation}):
\begin{equation}
    q(\weight\given\inducing) = q(\vect(\weight)\given\vect(\inducing)) = \gauss(\vmu_{\weight\given\inducing}, \lambda^2 \fullcov_{\weight\given\inducing}),
\label{eq:conditional_q_distribution_main}
\end{equation}
\begin{equation}
    R(\lambda) := \kl\left[ q(\weight\given\inducing) \vert\vert p(\weight\given\inducing ) \right]
    = d_{in} d_{out} (0.5\lambda^2 - \log \lambda - 0.5).
\label{eq:conditional_kl}
\end{equation}

Plugging $\params = \allweights, a = \allinducing$ and \cref{eq:conditional_kl} into
\cref{eq:inducing_elbo_general} returns the following variational lower-bound
\begin{equation}
    \mathcal{L}(q(\allweights, \inducing_{1:L})) = \mathbb{E}_{q(\allweights)}[\log p(\data | \allweights)]\,-
    \sum\nolimits_{l=1}^L (R(\lambda_l) + \kl[q(\inducing_l) || p(\inducing_l)]),
\label{eq:elbo_inducing_weight_reduced}
\end{equation}
with $\lambda_l$ the associated scaling parameter for $q(\weight_l | \inducing_l)$. Again as the choices of $\covparam_c, \covparam_r, \precparam_c, \precparam_r$ do not change the marginal prior $p(\weight)$, we are safe to optimise them as well. Therefore the variational parameters are now $\phi = \{\covparam_c, \covparam_r, \precparam_c, \precparam_r, \lambda, \text{dist.~params.~of } q(\inducing) \}$ for each layer.

\paragraph{Two choices of $q(\inducing)$}
A simple choice is \acrshort{ffg} $q(\vect(\inducing)) = \gauss(\bm{m}_u, \text{diag}(\bm{v}_u))$, which performs mean-field inference in $\inducing$ space \citep[c.f.][]{blundell2015weight}, and here $\kl[q(\inducing) || p(\inducing)]$ has a closed-form solution.
Another choice is a ``mixture of delta measures'' $q(\inducing) = \frac{1}{K}\sum_{k=1}^K \delta(\inducing = \inducing^{(k)})$, i.e.~we keep $K$ distinct sets of parameters $\{U_{1:L}^{(k)}\}_{k=1}^K$ in inducing space that are projected back into the original parameter space via the \emph{shared} conditionals $q(\weight_l\given\inducing_l)$ to obtain the weights. This approach can be viewed as constructing ``deep ensembles'' in $\inducing$ space, and we follow ensemble methods \citep[e.g.][]{lakshminarayanan2017simple} to drop $\kl[q(\inducing) || p(\inducing)]$ in \cref{eq:elbo_inducing_weight_reduced}. %

Often $\inducing$ is chosen to have significantly lower dimensions than $\weight$, i.e.~$M_{in} << d_{in}$ and $M_{out} << d_{out}$. As $q(\weight|\inducing)$ and $p(\weight | \inducing)$ only differ in the covariance scaling constant, $\inducing$ can be regarded as a \emph{sparse representation of uncertainty} for the network layer, as the major updates in (approximate) posterior belief is quantified by $q(\inducing)$.


\subsection{Efficient sampling with extended Matheron's rule}
\label{sec:conditional_sampling_matheron}

\begin{figure}[t]
    \centering
    \includegraphics[width=0.95\linewidth]{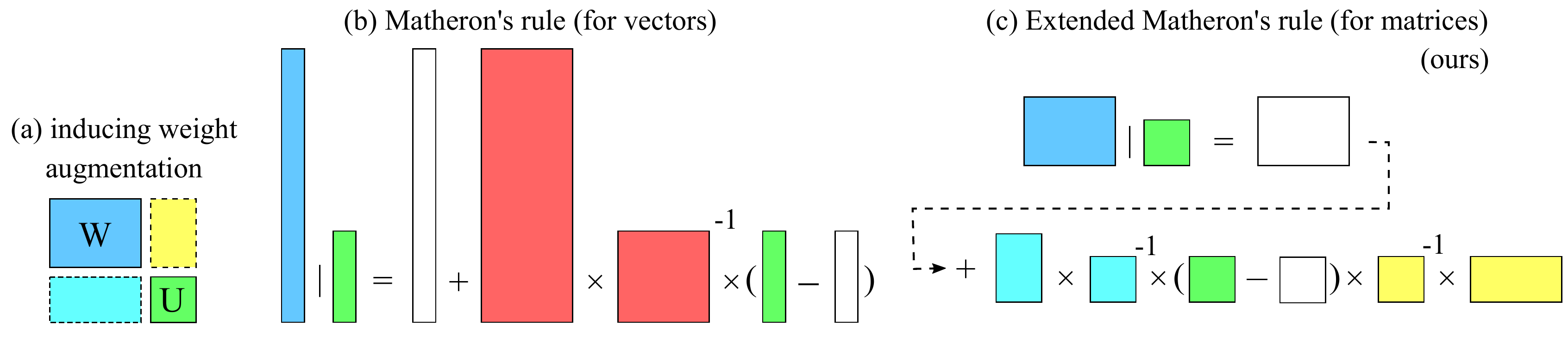}
    \caption{Visualisation of (a) the inducing weight augmentation, and compare (b) the original Matheron's rule to (c) our extended version. White blocks represent samples from the joint Gaussian.}
    \label{fig:visualisation_matheron}
    \vspace{-1em}
\end{figure}

Computing the variational lower-bound \cref{eq:elbo_inducing_weight_reduced} requires samples from $q(\weight)$, which requires an efficient sampling procedure for $q(\weight | \inducing)$. Unfortunately, $q(\weight | \inducing)$ derived from \cref{eq:inducing_joint_khatri_rao} \& \cref{eq:conditional_q_distribution_main} is not a matrix normal, so \emph{direct} sampling is prohibitive.
To address this challenge, we extend Matheron's rule \citep{journel1978mining,hoffman1991constrained,doucet2010gaussian} to efficiently sample from $q(\weight | \inducing)$, with derivations provided in \cref{app:extended_matheron}. 

The original Matheron's rule applies to multivariate Gaussian distributions. As a running example, consider two vector-valued random variables $\bm{w}$, $\bm{u}$ with joint distribution $p(\bm{w}, \bm{u}) = \mathcal{N}(\bm{0}, \fullcov)$. Then the conditional distribution $p(\bm{w} | \bm{u}) = \mathcal{N}(\bm{\mu}_{\bm{w} | \bm{u}}, \fullcov_{\bm{w} | \bm{u}})$ is also Gaussian, and direct sampling from it requires decomposing the conditional covariance matrix $\fullcov_{\bm{w} | \bm{u}}$ which can be costly. The main idea of Matheron's rule is that we can transform a sample from the joint Gaussian to obtain a sample from the conditional distribution $p(\bm{w} | \bm{u})$ as follows:
\begin{align}
    \bm{w} = \bm{\bar{w}} + \fullcov_{\bm{w} \bm{u}} \fullcov_{\bm{u}\bm{u}}^{-1} (\bm{u} - \bm{\bar{u}}), \quad
    \bm{\bar{w}}, \bm{\bar{u}} \sim \gauss(\bm{0}, \fullcov), \quad \fullcov = \pmat{\fullcov_{\bm{w} \bm{w}} & \fullcov_{\bm{w} \bm{u}} \\ \fullcov_{\bm{u} \bm{w}} & \fullcov_{\bm{u} \bm{u}}}.
\end{align}
One can check the validity of Matheron's rule by computing the mean and variance of $\bm{w}$ above: 
\begin{equation*}
\mathbb{E}_{\bm{\bar{w}}, \bm{\bar{u}}}[\bm{w}] = \fullcov_{\bm{w} \bm{u}}  \fullcov_{\bm{u}\bm{u}}^{-1} \bm{u} = \bm{\mu}_{\bm{w} | \bm{u}}, 
\quad \mathbb{V}_{\bm{\bar{w}}, \bm{\bar{u}}}[\bm{w}] = \fullcov_{\bm{w} \bm{w}} - \fullcov_{\bm{w} \bm{u}} \fullcov_{\bm{u}\bm{u}}^{-1} \fullcov_{\bm{u} \bm{w}} = \fullcov_{\bm{w} | \bm{u}}.
\end{equation*}
It might seem counter-intuitive at first sight in that this rule requires samples from a higher dimensional space. However, in the case where decomposition/inversion of $\fullcov$ and $\fullcov_{\bm{u} \bm{u}}$ can be done efficiently, sampling from the joint Gaussian $p(\bm{w}, \bm{u})$ can be significantly cheaper than directly sampling from the conditional Gaussian $p(\bm{w} | \bm{u})$. This happens e.g.~when $\fullcov$ is directly parameterised by its Cholesky decomposition and $\dim(\bm{u}) << \dim(\bm{w})$, so that sampling $\bm{\bar{w}}, \bm{\bar{u}} \sim \gauss(\bm{0}, \fullcov)$ is straight-forward, and computing $\fullcov_{\bm{u} \bm{u}}^{-1}$ is significantly cheaper than decomposing $\fullcov_{\bm{w} | \bm{u}}$.

Unfortunately, the original Matheron's rule cannot be applied directly to sample from $q(\weight | \inducing)$. This is because $q(\weight | \inducing) = q(\vect(\weight) | \vect(\inducing))$ differs from $p(\vect(\weight) | \vect(\inducing))$ only in the variance scaling $\lambda$, and for $p(\vect(\weight) | \vect(\inducing))$, its joint distribution counter-part \cref{eq:inducing_joint_khatri_rao} does not have an efficient representation for the covariance matrix. Therefore a naive application of Matheron's rule requires decomposing the covariance matrix of $p(\vect(\weight), \vect(\inducing))$ which is even more expensive than direct conditional sampling. However, notice that for the joint distribution $p(\weight, \inducing_c, \inducing_r, \inducing)$ in an even higher dimensional space, the row and column covariance matrices $\fullcov_r$ and $\fullcov_c$ are parameterised by their Cholesky decompositions, so that sampling from this joint distribution can be done efficiently. This inspire us to extend the original Matheron's rule for efficient sampling from $q(\weight | \inducing)$ (details in \cref{app:extended_matheron}, when $\lambda = 1$ it also applies to sampling from $p(\weight | \inducing)$):
\begin{equation}
     \weight = \lambda \bar{\weight} + \sigma \covparam_r^\top\inducingprec_r (U - \lambda \bar{\inducing}) \inducingprec_c\covparam_c;\quad
     \bar{\weight}, \bar{\inducing} \sim p(\bar{\weight}, \bar{\inducing_c}, \bar{\inducing_r}, \bar{\inducing}) = \matgauss(0, \fullcov_r, \fullcov_c).
\end{equation}
Here $\bar{\weight}, \bar{\inducing} \sim p(\bar{\weight}, \bar{\inducing_c}, \bar{\inducing_r}, \bar{\inducing})$ means we first sample $\bar{\weight}, \bar{\inducing_c}, \bar{\inducing_r}, \bar{\inducing}$ from the joint then drop $\bar{\inducing_c}, \bar{\inducing_r}$; in fact $\bar{\inducing_c}, \bar{\inducing_r}$ are never computed, and the other samples $\bar{\weight}, \bar{\inducing}$ can be obtained by:
\begin{align} \label{eq:matheron_rule_q_conditional}
    \bar{\weight} &= \sigma \mEps_1, \
    \bar{\inducing} = \covparam_r \mEps_1\covparam_c^\top + \hat{\inducingchol}_r\tilde{\mEps}_2\precparam_c + \precparam_r\tilde{\mEps}_3 \hat{\inducingchol}_c^\top + \precparam_r \mEps_4\precparam_c,\nonumber\\
    \mEps_1 &\sim \matgauss(0, \eye_{d_{out}}, \eye_{d_{in}});\quad
    \tilde{\mEps}_2, \tilde{\mEps}_3, \mEps_4 \sim \matgauss(0, \eye_{M_{out}}, \eye_{M_{in}}),\\
    \hat{\inducingchol}_r &= \text{chol}(\covparam_r\covparam_r^\top), \ \hat{\inducingchol}_c = \text{chol}(\covparam_c\covparam_c^\top).\nonumber
\end{align}
The run-time cost is $\mathcal{O}(2M_{out}^3 + 2M_{in}^3 + d_{out}M_{out}M_{in} + M_{in}d_{out}d_{in})$ required by inverting $\inducingcov_r, \inducingcov_c$, computing $\hat{\inducingchol}_r$, $\hat{\inducingchol}_c$, and the matrix products. The extended Matheron's rule is visualised in \cref{fig:visualisation_matheron} with a comparison to the original Matheron's rule for sampling from $q(\vect(\weight) | \vect(\inducing))$. Note that the original rule requires joint sampling from \cref{eq:inducing_joint_khatri_rao} (i.e.~sampling the white blocks in \cref{fig:visualisation_matheron}(b)) which has $\mathcal{O}((d_{out}d_{in} + M_{out}M_{in})^3)$ cost. Therefore our recipe avoids inverting and multiplying big matrices, resulting in a significant speed-up for conditional sampling.


\begin{table*}[t]
    \centering
    \caption{Computational complexity per layer. We assume $\weight \in \mathbb{R}^{d_{out} \times d_{in}}$, $\inducing \in \mathbb{R}^{M_{out} \times M_{in}}$, and $K$ forward passes for each of the $N$ inputs. ($^{*}$ uses a parallel computing friendly vectorisation technique \citep{wen2019batchensemble} for further speed-up.)}
    \label{tab:computational_complexity}
    \vspace{-0.5em}
    \scalebox{0.95}{
    \begin{tabular}{ccc}
        \toprule
        Method & Time complexity & Storage complexity \\ 
        \midrule
        Deterministic-$\weight$ & $\mathcal{O}(N d_{in} d_{out})$ & $\mathcal{O}(d_{in} d_{out})$\\
        \midrule
        \acrshort{ffg}-$\weight$ & $\mathcal{O}(N K d_{in} d_{out})$ & $\mathcal{O}(2 d_{in} d_{out})$ \\
        Ensemble-$\weight$ & $\mathcal{O}(N K d_{in} d_{out})$ & $\mathcal{O}(K d_{in} d_{out})$ \\
        Matrix-normal-$\weight$ & $\mathcal{O}(N K d_{in} d_{out})$ & $\mathcal{O}( d_{in} d_{out} + d_{in} + d_{out})$ \\
        $k$-tied \acrshort{ffg}-$\weight$ & $\mathcal{O}(N K d_{in} d_{out})$ & $\mathcal{O}( d_{in} d_{out} + k(d_{in} + d_{out}))$ \\
        rank-1 \acrshort{bnn} & $\mathcal{O}(N K d_{in} d_{out})^{*}$ & $\mathcal{O}( d_{in} d_{out} + 2(d_{in} + d_{out}))$ \\
        \midrule
        \acrshort{ffg}-$\inducing$ & $\mathcal{O}(N K d_{in} d_{out} + 2M_{in}^3 + 2M_{out}^3$ & $\mathcal{O}(d_{in}M_{in} + d_{out}M_{out} + 2M_{in}M_{out})$ \\
        & $+ K(d_{out}M_{out}M_{in} + M_{in}d_{out}d_{in}))$ &  \\
        Ensemble-$\inducing$ & same as above & $\mathcal{O}(d_{in}M_{in} + d_{out}M_{out} + K M_{in}M_{out})$ \\
        \bottomrule
    \end{tabular}
    }
    \vspace{-1em}
\end{table*}

\subsection{Computational complexities}
\label{sec:computation}

In \cref{tab:computational_complexity} we report the complexity figures for two types of inducing weight approaches: \acrshort{ffg} $q(\inducing)$ (\acrshort{ffg}-$\inducing$) and Delta mixture $q(\inducing)$ (Ensemble-$\inducing$). Baseline approaches include: Deterministic-$\weight$, variational inference with \acrshort{ffg} $q(\weight)$ \citep[\acrshort{ffg}-$\weight$,][]{blundell2015weight}, deep ensemble in $\weight$ \citep[Ensemble-$\weight$,][]{lakshminarayanan2017simple}, as well as parameter efficient approaches such as matrix-normal $q(\weight)$ (Matrix-normal-$\weight$, \citet{louizos2017multiplicative}), variational inference with $k$-tied \acrshort{ffg} $q(\weight)$ ($k$-tied \acrshort{ffg}-$\weight$, \citet{swiatkowski2020k}), and rank-1 \acrshort{bnn} \citep{dusenberry2020efficient}. The gain in memory is significant for the inducing weight approaches, in fact with $M_{in} < d_{in}$ and $M_{out} < d_{out}$ the parameter storage requirement is smaller than a single deterministic neural network. The major overhead in run-time comes from the extended Matheron's rule for sampling $q(\weight | \inducing)$. Some of the computations there are performed only once, and in our experiments we show that by using a relatively low-dimensional $\inducing$ and large batch-sizes, the overhead is acceptable.

\section{Experiments}



We evaluate the inducing weight approaches on regression, classification and related uncertainty estimation tasks. The goal is to demonstrate competitive performance to popular $\weight$-space uncertainty estimation methods while using significantly fewer parameters. 
We acknowledge that existing parameter efficient approaches for uncertainty estimation (e.g.~k-tied or rank-1 BNNs) have achieved close performance to deep ensembles. However, \emph{none} of them reduces the parameter count to be smaller than that of a \emph{single} network. Therefore we decide \emph{not} to include these baselines and instead focus on comparing: (1) variational inference with \acrshort{ffg} $q(\weight)$ \citep[\acrshort{ffg}-$\weight$,][]{blundell2015weight} v.s.~\acrshort{ffg} $q(\inducing)$ (\acrshort{ffg}-$\inducing$, ours); (2) deep ensemble in $\weight$ space \citep[Ensemble-$\weight$,][]{lakshminarayanan2017simple} v.s.~ensemble in $\inducing$ space (Ensemble-$\inducing$, ours). Another baseline is training a deterministic neural network with maximum likelihood. Details and additional results are in \cref{app:experiment_details,app:additional_results}.

\subsection{Synthetic 1-D regression}

\begin{figure}[t]
\centering     
\subfigure[\acrshort{ffg}-$\inducing$]{\label{fig:regression_inducing_ffg}
\includegraphics[width=0.15\linewidth]{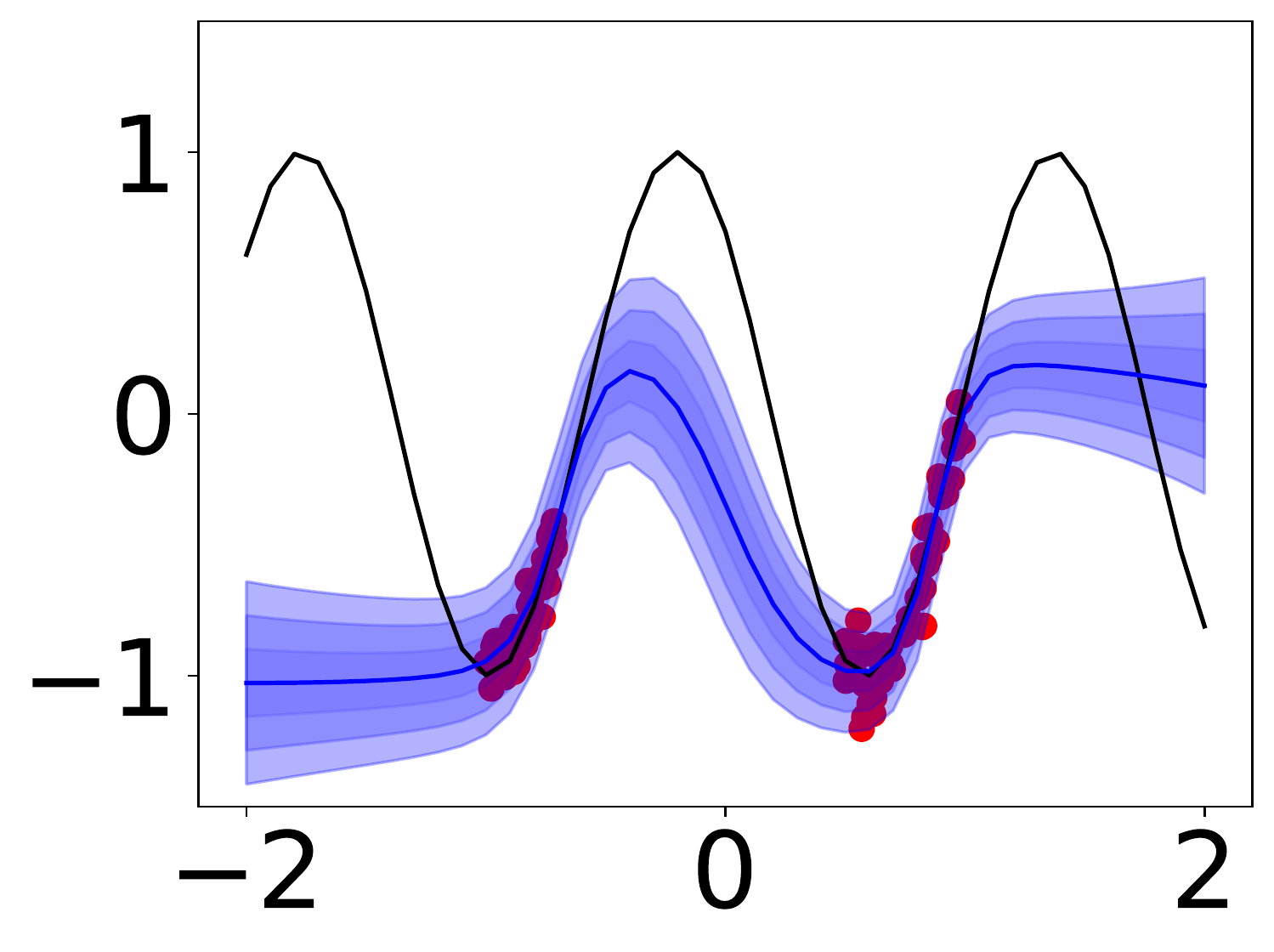}}
\hfill
\subfigure[\acrshort{fcg}-$\inducing$]{\label{fig:regression_inducing_fcg}
\includegraphics[width=0.15\linewidth]{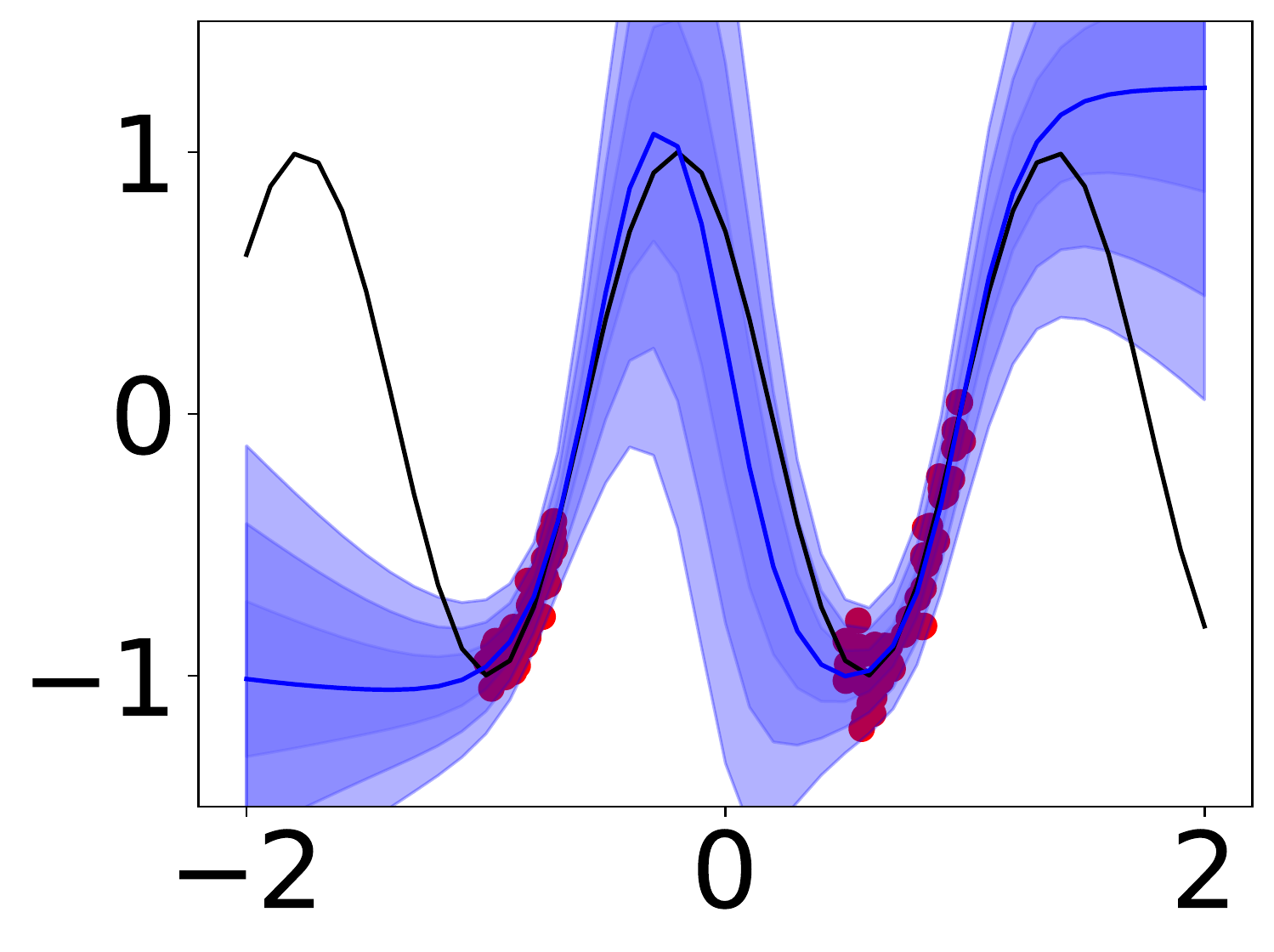}}
\hfill
\subfigure[Ensemble-$\inducing$]{\label{fig:regression_inducing_ensemble}
\includegraphics[width=0.15\linewidth]{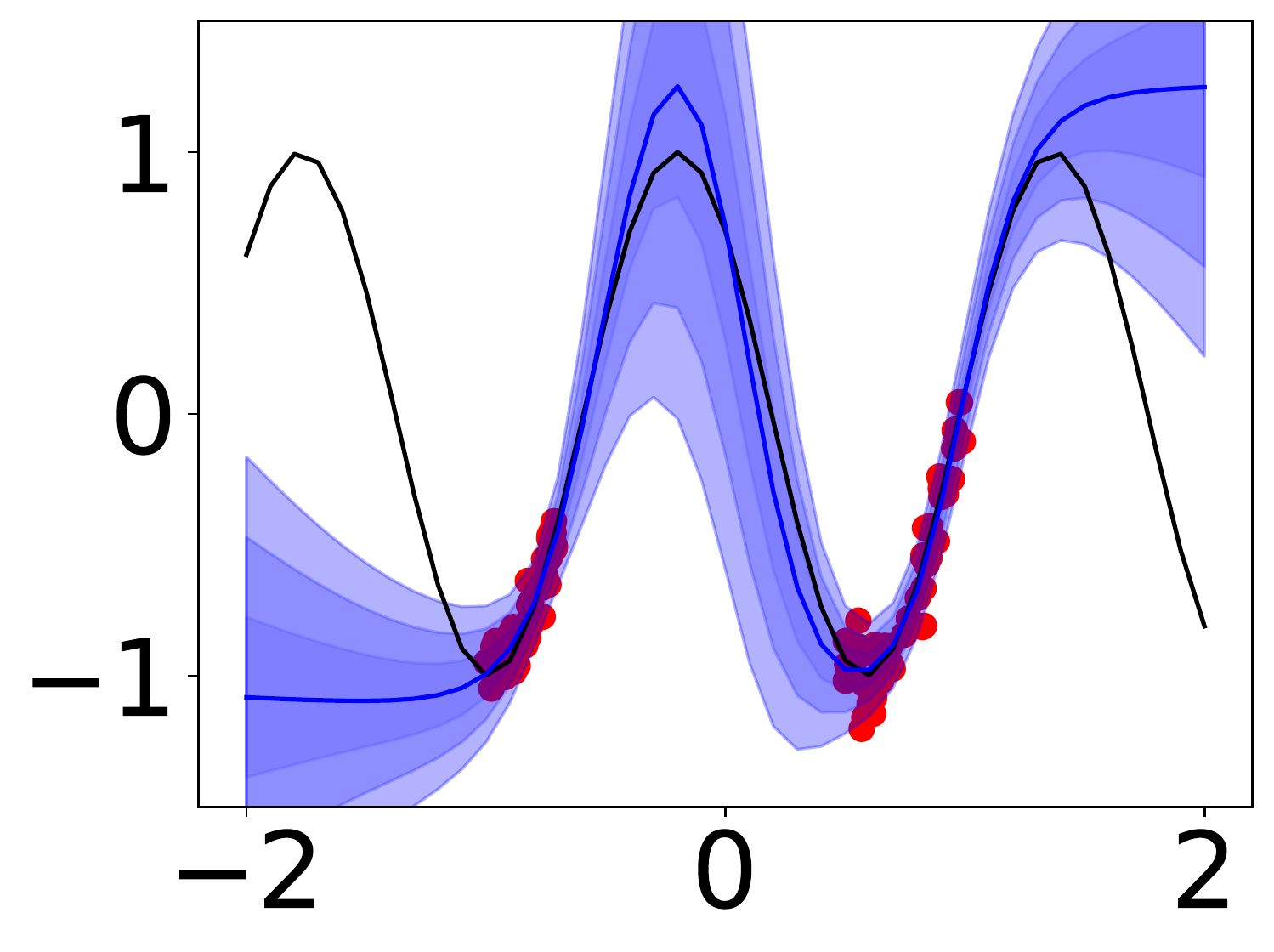}}
\hfill
\subfigure[\acrshort{ffg}-$\weight$]{\label{fig:regression_mfvi}
\includegraphics[width=0.15\linewidth]{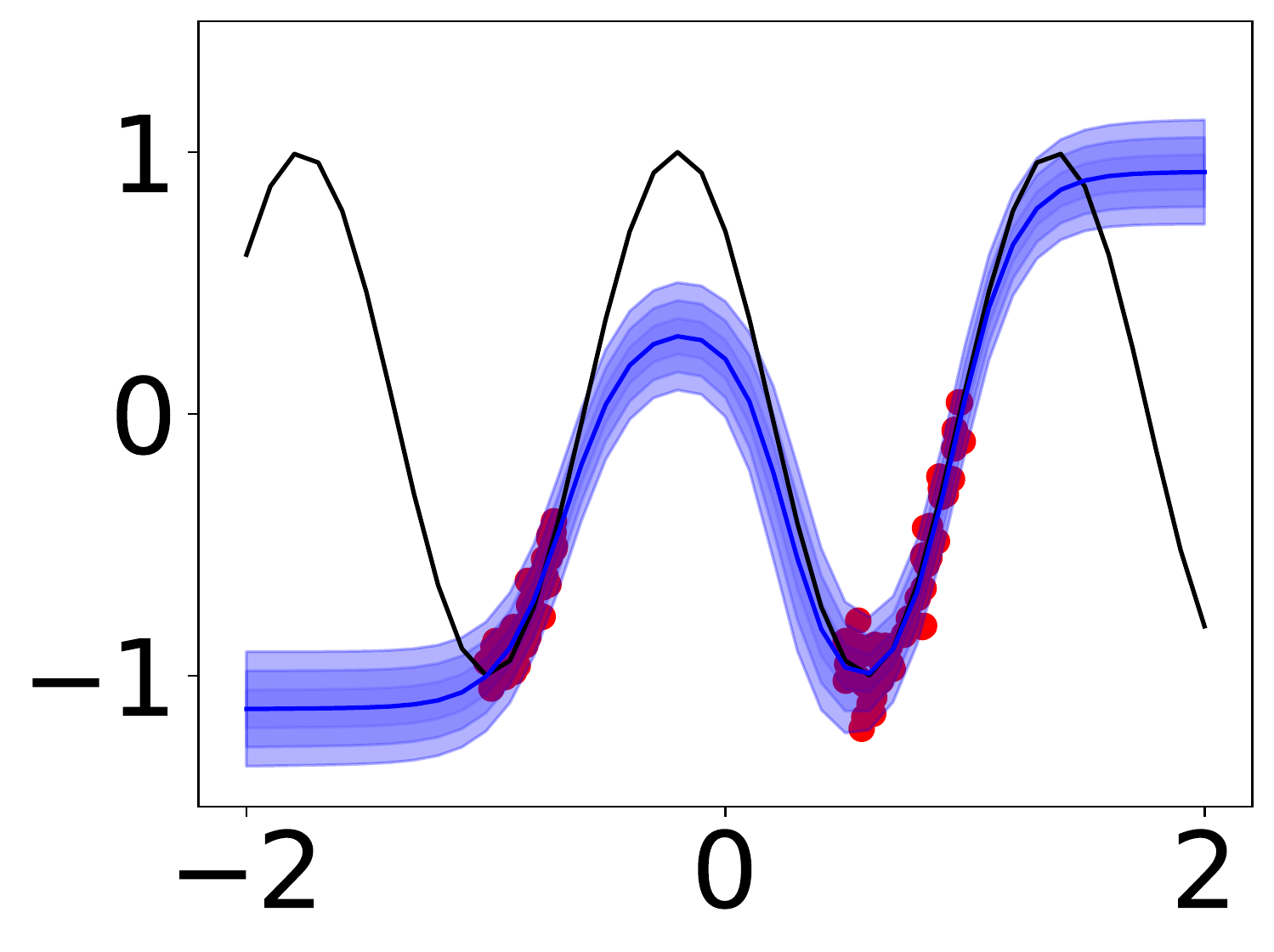}}
\hfill
\subfigure[\acrshort{fcg}-$\weight$]{\label{fig:regression_fcg}
\includegraphics[width=0.15\linewidth]{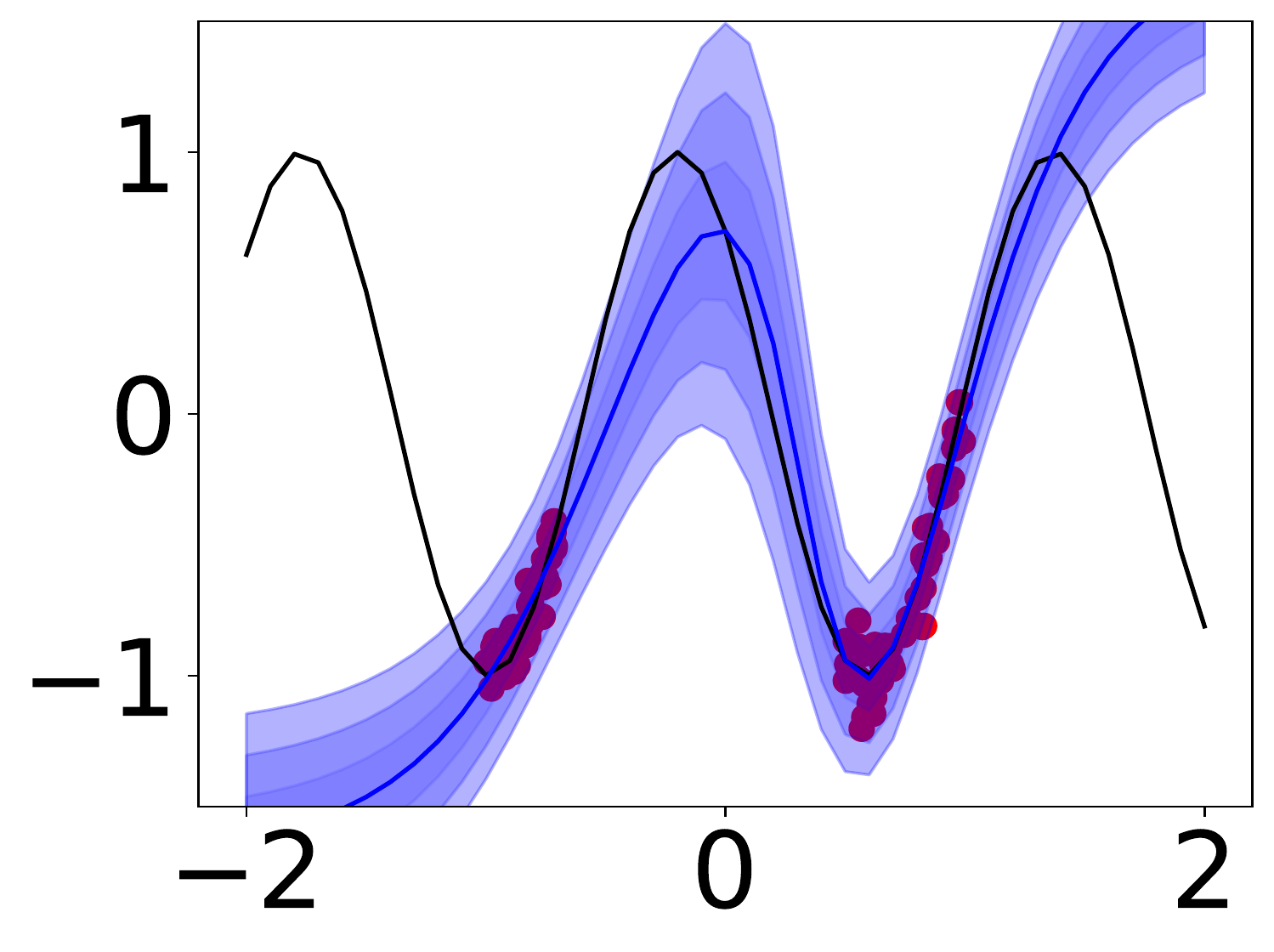}}
\hfill
\subfigure[NUTS]{\label{fig:regression_nuts}
\includegraphics[width=0.15\linewidth]{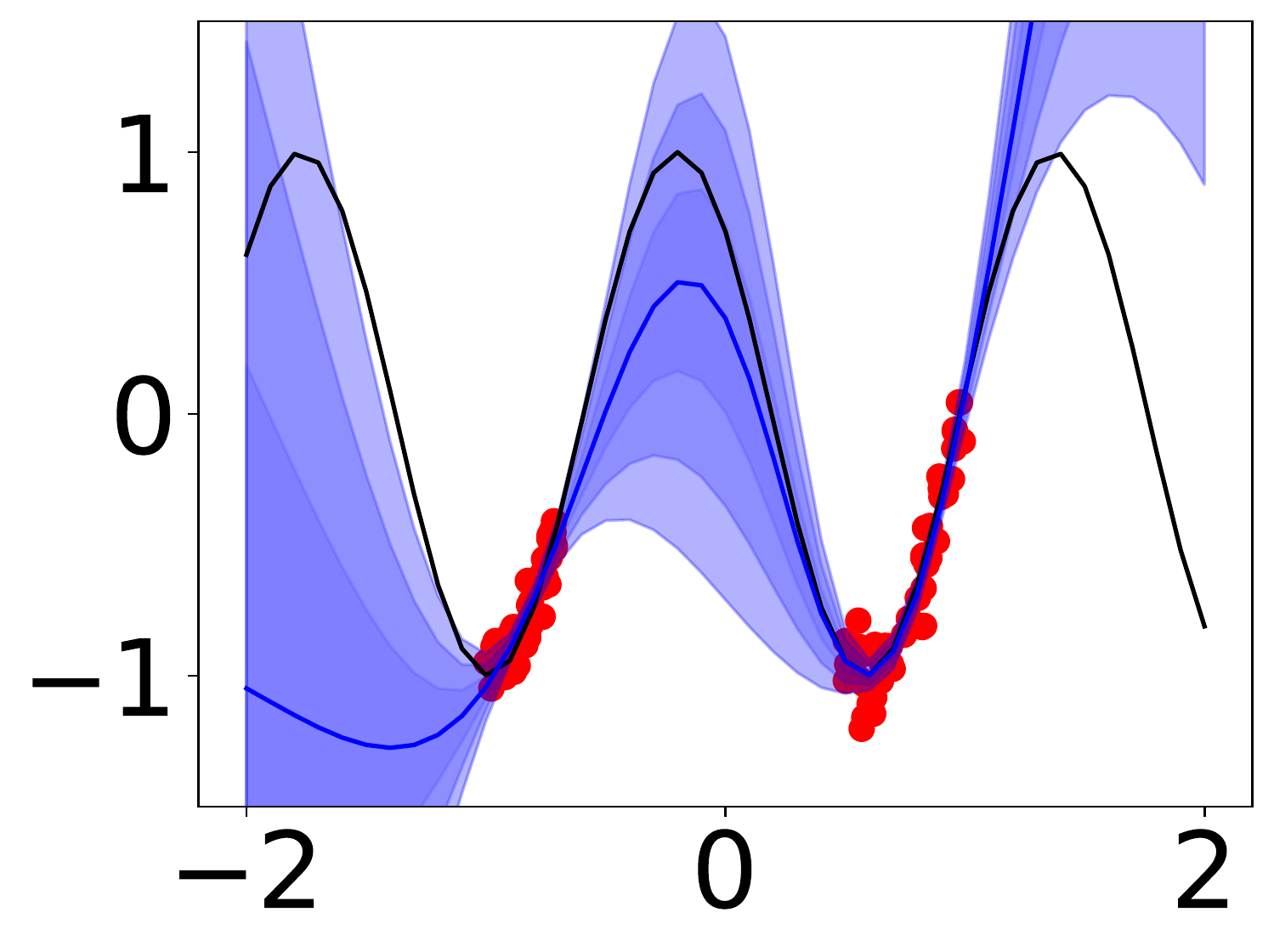}}
\vspace{-0.5em}
\caption{Toy regression results, with observations in red dots and the ground truth function in black.}
\label{fig:regression_results}
\vspace{-0.5em}
\end{figure}

The regression task follows \citet{foong2019between}, which has two input clusters $x_1 \sim \mathcal{U}\left[-1, -0.7\right]$, $x_2 \sim \mathcal{U}\left[0.5, 1\right]$, and targets $y \sim \mathcal{N}(\cos(4x + 0.8), 0.01)$.
For reference we show the exact posterior results using the NUTS sampler \citep{hoffman2014no}.
The results are visualised in \cref{fig:regression_results} with predictive mean in blue, and up to three standard deviations as shaded area.
Similar to historical results, \acrshort{ffg}-$\weight$ fails to represent the increased uncertainty away from the data and in between clusters.
While underestimating predictive uncertainty overall, \acrshort{ffg}-$\inducing$ shows a small increase in predictive uncertainty away from the data.
In contrast, a per-layer \acrfull{fcg} in both weight (\acrshort{fcg}-$\weight$) and inducing space (\acrshort{fcg}-$\inducing$) as well as Ensemble-$\inducing$ better capture the increased predictive variance, although the mean function is more similar to that of \acrshort{ffg}-$\weight$.

\subsection{\mbox{Classification and in-distribution calibration}}

\begin{figure}[t]
\centering
\begin{minipage}{0.52\linewidth}
\captionof{table}{CIFAR in-distribution metrics (in $\%$).}
\label{tab:indistribution}
\newcommand{\mainuncertaintytable}{\input{tables/main_uncertainty.tex}}
\scalebox{0.9}{
    \begin{tabular}{l|cccc}
    \input{tables/main_uncertainty.tex}
    \end{tabular}
}
\end{minipage}
\hfill
\begin{minipage}{0.45\linewidth}
\includegraphics[width=1\linewidth]{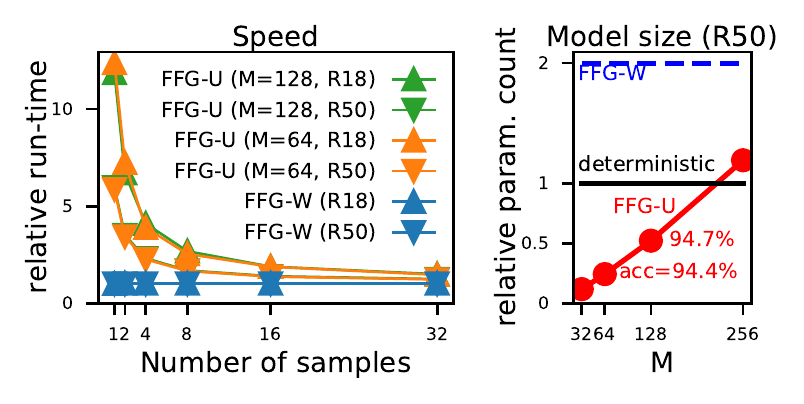}
\vspace{-1.5em}
\captionof{figure}{Resnet run-times \& model sizes.}
\label{fig:speed_memory}
\end{minipage}
\vspace{-1em}
\end{figure}

As the core empirical evaluation, we train Resnet-50 models \citep{he2016identity} on CIFAR-10 and CIFAR-100 \citep{krizhevsky2009cifar}.
To avoid underfitting issues with \acrshort{ffg}-W, a useful trick is to set an upper limit $\sigma_{max}^2$ on the variance of $q(\weight)$ \citep{louizos2017multiplicative}.
This trick is similarly applied to the $\inducing$-space methods, where we cap $\lambda \leq \lambda_{max}$ for $q(\weight | \inducing)$, and for \acrshort{ffg}-$\inducing$ we also set $\sigma_{max}^2$ for the variance of $q(\inducing)$.
In convolution layers, we treat the 4D weight tensor $\weight$ of shape $(c_{out}, c_{in}, h, w)$ as a $c_{out} \times c_{in}hw$ matrix.
We use $\inducing$ matrices of shape $64 \times 64$ for all layers (i.e.~$M=M_{in}=M_{out}=64$), except that for CIFAR-10 we set $M_{out}=10$ for the last layer.
%

In \cref{tab:indistribution} we report test accuracy and test \acrfull{ece} \citep{guo2017calibration} as a first evaluation of the uncertainty estimates.
Overall, Ensemble-$\weight$ achieves the highest accuracy, but is not as well-calibrated as variational methods.
For the inducing weight approaches, Ensemble-$\inducing$ outperforms \acrshort{ffg}-$\inducing$ on both datasets; overall it performs the best on the more challenging CIFAR-100 dataset (close-to-Ensemble-$\weight$ accuracy and lowest \acrshort{ece}).
\cref{tab:indistribution_full_resnet50_c10,tab:indistribution_full_resnet50_c100} in \cref{app:additional_results} show that increasing the $\inducing$ dimensions to $M=128$ improves accuracy but leads to slightly worse calibration.

In \cref{fig:speed_memory} we show prediction run-times for batch-size $=500$ on an NVIDIA Tesla V100 GPU, relative to those of an ensemble of deterministic nets, as well as relative parameter sizes to a single ResNet-50.
The extra run-times for the inducing methods come from computing the extended Matheron's rule. However, as they can be calculated once and cached for drawing multiple samples, the overhead reduces to a small factor when using larger number of samples $K$, especially for the bigger Resnet-$50$.
More importantly, when compared to a \emph{deterministic} ResNet-50, the inducing weight models reduce the parameter count by over $75\%$ ($5,710,902$ vs.~$23,520,842$) for  $M=64$.

\begin{figure}[b]
\centering
\includegraphics[width=0.95\linewidth]{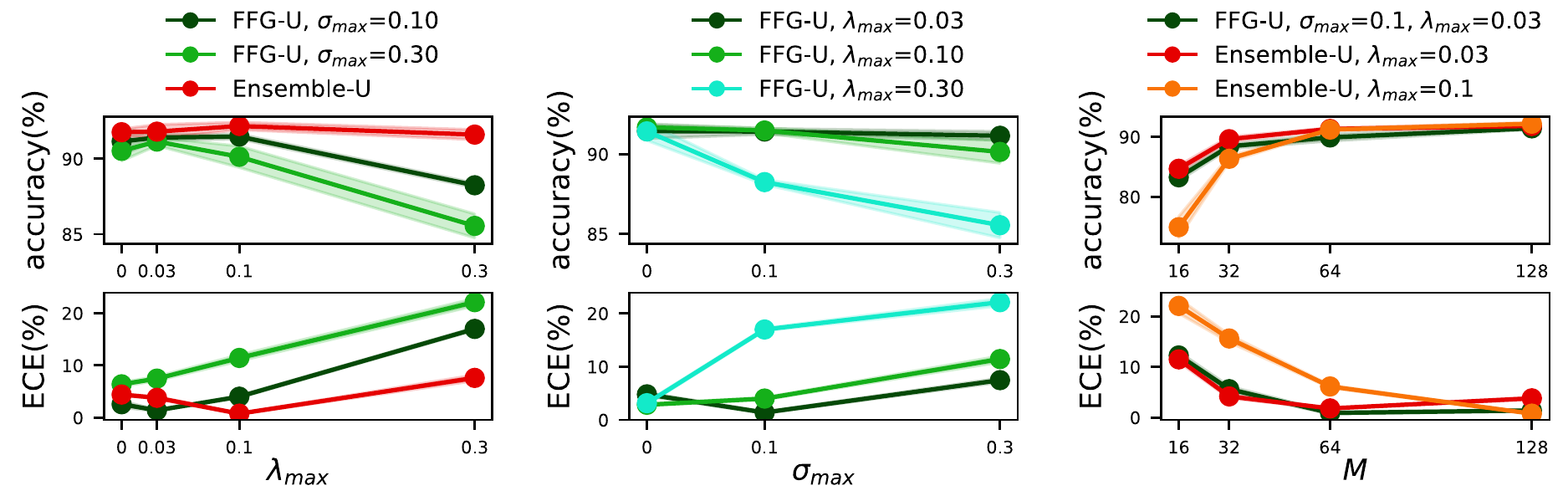}
\vspace{-0.5em}
\caption{Ablation study: average CIFAR-10 accuracy ($\uparrow$) and \acrshort{ece} ($\downarrow$) for the inducing weight methods on ResNet-18. In the first two columns $M=128$ for $\inducing$ dimensions. For $\lambda_{max}, \sigma_{max} = 0$ we use point estimates for $\inducing, \weight$ respectively.}
\label{fig:cifar10_ablation_study}
\vspace{-1.5em}
\end{figure}

\paragraph{Hyper-parameter choices} We visualise in \cref{fig:cifar10_ablation_study} the accuracy and \acrshort{ece} results for computationally lighter inducing weight ResNet-18 models with different hyper-parameters (see \cref{app:experiment_details}). Performance in both metrics improves as the $\inducing$ matrix size $M$ is increased (right-most panels), and the results for $M=64$ and $M=128$ are fairly similar. Also setting proper values for $\lambda_{max}, \sigma_{max}$ is key to the improved results. The left-most panels show that with fixed $\sigma_{max}$ values (or Ensemble-$\inducing$), the preferred conditional variance cap values $\lambda_{max}$ are fairly small (but still larger than $0$ which corresponds to a point estimate for $\weight$ given $\inducing$). For $\sigma_{max}$ which controls variance in $\inducing$ space, we see from the top middle panel that the accuracy metric is fairly robust to $\sigma_{max}$ as long as $\lambda_{max}$ is not too large. But for \acrshort{ece}, a careful selection of $\sigma_{max}$ is required (bottom middle panel).

\subsection{Robustness, out-of-distribution detection and pruning}

\begin{figure}[t]
\centering
\includegraphics[width=0.95\linewidth]{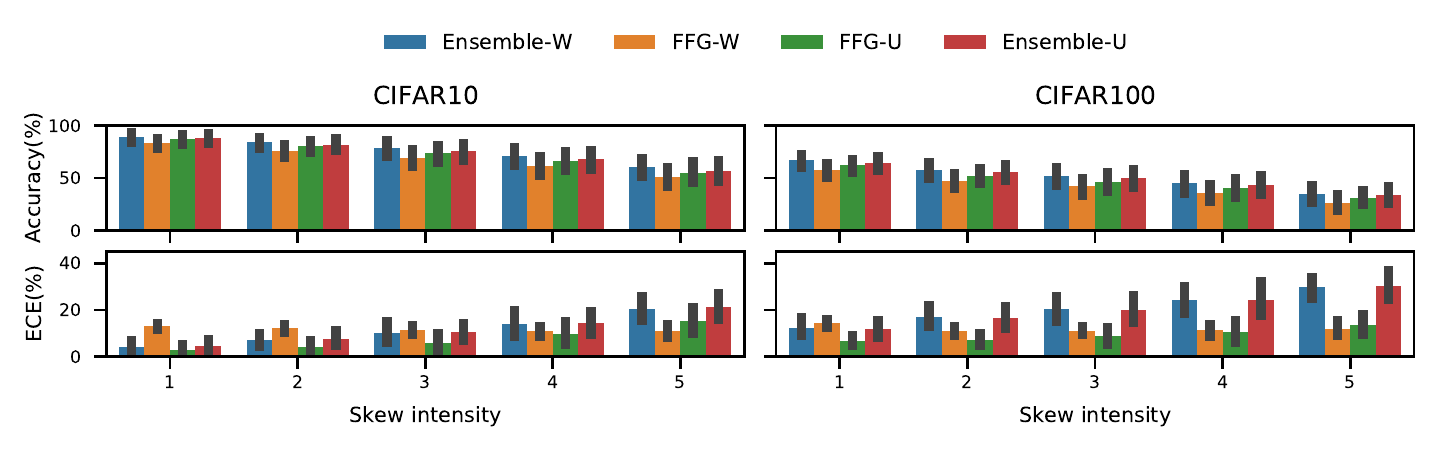}
\vspace{-0.5em}
\caption{Mean${\pm}$two errs. for Acc$\uparrow$ and \acrshort{ece}$\downarrow$ on corrupted CIFAR \citep{hendrycks2019benchmarking}.}
\label{fig:cifar_corrupted}
\vspace{-1em}
\end{figure}

\begin{table}[t]
\centering
\caption{\acrshort{ood} detection metrics for Resnet-$50$ trained on CIFAR$10$/$100$.}
\label{tab:ood_detection}
\newcommand{\mainaucoodtable}{\toprule
In-dist $\rightarrow$ OOD & \multicolumn{2}{c}{C10 $\rightarrow$ C100} & \multicolumn{2}{c}{C10 $\rightarrow$ SVHN} & \multicolumn{2}{c}{C100 $\rightarrow$ C10} & \multicolumn{2}{c}{C100 $\rightarrow$ SVHN}\\
Method / Metric & AUROC & AUPR & AUROC & AUPR & AUROC & AUPR & AUROC & AUPR\\
\midrule
Deterministic & $.84 {\scriptstyle {\pm} .00}$ & $.80 {\scriptstyle {\pm} .00}$ & $.93 {\scriptstyle {\pm} .01}$ & $.85 {\scriptstyle {\pm} .01}$ & $.74 {\scriptstyle {\pm} .00}$ & $.74 {\scriptstyle {\pm} .00}$ & $.81 {\scriptstyle {\pm} .01}$ & $.72 {\scriptstyle {\pm} .02}$ \\
Ensemble-W & $.89 {\scriptstyle \phantom{{\pm} .00}}$ & $.89 {\scriptstyle \phantom{{\pm} .00}}$ & $.95 {\scriptstyle \phantom{{\pm} .00}}$ & $.92 {\scriptstyle \phantom{{\pm} .00}}$ & $.78 {\scriptstyle \phantom{{\pm} .00}}$ & $.79 {\scriptstyle \phantom{{\pm} .00}}$ & $.86 {\scriptstyle \phantom{{\pm} .00}}$ & $.78 {\scriptstyle \phantom{{\pm} .00}}$ \\
FFG-W & $.88 {\scriptstyle {\pm} .00}$ & $.90 {\scriptstyle {\pm} .00}$ & $.90 {\scriptstyle {\pm} .01}$ & $.86 {\scriptstyle {\pm} .01}$ & $.76 {\scriptstyle {\pm} .00}$ & $.79 {\scriptstyle {\pm} .00}$ & $.80 {\scriptstyle {\pm} .01}$ & $.69 {\scriptstyle {\pm} .01}$ \\
FFG-U & $.89 {\scriptstyle {\pm} .00}$ & $.91 {\scriptstyle {\pm} .00}$ & $.94 {\scriptstyle {\pm} .01}$ & $.91 {\scriptstyle {\pm} .01}$ & $.77 {\scriptstyle {\pm} .00}$ & $.79 {\scriptstyle {\pm} .00}$ & $.83 {\scriptstyle {\pm} .01}$ & $.74 {\scriptstyle {\pm} .01}$ \\
Ensemble-U & $.90 {\scriptstyle {\pm} .00}$ & $.91 {\scriptstyle {\pm} .00}$ & $.93 {\scriptstyle {\pm} .00}$ & $.91 {\scriptstyle {\pm} .00}$ & $.77 {\scriptstyle {\pm} .00}$ & $.79 {\scriptstyle {\pm} .00}$ & $.82 {\scriptstyle {\pm} .01}$ & $.72 {\scriptstyle {\pm} .02}$ \\
\bottomrule
}
\scalebox{0.9}{
    \begin{tabular}{l|cccccccc}
    \toprule
In-dist $\rightarrow$ OOD & \multicolumn{2}{c}{C10 $\rightarrow$ C100} & \multicolumn{2}{c}{C10 $\rightarrow$ SVHN} & \multicolumn{2}{c}{C100 $\rightarrow$ C10} & \multicolumn{2}{c}{C100 $\rightarrow$ SVHN}\\
Method / Metric & AUROC & AUPR & AUROC & AUPR & AUROC & AUPR & AUROC & AUPR\\
\midrule
Deterministic & $.84 {\scriptstyle {\pm} .00}$ & $.80 {\scriptstyle {\pm} .00}$ & $.93 {\scriptstyle {\pm} .01}$ & $.85 {\scriptstyle {\pm} .01}$ & $.74 {\scriptstyle {\pm} .00}$ & $.74 {\scriptstyle {\pm} .00}$ & $.81 {\scriptstyle {\pm} .01}$ & $.72 {\scriptstyle {\pm} .02}$ \\
Ensemble-W & $.89 {\scriptstyle \phantom{{\pm} .00}}$ & $.89 {\scriptstyle \phantom{{\pm} .00}}$ & $.95 {\scriptstyle \phantom{{\pm} .00}}$ & $.92 {\scriptstyle \phantom{{\pm} .00}}$ & $.78 {\scriptstyle \phantom{{\pm} .00}}$ & $.79 {\scriptstyle \phantom{{\pm} .00}}$ & $.86 {\scriptstyle \phantom{{\pm} .00}}$ & $.78 {\scriptstyle \phantom{{\pm} .00}}$ \\
FFG-W & $.88 {\scriptstyle {\pm} .00}$ & $.90 {\scriptstyle {\pm} .00}$ & $.90 {\scriptstyle {\pm} .01}$ & $.86 {\scriptstyle {\pm} .01}$ & $.76 {\scriptstyle {\pm} .00}$ & $.79 {\scriptstyle {\pm} .00}$ & $.80 {\scriptstyle {\pm} .01}$ & $.69 {\scriptstyle {\pm} .01}$ \\
FFG-U & $.89 {\scriptstyle {\pm} .00}$ & $.91 {\scriptstyle {\pm} .00}$ & $.94 {\scriptstyle {\pm} .01}$ & $.91 {\scriptstyle {\pm} .01}$ & $.77 {\scriptstyle {\pm} .00}$ & $.79 {\scriptstyle {\pm} .00}$ & $.83 {\scriptstyle {\pm} .01}$ & $.74 {\scriptstyle {\pm} .01}$ \\
Ensemble-U & $.90 {\scriptstyle {\pm} .00}$ & $.91 {\scriptstyle {\pm} .00}$ & $.93 {\scriptstyle {\pm} .00}$ & $.91 {\scriptstyle {\pm} .00}$ & $.77 {\scriptstyle {\pm} .00}$ & $.79 {\scriptstyle {\pm} .00}$ & $.82 {\scriptstyle {\pm} .01}$ & $.72 {\scriptstyle {\pm} .02}$ \\
\bottomrule

    \end{tabular}
}
\vspace{-1em}
\end{table}

To investigate the models' robustness to distribution shift, we compute predictions on corrupted CIFAR datasets \citep{hendrycks2019benchmarking} after training on clean data.
\cref{fig:cifar_corrupted} shows accuracy and \acrshort{ece} results for the ResNet-50 models.
Ensemble-$\weight$ is the most accurate model across skew intensities, while \acrshort{ffg}-$\weight$, though performing well on clean data, returns the worst accuracy under perturbation. The inducing weight methods perform competitively to Ensemble-$\weight$ with Ensemble-$\inducing$ being slightly more accurate than \acrshort{ffg}-$\inducing$ as on the clean data.
For \acrshort{ece}, \acrshort{ffg}-$\inducing$ outperforms Ensemble-$\inducing$ and Ensemble-$\weight$, which are similarly calibrated. Interestingly, while the accuracy of \acrshort{ffg}-$\weight$ decays quickly as the data is perturbed more strongly, its \acrshort{ece} remains roughly constant.

\cref{tab:ood_detection} further presents the utility of the maximum predicted probability for \acrfull{ood} detection. 
The metrics are the area under the receiver operator characteristic (AUROC) and the precision-recall curve (AUPR).
The inducing-weight methods perform similarly to Ensemble-$\weight$; all three outperform \acrshort{ffg}-$\weight$ and deterministic networks across the board.

\begin{wrapfigure}[13]{r}{0.3\linewidth}
    \vspace{-2em}
    \centering
    \includegraphics[width=\linewidth]{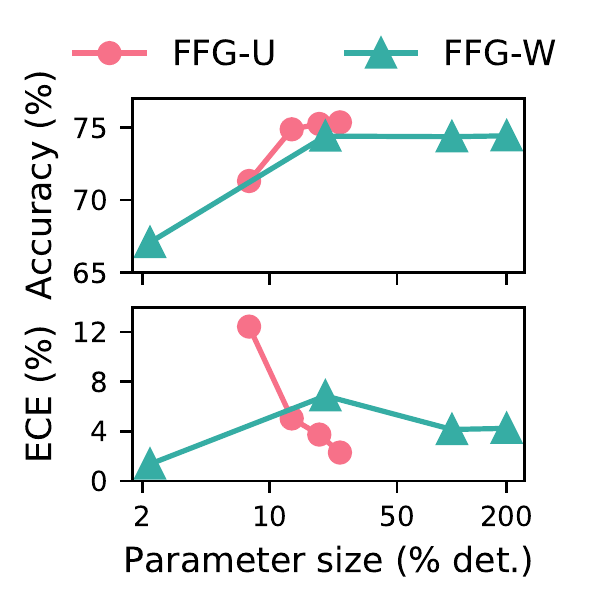}
    \vspace{-1em}
    \caption{CIFAR100 pruning accuracy($\uparrow$) \& \acrshort{ece}($\downarrow$). Rightmost points are w/out pruning.}
    \label{fig:pruning}
\end{wrapfigure}

\paragraph{Parameter pruning} We further investigate pruning as a pragmatic alternative for more parameter-efficient inference.
For \acrshort{ffg}-$\inducing$, we prune entries of the $Z$ matrices, which contribute the largest number of parameters to the inducing methods, with the smallest magnitude.
For \acrshort{ffg}-$\weight$ we follow \citet{graves2011practical} in setting different fractions of $\weight$ to $0$  depending on their variational mean-to-variance ratio and repeat the previous experiments after fine-tuning the distributions on the remaining variables.
We stress that, unlike \acrshort{ffg}-$\inducing$, the \acrshort{ffg}-$\weight$ pruning corresponds to a post-hoc change of the probabilistic model and no longer performs inference in the original weight-space.

For \acrshort{ffg}-$\weight$, pruning $90\%$ of the parameters (leaving $20\%$ of parameters as compared to its deterministic counterpart) worsens the \acrshort{ece}, in particular on CIFAR100, see \cref{fig:pruning}.
Further pruning to $1\%$ worsens the accuracy and the \acrshort{ood} detection results as well.
On the other hand, pruning $50\%$ of the $\covparam$ matrices for \acrshort{ffg}-$\inducing$ reduces the parameter count to $13.2\%$ of a deterministic net, at the cost of only slightly worse calibration.
See \cref{app:additional_results} for the full results.

\section{Discussions}
\subsection{A function-space perspective on inducing weights}
\label{sec:function_space_view}

Although the inducing weight approach performs approximate inference in weight space, we present in \cref{app:function_space_inducing_weight} a function-space inference perspective of the proposed method, showing its deep connections to sparse \acrshortpl{gp}. Our analysis considers the function-space behaviour of each network layer's output and discusses the corresponding interpretations of the $\inducing$ variables and $\covparam$ parameters. 

\begin{wrapfigure}[13]{r}{0.3\linewidth}
    \vspace{-1em}
    \centering
    \includegraphics[width=\linewidth]{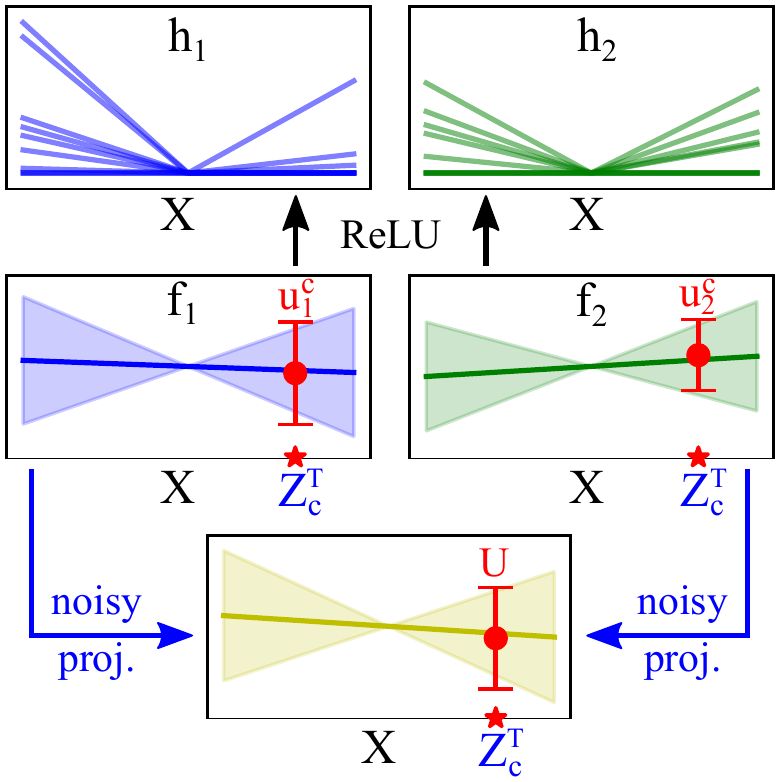}
    \caption{Visualising $\inducing$ variables in pre-activation spaces.}
    \label{fig:inducing_weights_function_space}
\end{wrapfigure}

The interpretations are visualised in \cref{fig:inducing_weights_function_space}. Similar to sparse \acrshortpl{gp}, in each layer, the $\covparam_c$ parameters can be viewed as the (transposed) inducing input locations which lie in the same space as the layer's input. The $\inducing_c$ variables can also be viewed as the corresponding (noisy) inducing outputs that lie in the pre-activation space.  
Given that the output dimension $d_{out}$ can still be high (e.g.~$>1000$ in a fully connected layer), our approach performs further dimension reduction in a similar spirit as probabilistic PCA \citep{tipping1999probabilistic}, which projects the \emph{column vectors} of $\inducing_c$ to a lower-dimensional space. This returns the inducing weight variables $\inducing$, and the projection parameters are $\{\covparam_r, \precparam_r\}$. Combining the two steps, it means the column vectors of $\inducing$ can be viewed as collecting the ``noisy projected inducing outputs'' whose corresponding ``inducing inputs'' are row vectors of $\covparam_c$ (see the red bars in \cref{fig:inducing_weights_function_space}). 

In \cref{app:function_space_inducing_weight} we further derive the resulting variational objective from the function-space view, which is almost identical to \cref{eq:elbo_inducing_weight_reduced}, except for scaling coefficients on the $R(\lambda_l)$ terms to account for the change in dimensionality from weight space to function space. This result nicely connects posterior inference in weight- and function-space.

\subsection{Related work}

\paragraph{Parameter-efficient uncertainty quantification methods}
Recent research has proposed Gaussian posterior approximations for \acrshortpl{bnn} with efficient covariance structure \citep{ritter2018scalable,zhang2018noisy,mishkin2018slang}. The inducing weight approach differs from these in introducing structure via a hierarchical posterior with low-dimensional auxiliary variables.
Another line of work reduces the memory overhead via efficient parameter sharing \citep{louizos2017multiplicative,wen2019batchensemble,swiatkowski2020k,dusenberry2020efficient}. The third category of work considers a hybrid approach, where only a selective part of the neural network receives Bayesian treatments, and the other weights remain deterministic \citep{bradshaw2017adversarial,daxberger2020expressive}. However, both types of approaches maintain a ``mean parameter'' for the weights, making the memory footprint at least that of storing a deterministic neural network. Instead, our approach shares parameters via the augmented prior with efficient low-rank structure, \emph{reducing} the memory use compared to a deterministic network.
In a similar spirit, \citet{izmailov2020subspace} perform inference in a $d$-dimensional sub-space obtained from PCA on weights collected from an SGD trajectory. But this approach does not leverage the layer-structure of neural networks and requires $d\times$ memory of a single network.

\paragraph{Network pruning in uncertainty estimation context}
There is a large amount of existing research advocating network pruning approaches for parameter-efficient deep learning, e.g. see \citet{han2015deep,frankle2018lottery,lee2018snip}. In this regard, mean-field VI approaches have also shown success in network pruning, but only in terms of maintaining a minimum accuracy level \citep{graves2011practical,louizos2017bayesian,havasi2018minimal}. To the best of our knowledge, our empirical study presents the first evaluation for VI-based pruning methods in maintaining uncertainty estimation quality. \citet{deng2019adaptive} considers pruning BNNs with stochastic gradient Langevin dynamics \citep{welling2011bayesian} as the inference engine. The inducing weight approach is orthogonal to these BNN pruning approaches, as it leaves the prior on the network parameters intact, while the pruning approaches correspond to a post-hoc change of the probabilistic model to using a sparse weight prior. Indeed our parameter pruning experiments showed that our approach can be combined with network pruning to achieve further parameter efficiency improvements.

\paragraph{Sparse \acrshort{gp} and function-space inference}
As \acrshortpl{bnn} and \acrshortpl{gp} are closely related \citep{neal1995bayesian,matthews2018gaussian,lee2018deep}, recent efforts have introduced \acrshort{gp}-inspired techniques to \acrshortpl{bnn} \citep{ma2019variational,sun2018functional,khan2019approximate,ober2020bnn}. Compared to weight-space inference, function-space inference is appealing as its uncertainty is more directly relevant for predictive uncertainty estimation. While the inducing weight approach performs computations in weight-space, \cref{sec:function_space_view} establishes the connection to function-space posteriors. Our approach is related to sparse deep \acrshort{gp} methods with $\inducing_c$ having similar interpretations as inducing outputs in e.g. \citet{salimbeni2017doubly}. The major difference is that $\inducing$ lies in a low-dimensional space, projected from the pre-activation output space of a network layer. 

The original Matheron's rule \citep{journel1978mining,hoffman1991constrained,doucet2010gaussian} for sampling from conditional multivariate Gaussian distributions has recently been applied to speed-up sparse GP inference \citep{wilson2020efficient,wilson2021pathconditioning}. As explained in \cref{sec:conditional_sampling_matheron}, direct application of the original rule to sampling $\weight$ conditioned on $\inducing$ still incurs prohibitive cost as $p(\vect(\weight), \vect(\inducing))$ does not have a convenient factorisation form. Our extended Matheron's rule addresses this issue by exploiting the efficient factorisation structure of the joint matrix normal distribution $p(\weight, \inducing_c, \inducing_r, \inducing)$, reducing the dominating factor of computation cost from cubic ($\mathcal{O}(d_{out}^3 d_{in}^3)$) to linear ($\mathcal{O}(d_{out} d_{in})$). We expect this new rule to be useful for a wide range of models/applications beyond BNNs, such as matrix-variate Gaussian processes \citep{stegle2011efficient}.

\paragraph{Priors on neural network weights}
Hierarchical priors for weights has also been explored \citep{louizos2017bayesian,krueger2017bayesian,atanov2018the,ghosh2019model,karaletsos2020hierarchical}. However, we emphasise that $\tilde{p}(\weight, \inducing)$ is a \emph{pseudo prior} that is constructed to \emph{assist posterior inference} rather than to \emph{improve model design}. Indeed, parameters associated with the inducing weights are optimisable for improving posterior approximations. Our approach can be adapted to other priors, e.g.~for a Horseshoe prior $p(\theta, \nu) = p(\theta | \nu) p(\nu) = \mathcal{N}(\theta; 0, \nu^2) C^{+}(\nu; 0, 1)$, the pseudo prior can be defined as $\tilde{p}(\theta, \nu, a) = \tilde{p}(\theta | \nu, a) \tilde{p}(a) p(\nu)$ such that $\int \tilde{p}(\theta | \nu, a) \tilde{p}(a) d a = p(\theta | \nu)$. In general, pseudo priors have found broader success in Bayesian computation \citep{carlin1995bayesian}.

\section{Conclusion}
We have proposed a parameter-efficient uncertainty quantification framework for neural networks. It augments each of the network layer weights with a small matrix of inducing weights, and by extending Matheron's rule to matrix-normal related distributions, maintains a relatively small run-time overhead compared with ensemble methods. Critically, experiments on prediction and uncertainty estimation tasks show the competence of the inducing weight methods to the state-of-the-art, while reducing the parameter count to under a quarter of a deterministic ResNet-50 before pruning. This represents a significant improvement over prior Bayesian and deep ensemble techniques, which so far have not managed to go below this threshold despite various attempts of matching it closely.

Several directions are to be explored in the future. First, modelling correlations across layers might further improve the inference quality. We outline an initial approach leveraging inducing variables in \cref{app:hierarchical_inducing}.
Second, based on the function-space interpretation of inducing weights, better initialisation techniques can be inspired from the sparse \acrshort{gp} and dimension reduction literature. 
Similarly, this interpretation might suggest other innovative pruning approaches for the inducing weight method, thereby achieving further memory savings. 
Lastly, the run-time overhead of our approach can be mitigated by a better design of the inducing weight structure as well as vectorisation techniques amenable to parallelised computation. Designing hardware-specific implementations of the inducing weight approach is also a viable alternative for such purposes.



\bibliography{references}
\bibliographystyle{icml2021}

\clearpage
\appendix

\section{Notation}

Generally, we use regular font letters $a, A$ to denote scalars, lower-case bold symbols $\rva$ to denote vectors and upper-case bold symbols $\rmA$ for matrices.

\begin{table}[h]
\caption{Overview of notation and matrix shapes}
\begin{tabularx}{\textwidth}{@{}XX@{}}
\toprule
  \underline{Random variables:}  \\
  \enskip$\weight: d_{out} \times d_{in}$ & Weight matrix of a layer \\
  \enskip$\inducing: M_{out} \times M_{in}$ & Inducing weight matrix \\
  \enskip$\inducing_r: M_{out} \times d_{in}$ & Inducing row matrix \\
  \enskip$\inducing_c: d_{out} \times M_{in}$ & Inducing column matrix \\
  \\
  \underline{Parameters:}  \\
  \enskip$\covparam_r: M_{out} \times d_{out}, \covparam_c: M_{in} \times d_{in}$ &  Inducing covariance parameters\\
  \enskip$\precparam_r: M_{out} \times M_{out}, \precparam_c: M_{in} \times M_{in}$ &  Inducing precision parameters (we use diagonal matrices in our approach)\\
  \enskip$\lambda^2$ & Conditional rescaling factor ($<1$ reduces variance of weight distribution)\\
  \\
  \underline{Hyperparameters:}  \\
  \enskip$\sigma^2$ & Prior variance \\
  \enskip$\sigma^2_{max}$ & Maximum approximate posterior variance \\
  \enskip$\lambda^2_{max}$ & Maximum conditional rescaling \\
  \\
  \underline{Other variables:}  \\
  \enskip$\fullcov_r: (d_{out} + M_{out}) \times (d_{out} + M_{out})$ & Joint row covariance of $\weight$ and $\inducing$ \\
  \enskip$\fullcov_c: (d_{in} + M_{in}) \times (d_{in} + M_{in})$ & Joint column covariance of $\weight$ and $\inducing$ \\
  \enskip$\inducingchol_r, \inducingchol_c$ & Lower Cholesky factor of the joint row/column covariance\\
  \enskip$\inducingcov_r: M_{out} \times M_{out}, \inducingcov_c: M_{in} \times M_{in}$ & Marginal row/column inducing covariance\\
\bottomrule
\end{tabularx}
\end{table}

\section{Derivations of the auxiliary variational objective}
\label{app:auxiliary_variational_objective_derivations}
When the variational distribution is constructed as a mixture distribution $q(\params) = \int q(\params | \auxvars) q(\auxvars) d\auxvars$, the original variational lower-bound becomes intractable: 
\begin{equation}
    \log p(\data) \geq \mathcal{L}(q(\params)) = \mathbb{E}_{q(\params)}[\log p(\data | \params)] + \mathbb{E}_{q(\params)}\left[ \log \frac{p(\params)}{\int q(\params | \auxvars) q(\auxvars) d\auxvars} \right].
\end{equation}
However, notice that we can also rewrite the ``marginal'' $q(\params)$ using Bayes' rule: $q(\params) = \frac{q(\params | \auxvars) q(\auxvars)}{q(\auxvars | \params)}$,
meaning that the variational lower-bound can be re-formulated as
\begin{equation}
    \mathcal{L}(q(\params)) = \mathbb{E}_{q(\params, \auxvars)}[\log p(\data | \params)] + \mathbb{E}_{q(\params, \auxvars)}\left[ \log \frac{p(\params) q(\auxvars | \params)}{q(\params | \auxvars) q(\auxvars)} \right].
\end{equation}
For many flexible mixture distributions, $q(\auxvars | \params)$ remains intractable. Fortunately, notice that for \emph{any} distribution $r(\auxvars | \params)$ with the same support as $q(\auxvars | \params)$, we have:
\begin{equation}
    \mathbb{E}_{q(\params, \auxvars)}\left[ \log \frac{p(\params) q(\auxvars | \params)}{q(\params | \auxvars) q(\auxvars)} \right] = \mathbb{E}_{q(\params, \auxvars)}\left[ \log \frac{p(\params) r(\auxvars | \params)}{q(\params | \auxvars) q(\auxvars)} \right] + \mathbb{E}_{q(\params, \auxvars)}\left[ \log \frac{q(\auxvars | \params)}{r(\auxvars | \params)} \right],
\end{equation}
and more importantly, the second term on the RHS of the above equation satisfies
\begin{equation}
     \mathbb{E}_{q(\params, \auxvars)}\left[ \log \frac{q(\auxvars | \params)}{r(\auxvars | \params)} \right] = \mathbb{E}_{q(\params)}[\kl[q(\auxvars | \params) || r(\auxvars | \params)]] \geq 0.
\end{equation}
This means we can remove this KL term and construct a lower-bound to the variational lower-bound:
\begin{equation}
    \log{} p(\data) \geq \mathcal{L}(q(\params)) \geq \mathbb{E}_{q(\params, \auxvars)}[\log p(\data | \params)] + \mathbb{E}_{q(\params, \auxvars)} \left[ \log \frac{p(\params) r(\auxvars | \params)}{q(\params | \auxvars)q(\auxvars)} \right] := \mathcal{L}(q(\params, \auxvars)),
\end{equation}
which corresponds to the auxiliary variational lower-bound \cref{eq:auxiliary_variational_lowerbound} presented in the main text. This auxiliary bound can be improved by optimising $r(\auxvars | \params)$ towards better approximating $q(\auxvars | \params)$, and it recovers the original variational lower-bound iff.~$r(\auxvars | \params) = q(\auxvars | \params)$. Still we emphasise that the auxiliary bound is valid for \emph{any} $r(\auxvars | \params)$ with the same support as $q(\auxvars | \params)$, which enables our design presented in the main text to improve memory efficiency.

\section{An introduction to SVGP}
\label{app:gp_intro}
This section provides a brief introduction to sparse variational approximation for \acrfull{svgp}. We use regression as a running example, but the principles of \acrshort{svgp} also apply to other supervised learning tasks such as classification. Readers are also referred to e.g.~\citet{leibfried2020tutorial} for a modern tutorial.

Assume we have a regression dataset $\mathcal{D} = \{ \X, \Y \}$ where $\X = [\x_1, ..., \x_N]$ and $\Y = [\bm{y}_1, ..., \bm{y}_N]$. In GP regression we build the following probabilistic model to address this regression task:
\begin{equation}
    \bm{y}_n = f(\x_n) + \bm{\epsilon}_n, \quad \bm{\epsilon}_n \sim \mathcal{N}(\bm{0}, \sigma^2 \mathbf{I}), \quad f(\cdot) \sim \mathcal{GP}(0, \mathcal{K}(\cdot, \cdot)).
\end{equation}
Here we put on the regression function $f(\cdot)$ a \acrshort{gp} prior with zero mean function and covariance function defined by kernel $\mathcal{K}(\cdot, \cdot)$. In practice we can only evaluate the function $f(\cdot)$ on finite number of inputs, but fortunately by construction, \acrshort{gp} allows sampling of function values from a joint Gaussian distribution. In detail, we can sample from the following Gaussian distribution to get the function value samples $\f = [f(\x_1), ..., f(\x_N)]$ given the input locations $\X$:
\begin{equation}
    \f \sim \mathcal{N}(\bm{0}, \K_{\X\X}), \quad \K_{\X\X}(m, n) = \mathcal{K}(\x_m, \x_n).
\end{equation}
Therefore we can rewrite the probabilistic model in finite dimension as
\begin{equation}
    p(\Y | \X, f(\cdot)) = p(\Y | \f) = \prod_{n=1}^N \mathcal{N}(\bm{y}_n; f(\x_n), \sigma^2 \mathbf{I}), \quad p(\f | \X) = \mathcal{N}(\f; \bm{0}, \K_{\X\X}).
\end{equation}
Note that the \acrshort{gp} prior can be extended to a larger set of inputs $\X \cup \X^*$ as $p(\f, \f^* | \X, \X^*) = \mathcal{N}([\f, \f^*]; \bm{0}, \K_{[\X, \X^*], [\X, \X^*]})$, where $\f^*$ denotes the function values given the inputs $\X^*$, and 
\begin{equation*}
\begin{aligned}
\K_{[\X, \X^*], [\X, \X^*]} &= \pmat{\K_{\X\X} & \K_{\X \X^*} \\ \K_{\X^* \X} & \K_{\X^* \X^*} } .
\end{aligned}
\end{equation*}
Importantly, this definition leaves the marginal prior unchanged ($\int p(\f, \f^* | \X, \X^*) d\f^* = p(\f | \X)$), and the conditional distribution $p(\f | \f^*, \X, \X^*)$ is also Gaussian. This means predictive inference can be done in the following way: given test inputs $\X^*$, the posterior predictive for $\f^*$ is
\begin{equation}
p(\Y | \X, \X^*) = \int p(\Y | \f) p(\f| \X) p(\f^* | \f, \X, \X^*) d\f d\f^* = \int p(\Y | \f) p(\f| \X) d\f = p(\Y | \X),
\end{equation}
\begin{equation}
\begin{aligned}
    p(\f^* | \X, \Y, \X^*) &= \int \frac{p(\Y | \f) p(\f, \f^* | \X, \X^*)}{p(\Y | \X, \X^*)} d\f \\
    &= \int \frac{p(\Y | \f) p(\f| \X) p(\f^* | \f, \X, \X^*)}{p(\Y | \X)} d\f \\
    &= \int \underbrace{p(\f | \X, \Y) p(\f^* | \f, \X, \X^*)}_{=p(\f, \f^* | \X, \Y, \X^*)} d\f.
\end{aligned}
\end{equation}
Unfortunately the exact posterior $p(\f | \X, \Y)$ is intractable even for \acrshort{gp} regression. Although in such case $p(\f | \X, \Y)$ is Gaussian, evaluating this posterior requires inverting/decomposing an $N \times N$ covariance matrix which has time complexity $\mathcal{O}(N^3)$. Also the storage cost of $\mathcal{O}(N^2)$ for the posterior covariance is prohibitively expensive when $N$ is large.

To address the intractability issue, we seek to define an \emph{approximate posterior} $q(f(\cdot))$, so that we can evaluate it on $\X \cup \X^*$ and approximate the posterior as $p(\f, \f^* | \X, \Y, \X^*) \approx q(\f, \f^*)$. \acrshort{svgp} \citep{snelson2006sparse,titsias2009variational} defines such approximate posterior by introducing \emph{inducing} inputs and outputs. Again this is done by noticing that we can extend the \acrshort{gp} prior to an even larger set of inputs $\X \cup \X^* \cup \mathbf{Z}$, where $\mathbf{Z} = [\bm{z}_1, ..., \bm{z}_M]$ are called \emph{inducing inputs}, and the corresponding function values $\mathbf{u} = [f(\bm{z}_1), ..., f(\bm{z}_M)]$ are named \emph{inducing outputs}:
\begin{equation}
\begin{aligned}
    p(\f, \f^*, \mathbf{u} | \X, \X^*, \mathbf{Z}) &= \mathcal{N}([\f, \f^*, \bm{u}]; \bm{0}, \K_{[\X, \X^*, \mathbf{Z}], [\X, \X^*, \mathbf{Z}]}) \\
    &= p(\mathbf{u}|\mathbf{Z})p(\f, \f^* | \mathbf{u}, \X, \X^*, \mathbf{Z}).
\end{aligned}
\end{equation}
Importantly, marginal consistency still holds for any $\mathbf{Z}$: $\int p(\f, \f^*, \mathbf{u} | \X, \X^*, \mathbf{Z}) d\mathbf{u} = p(\f, \f^* | \X, \X^*)$ (c.f.~eq.(\ref{eq:marginalisation_constraint}) in the main text). Furthermore, the conditional distribution $p(\f, \f^* | \mathbf{u}, \X, \X^*, \mathbf{Z})$ is a Gaussian distribution. Observing these, the \acrshort{svgp} approach defines the approximate posterior as
\begin{equation}
    q(\f, \f^*, \mathbf{u}) = p(\f, \f^* | \X, \X^*, \mathbf{Z}^*) q(\mathbf{u}), \quad q(\f, \f^*) = \int q(\f, \f^*, \mathbf{u}) d\mathbf{u},
\end{equation}
and minimises an upper-bound of the KL divergence to find the optimal $q(\f, \f^*)$:
\begin{equation}
\begin{aligned}
    \kl[q(\f, \f^*) || p(\f, \f^* | \X, \Y, \X^*)] &= \mathbb{E}_{q(\f, \f^*)}\left[ \log \frac{q(\f, \f^*) p(\Y | \X)}{p(\f, \f^*, \Y | \X, \X^*)} \right] \\
    = \log p(\Y | \X) + &\mathbb{E}_{q(\f, \f^*, \mathbf{u})}\left[ \log \frac{q(\f, \f^*, \mathbf{u})}{p(\Y | \f) p(\f, \f^*, \mathbf{u} | \X, \X^*, \mathbf{Z})} + \frac{p(\mathbf{u} | \f, \f^*, \X, \X^*, \mathbf{Z})}{q(\mathbf{u} | \f, \f^*)} \right] \\
    = \log p(\Y | \X) + &\mathbb{E}_{q(\f, \f^*, \mathbf{u})}\left[ \log \frac{ \cancel{p(\f, \f^*| \mathbf{u}, \X, \X^*, \mathbf{Z})} q(\mathbf{u})}{p(\Y | \f) \cancel{p(\f, \f^* | \mathbf{u}, \X, \X^*, \mathbf{Z})} p(\mathbf{u} | \mathbf{Z})}\right] \\ -& \mathbb{E}_{q(\f, \f^*)} \kl[q(\mathbf{u} | \f, \f^*) || p(\mathbf{u} | \f, \f^*, \X, \X^*, \mathbf{Z})] \\
    \leq \log p(\Y | \X) -& \lbrace \underbrace{ \mathbf{E}_{q(\f)}[\log p(\Y|\f)] - \kl[q(\mathbf{u}) || p(\mathbf{u} | \mathbf{Z})] }_{:= \mathcal{L}(q(\f, \mathbf{u})} \rbrace. 
\end{aligned}
\end{equation}
As KL divergences are non-negative, the above derivations also means $\log p(\Y | \X) \geq \mathcal{L}(q(\f, \mathbf{u}))$, and we can optimise the parameters in $q$ to tighten the lower-bound. These variational parameters include the distributional parameters of $q(\mathbf{u})$ (e.g.~mean and covariance if $q(\mathbf{u})$ is Gaussian), as well as the inducing inputs $\mathbf{Z}$, since the variational lower-bound $\mathcal{L}(q(\f, \mathbf{u}))$ is valid for any settings of $\mathbf{Z}$. 
The lower-bound requires evaluating 
$$q(\f) = \int p(\f, \f^* | \mathbf{u}, \X, \X^*, \mathbf{Z})q(\mathbf{u}) d\f^* d\mathbf{u} = \int p(\f | \mathbf{u}, \X, \mathbf{Z})q(\mathbf{u}) d\mathbf{u},$$
which can be done efficiently when $M << N$, as evaluating the conditional Gaussian $p(\f | \mathbf{u}, \X, \mathbf{Z})$ has $\mathcal{O}(NM^2 + M^3)$ run-time cost. Similarly, once $q$ is optimised, in prediction time one can directly sample from $q(\f^*)$ by computing
$q(\f^*) = \int p(\f^* | \mathbf{u}, \X^*, \mathbf{Z})q(\mathbf{u}) d\mathbf{u}$, and by caching the inverse/decomposition of both the covariacne matrix of $q(\mathbf{u})$ and $\K_{\mathbf{Z} \mathbf{Z}}$, predictive inference can be approximated efficiently.

Using shorthand notations by dropping $\mathbf{Z}$, e.g.~$p(\mathbf{u}) = p(\mathbf{u}|\mathbf{Z})$ and $p(\f | \X, \mathbf{u}) = p(\f | \mathbf{u}, \X, \mathbf{Z})$, returns the desired results discussed in \cref{sec:background} of the main text, if we set $a = \mathbf{u}$ and $\theta = \f$. In fact from the above discussions, we see that $\theta$ in \acrshort{gp} inference is infinite dimensional: $\theta = f_{\neq \mathbf{u}}$, as we can extend the finite collection of function values to include both $\f$ and $\f^*$ for any $\X^* \neq \mathbf{Z}$. By tying the conditional distribution given $\mathbf{u}$ in both the \acrshort{gp} prior and the approximate posterior $q(f(\cdot))$, the posterior belief updates are ``compressed'' into $\mathbf{u}$ space, which is also reflected by the name ``sparse approximation'' of the approach.

\section{Derivations of the augmented (pseudo) prior}
\label{app:inducing_weight_derivations}

\subsection{Inducing auxiliary variables: multivariate Gaussian case}
\label{app:inducing_weight_vector_normal}
Suppose each weight matrix has an isotropic Gaussian prior with zero mean, i.e. $\vect(\weight) \sim \gauss(0, \sigma^2 \eye)$ where $\vect$ concatenates the columns of a matrix into a vector and $\sigma$ is the standard deviation.
Augmenting this Gaussian with an auxiliary variable $\inducing$ that also has a mean of zero and some covariance that we are free to parameterise, the joint distribution is
\begin{align*}
    \pmat{\vect(\weight)\\\vect(\inducing)} &\sim \gauss(0, \fullcov)
    \quad\text{with}\quad
    L = \pmat{\sigma\eye & 0\\\covparam & \precparam}\\
    \quad\text{s.t.}\quad
    \fullcov &= \inducingchol\inducingchol^\top = \pmat{\sigma^2\eye & \sigma \covparam^\top\\\sigma \covparam & \covparam\covparam^\top + \precparam^2}
\end{align*}
where $\precparam$ is a positive diagonal matrix and $\covparam$ a matrix with arbitrary entries.
Through defining the Cholesky decomposition of $\fullcov$ we ensure its positive definiteness.
By the usual rules of Gaussian marginalisation, the augmented model leaves the marginal prior on $\weight$ unchanged.
Further, we can analytically derive the conditional distribution on the weights given the inducing weights:
\begin{align}
    p(\vect(\weight)\given\vect(\inducing)) &= \gauss(\vmu_{\weight\given\inducing}, \fullcov_{\weight\given\inducing}), \\
    \vmu_{\weight\given\inducing} &= \sigma \covparam^\top \inducingprec \vect(U),
    \\
    \fullcov_{\weight\given\inducing} &= \sigma^2 (\eye - \covparam^\top \inducingprec\covparam), 
    \\
    \inducingcov &= \covparam\covparam^\top + \precparam^2. \nonumber
\end{align}
For inference, we now need to define an approximate posterior over the joint space $q(\weight, \inducing)$.
We will do so by factorising it as $q(\weight, \inducing) = q(\weight\given\inducing) q(\inducing)$.
Factorising the prior in the same way leads to the following KL term in the ELBO:
\begin{equation}
    \kl\left[q(\weight, \inducing) \vert\vert p(\weight, \inducing) \right]
    =
    \avg_{q(\inducing)}\left[ \kl\left[ q(\weight\given\inducing) \vert\vert p(\weight\given\inducing \right]\right]
    +
    \kl\left[ q(\inducing) \vert\vert p(\inducing) \right]
\end{equation}

\subsection{Inducing auxiliary variables: matrix normal case}
\label{app:inducing_weight_matrix_normal}
Now we introduce the inducing variables in matrix space, and, in addition to the inducing weight $\inducing$, we pad in two inducing matrices $\inducing_r$, $\inducing_c$, such that the full augmented prior is:
\begin{align}
    \pmat{\weight & \inducing_c\\\inducing_r & \inducing} &\sim p(\weight, \inducing_c, \inducing_r, \inducing) := \matgauss(0, \fullcov_r, \fullcov_c), \\
    \text{with}\quad
    \inducingchol_r &= \pmat{\sigma_r \eye & 0\\\covparam_r & \precparam_r}
    \quad\text{and}\quad
    \inducingchol_c = \pmat{\sigma_c \eye & 0\\\covparam_c & \precparam_c}\nonumber\\
    \text{s.t.}\quad
    \fullcov_r &= \inducingchol_r\inducingchol_r^\top = \pmat{\sigma_r^2\eye & \sigma_r \covparam_r^\top\\\sigma_r \covparam_r & \covparam_r\covparam_r^\top + \precparam_r^2}, \nonumber\\
    \text{and}\quad
    \fullcov_c &= \inducingchol_c\inducingchol_c^\top = \pmat{\sigma_c^2\eye & \sigma_c\covparam_c^\top\\\sigma_c\covparam_c & \covparam_c\covparam_c^\top + \precparam_c^2}. \nonumber
\end{align}
Matrix normal distributions have similar marginalisation and conditioning properties as multivariate Gaussians.
As such, the marginal both over some set of rows and some set of columns is still a matrix normal.
Hence, $p(\weight) = \matgauss(0, \sigma_r^2 \eye, \sigma_c^2 \eye)$, and by choosing $\sigma_r\sigma_c = \sigma$ this matrix normal distribution is equivalent to the multivariate normal $p(\vect(\weight)) = \gauss(0, \sigma^2\eye)$. Also $p(\inducing) = \matgauss(0, \inducingcov_r, \inducingcov_c)$, where again $\inducingcov_r = \covparam_r\covparam_r^\top + \precparam_r^2$ and $\inducingcov_c = \covparam_c\covparam_c^\top + \precparam_c^2$.
Similarly, the conditionals on some rows or columns are matrix normal distributed:
\begin{align}
    \inducing_c\given\inducing
    &\sim \matgauss(\sigma_r \covparam_r^\top\inducingprec_r\inducing, \sigma_r^2(\eye - \covparam_r^\top\inducingprec_r\covparam_r), \inducingcov_c),\nonumber\\
    \inducing_r \given \inducing
    &\sim \matgauss(\inducing\inducingprec_c \sigma_c\covparam_c, \inducingcov_r, \sigma_c^2(\eye - \covparam_c^\top \inducingprec_c \covparam_c)),\nonumber\\
    \weight \given \inducing_c
    &\sim \matgauss\left( \inducing_c \inducingcov_c^{-1} \sigma_c \covparam_c, \sigma_r^2 \eye, \sigma_c^{2} (\eye - \covparam_c^\top \inducingcov_c^{-1} \covparam_c ) \right),\nonumber\\
    \weight,\inducing_r\given\inducing_c,\inducing
    &\sim \matgauss(\pmat{\inducing_c\\\inducing}\inducingprec_c \sigma_c \covparam_c, \fullcov_r, \sigma_c^2(\eye - \covparam_c^\top\inducingprec_c\covparam_c)),\\
    \weight \given \inducing_r, \inducing_c, \inducing
    &\sim
    \matgauss(M_\weight, \sigma_r^2(\eye - \covparam_r^\top\inducingprec\covparam_r),\sigma_c^2(\eye - \covparam_c^\top\inducingprec_c\covparam_c)),\nonumber\\
    M_\weight &= \sigma (
    \covparam_r^\top\inducingprec_r\inducing_r + \inducing_c\inducingprec_c\covparam_c - \covparam_r^\top\inducingprec_r\inducing\inducingprec_c\covparam_c).\nonumber
\end{align}

\section{KL divergence for rescaled conditional weight distributions}
\label{app:kl_divergence_derivation}

For the conditional distribution on the weights, in the simplest case we set $q(\weight\given\inducing) = p(\weight\given\inducing)$, hence the KL divergence would be zero.
For the most general case of arbitrary Gaussian distributions with $q = \gauss(\vmu_q, \fullcov_q)$ and $p = \gauss(\vmu_p, \fullcov_p)$, the KL divergence is:
\begin{equation}
    \kl\left[q\vert\vert p\right]
    =
    \frac{1}{2} (\log \frac{\det\fullcov_p}{\det\fullcov_q} - d + \trace(\fullcov_p^{-1}\fullcov_q)
    + (\vmu_p - \vmu_q)^\top\fullcov_p^{-1}(\vmu_p - \vmu_q)),
\end{equation}
where $d$ is the number of elements of $\vmu$. 
As motivated, it is desirable to make $q(\weight | \inducing)$ similar to $p(\weight | \inducing)$. We consider a scalar rescaling of the covariance, i.e. for $p = \gauss(\vmu, \fullcov)$ we set $q = \gauss(\vmu, \lambda^2\fullcov)$.
This leads to the final term, which is the Mahalanobis distance between the means under $p$, cancelling out entirely and the log determinant and trace terms becoming a function of $\lambda$ only: with $d = \text{dim}(\vect(\weight))$,
\begin{align*}
    \kl\left[q\vert\vert p\right]
    &= \frac{1}{2} (\log \frac{\det\fullcov}{\det\lambda^2\fullcov} - d + \trace(\fullcov^{-1}\lambda^2\fullcov))\\
    &= \frac{1}{2} (\log \frac{\det\fullcov}{\lambda^{2d} \det\fullcov} - d + \trace(\lambda^2\eye))\\
    &= \frac{1}{2} (-2d\log \lambda - d + d\lambda^2)\\
    &= d (\frac{1}{2}\lambda^2 - \log \lambda - \frac{1}{2}).
\end{align*}

\section{The extended Matheron's rule to matrix normal distributions}
\label{app:extended_matheron}

The original Matheron's rule \citep{journel1978mining,hoffman1991constrained,doucet2010gaussian} for sampling conditional Gaussian variables states the following. If the joint multivariate Gaussian distribution is
\begin{align*}
    \pmat{\vect(\weight)\\\vect(\inducing)} &\sim p(\vect(\weight), \vect(\inducing)) := \gauss(\bm{0}, \fullcov),\\
    \fullcov &= \pmat{\fullcov_{\weight \weight} & \fullcov_{\weight \inducing} \\ \fullcov_{\inducing \weight} & \fullcov_{\inducing \inducing}},
\end{align*}
then, conditioned on $\inducing$, sampling $\weight \sim p(\vect(\weight), \vect(\inducing))$ can be done as
\begin{align*}
    \vect(\weight) &= \vect(\bar{\weight}) + \fullcov_{\weight \inducing} \fullcov_{\inducing \inducing}^{-1} (\vect(\inducing) - \vect(\bar{\inducing})),\\
    \vect(\bar{\weight}), \vect(\bar{\inducing}) &\sim \gauss(\bm{0}, \fullcov).
\end{align*}
Matheron's rule can provide significant speed-ups if $\vect(\inducing)$ has significantly smaller dimensions than that of $\vect(\weight)$, and the Cholesky decomposition of $\fullcov$ can be computed with low costs (e.g. due to the specific structure in $\fullcov$). Recall from the main text that the augmented prior is
\begin{equation}
    p(\vect(\weight), \vect(\inducing)) =
    \gauss \left(0, \pmat{\sigma_c^2 \eye \kron \sigma_r^2 \eye & \sigma_c\covparam_c^\top \kron \sigma_r\covparam_r^\top\\\sigma_c\covparam_c \kron \sigma_r\covparam_r & \inducingcov_c \kron \inducingcov_r} \right),
\end{equation}
and the corresponding conditional distribution is:
\begin{equation}
    p(\vect(\weight) \given \vect(\inducing))
    =
    \gauss(\sigma_c\sigma_r\vect(\covparam_r\inducingprec_r\inducing\inducingprec_c\covparam_c^\top), \sigma_c^2\sigma_r^2(\eye - \covparam_c^\top\inducingprec_c\covparam_c \kron \covparam_r^\top\inducingprec_r\covparam_r)).
\label{eq:conditional_distribution_vec}
\end{equation}
Therefore, while $\text{dim}(\vect(\inducing))$ is indeed significantly smaller than of $\text{dim}(\vect(\weight))$ by construction, the joint covariance matrix does not support fast Cholesky decompositions, meaning that Matheron's rule for efficient sampling does not directly apply here.

However, in the full augmented space, the joint distribution does have an efficient matrix normal form: $p(\weight, \inducing_c, \inducing_r, \inducing_c) = \matgauss(0, \fullcov_r, \fullcov_c)$. Furthermore, the row and column covariance matrices $\fullcov_r$ and $\fullcov_c$ are parameterised by their Cholesky decompositions, meaning that sampling from the joint distribution $p(\weight, \inducing_c, \inducing_r, \inducing)$ can be done in a fast way. Importantly, Cholesky decompositions for $p(\inducing)$'s row and column covariance matrices $\inducingcov_r$ and $\inducingcov_c$ can be computed in $\mathcal{O}(M_{out}^3)$ and $\mathcal{O}(M_{in}^3)$ time, respectively, which are much faster than the multi-variate Gaussian case that requires $\mathcal{O}(M_{in}^3 M_{out}^3)$ time. Observing these, we extend Matheron's rule to sample $p(\weight | \inducing)$ where $p(\weight, \inducing)$ is the marginal distribution of $p(\weight, \inducing_c, \inducing_r, \inducing_c) = \matgauss(0, \fullcov_r, \fullcov_c)$.

In detail, for drawing a sample from $p(\weight\given\inducing)$ we need to draw a sample from the joint $p(\weight, \inducing)$.
To do so, we sample from the augmented prior $\bar{\weight}, \bar{\inducing_c}, \bar{\inducing_r}, \bar{\inducing} \sim p(\bar{\weight}, \bar{\inducing_c}, \bar{\inducing_r}, \bar{\inducing}) = \matgauss(0, \fullcov_r, \fullcov_c)$, computed using the Cholesky decompositions of $\fullcov_r$ and $\fullcov_c$:
\begin{equation*}
    \pmat{\bar{\weight}&\bar{\inducing}_c\\\bar{\inducing}_r&\bar{\inducing}}
    =
    \pmat{\sigma_r\eye&0\\\covparam_r&\precparam_r}
    \pmat{\mEps_1&\mEps_2\\\mEps_3&\mEps_4}
    \pmat{\sigma_c\eye&\covparam_c^\top\\0&\precparam_c},
\end{equation*}
where $\mEps_1 \in \mathbb{R}^{d_{out}\times d_{in}}$, $\mEps_2 \in \mathbb{R}^{d_{out} \times M_{in}}$, $\mEps_3 \in \mathbb{R}^{M_{out} \times d_{in}}$, $\mEps_4 \in \mathbb{R}^{M_{out} \times M_{in}}$ are standard Gaussian noise samples, and $\bar{\weight} \in \mathbb{R}^{d_{out} \times d_{in}}$ and $\bar{\inducing} \in \mathbb{R}^{M_{out}\times M_{in}}$.
Then we construct the conditional sample $\weight \sim p(\weight | \inducing)$ as follows, similar to Matheron's rule in the multivariate Gaussian case:
\begin{equation}
     \weight = \bar{\weight} + \sigma_r \sigma_c \covparam_r^\top\inducingprec_r (U - \bar{U}) \inducingprec_c\covparam_c.
     \label{eq:matrix_matheron_rule}
\end{equation}
From the above equations we see that $\bar{\inducing_r}$ and $\bar{\inducing_c}$ do not contribute to the final $\weight$ sample. Therefore we do not need to compute $\bar{\inducing}_r$ and $\bar{\inducing}_c$, and we write the separate expressions for $\bar{\weight}$ and $\bar{\inducing}$ as:
\begin{equation}
    \bar{\weight} = \sigma_r \sigma_c \mEps_1, \quad
    \bar{\inducing} = \underbrace{\covparam_r \mEps_1 \covparam_c^\top}_{\bar{\inducing}_1} + \underbrace{\covparam_r \mEps_2 \precparam_c}_{\bar{\inducing}_2} + \underbrace{\precparam_r \mEps_3 \covparam_c^\top}_{\bar{\inducing}_3} + \underbrace{\precparam_r \mEps_4 \precparam_c}_{\bar{\inducing}_4}.
\label{eq:compute_wbar_ubar}
\end{equation}
Note that $\bar{\inducing}$ is a sum of four samples from matrix normal distributions. In particular, we have that:
\begin{equation*}
    \bar{\inducing}_2 \overset{d}{\sim} \matgauss(0, \covparam_r\covparam_r^\top, \precparam_c^2)
    \quad\text{and}\quad
    \bar{\inducing}_3 \overset{d}{\sim} \matgauss(0, \precparam_r^2, \covparam_c\covparam_c^\top).
\end{equation*}
Hence instead of sampling the ``long and thin'' Gaussian noise matrices $\mEps_2$ and $\mEps_3$, we can reduce variance by sampling standard Gaussian noise matrices $\tilde{E}_2, \tilde{E}_3 \in \mathbb{R}^{M_{out} \times M_{in}}$, and calculate $\bar{\inducing}$ as
\begin{equation}
    \bar{\inducing} = \covparam_r \mEps_1\covparam_c^\top + \hat{L}_r\tilde{E}_2\precparam_c + \precparam_r\tilde{E}_3 \hat{L}_c^\top + \precparam_r \mEps_4\precparam_c.
\end{equation}
This is enabled by calculating the Cholesky decompositions $\hat{\inducingchol}_r\hat{\inducingchol}_r^\top = \covparam_r\covparam_r^\top$ and $\hat{\inducingchol}_c\hat{\inducingchol}_c^\top = \covparam_c\covparam_c^\top$, which have $\mathcal{O}(M_{out}^3)$ and $\mathcal{O}(M_{in}^3)$ run-time costs, respectively.
As a reminder, the Cholesky factors are \emph{square} matrices, i.e. $\hat{\inducingchol}_r \in \mathbb{R}^{M_{out} \times M_{out}}$, $\hat{\inducingchol}_c \in \mathbb{R}^{M_{in} \times M_{in}}$). We name the approach the \emph{extended Matheron's rule} for sampling conditional Gaussians when the full joint has a matrix normal form.

To verify the proposed approach, we compute the mean and the variance of the random variable $\weight$ defined in \cref{eq:matrix_matheron_rule}, and check if they match the mean and variance of \cref{eq:conditional_distribution_vec}. First as $\bar{\weight}, \bar{\inducing}$ have zero mean, it is straightforward to verify that $\mathbb{E}[W] = \sigma_r \sigma_c \covparam_r^\top\inducingprec_r U \inducingprec_c\covparam_c$ which matches the mean of \cref{eq:conditional_distribution_vec}. For the variance of $\vect(\weight)$, it requires computing the following terms:

\begin{align}
    \mathbb{V}(\vect(\weight)) =&
    \mathbb{V}(\vect(\bar{\weight})) + \mathbb{V}(\vect(\sigma_r \sigma_c \covparam_r^\top\inducingprec_r \bar{U} \inducingprec_c\covparam_c))\nonumber\\
    &- 2 \text{Cov}[\vect(\weight), \vect(\sigma_r \sigma_c \covparam_r^\top\inducingprec_r \bar{U} \inducingprec_c\covparam_c)]\nonumber\\
    =:& \mA_1 + \mA_2 - 2 \mA_3.
\label{eq:variance_of_w_matheron}
\end{align}

First it can be shown that

\begin{align*}
    \mA_1 &= \sigma_r^2 \sigma_c^2 \eye \quad \text{since } \bar{\weight} \sim \matgauss(0, \sigma_r^2 \eye, \sigma_c^2 \eye), \\
    \mA_2 &= \sigma_r^2 \sigma_c^2 \covparam_c^\top\inducingprec_c\covparam_c \kron \covparam_r^\top\inducingprec_r\covparam_r\\
    \text{as} \quad \covparam_r^\top\inducingprec_r \bar{U} \inducingprec_c\covparam_c &\sim \matgauss(0, \covparam_r^\top\inducingprec_r \covparam_r, \covparam_c^\top\inducingprec_c\covparam_c).
\end{align*}

For the correlation term $\mA_3$, we notice that $\bar{\weight}$ and $\bar{\inducing}$ only share the noise matrix $\mEps_1$ in the joint sampling procedure \cref{eq:compute_wbar_ubar}. This also means 

\begin{align*}
\mA_3 &= \sigma_r^2\sigma_c^2 \text{Cov}[\vect(\mEps_1), \vect(\covparam_r^\top\inducingprec_r \covparam_r \mEps_1 \covparam_c^\top\inducingprec_c\covparam_c)]\\
&= \sigma_r^2 \sigma_c^2 \covparam_c^\top\inducingprec_c\covparam_c \kron \covparam_r^\top\inducingprec_r\covparam_r.
\end{align*}

Plugging in $\mA_1, \mA_2, \mA_3$ into \cref{eq:variance_of_w_matheron} verifies that $\mathbb{V}(\vect(\weight))$ matches the variance of the conditional distribution $p(\vect(\weight) | \vect(\inducing))$, showing that the proposed extended Matheron's rule indeed draws samples from the conditional distribution.

As for sampling $\weight$ from $q(\weight | \inducing)$, since it has the same mean but a rescaled covariance as compared with $p(\weight | \inducing)$, we can compute the samples by adapting the extend Matheron's rule as follows. Notice that the mean of $\weight$ in \cref{eq:matrix_matheron_rule} is $\mathbb{E}[\weight | \inducing] = \sigma_r \sigma_c \covparam_r^\top\inducingprec_r U \inducingprec_c\covparam_c$, therefore by rearranging terms, \cref{eq:matrix_matheron_rule} can be re-written as
\begin{align*}
     \weight =& \covparam_r^\top\inducingprec_r \inducing \inducingprec_c\covparam_c + [ \bar{\weight} - \sigma_r \sigma_c \covparam_r^\top\inducingprec_r \bar{\inducing} \inducingprec_c\covparam_c]\\
     :=& \text{mean} + \text{noise}.
\end{align*}
So sampling from $q(\weight | \inducing)$ can be done by rescaling the noise term in the above equation with the scale parameter $\lambda$. In summary, the extended Matheron's rule for sampling $q(\weight | \inducing)$ is as follows:
\begin{equation}
\begin{split}
     \weight &= \lambda \bar{\weight} + \sigma_r \sigma_c \covparam_r^\top\inducingprec_r (U - \lambda \bar{\inducing}) \inducingprec_c\covparam_c, \\
     \bar{\weight}, \bar{\inducing} &\sim p(\bar{\weight}, \bar{\inducing_c}, \bar{\inducing_r}, \bar{\inducing}).
\end{split}
\end{equation}
Plugging in $\sigma_r\sigma_c = \sigma$ here returns the conditional sampling rule \cref{eq:matheron_rule_q_conditional} in the main text.

\section{Function-space view of inducing weights}
\label{app:function_space_inducing_weight}

Here we present the detailed derivations of \cref{sec:function_space_view}. 
Assume a neural network layer with weight $\weight$ computes the following transformation of the input $\X = [\x_1, ..., \x_N], \x_i \in \mathbb{R}^{d_{in} \times 1}$:
\begin{equation*}
    \F = \weight \X, \ \BH = g(\F), \quad \weight \in \mathbb{R}^{d_{out} \times d_{in}}, \X \in \mathbb{R}^{d_{in} \times N},
\end{equation*}
where $g(\cdot)$ is the non-linearity.
Therefore the Gaussian prior $p(\weight) = \gauss(0, \sigma^2 \eye)$ induces a Gaussian distribution on the linear transformation output $\F$, in fact each of the rows in $\F = [\f_1, ..., \f_{d_{out}}]^{\top}, \f_i \in \mathbb{R}^{N \times 1}$ has a Gaussian process form with linear kernel:
\begin{equation}
    \f_i | \X \sim \mathcal{GP}(\bm{0}, \K_{\X\X}), \quad \K_{\X\X}(m, n) = \sigma^2 \x_m^{\top} \x_n.
\end{equation}
Performing inference on $\F$ directly has $\mathcal{O}(N^3 + d_{out}N^2)$ cost, so a sparse approximation is needed.
Slightly different from the usual approach, we introduce ``scaled noisy inducing outputs'' $\inducing_c = [\bu_1^c, ..., \bu^c_{d_{out}}]^\top \in \mathbb{R}^{d_{out} \times M_{in}}$ in the following way, using shared inducing inputs $\covparam_c^\top \in \mathbb{R}^{d_{in} \times M_{in}}$:
\begin{equation*}
\begin{aligned}
p(\f_i, \hat{\bu}_i | \X) &= \mathcal{GP} \left( \bm{0}, \pmat{\K_{\X\X} & \K_{\X \covparam_c} \\ \K_{\covparam_c \X} & \K_{\covparam_c \covparam_c} } \right), \\
p(\bu^c_i | \hat{\bu}_i) &= \gauss \left( \frac{\hat{\bu}_i}{\sigma_c}, \sigma_r^2 \precparam_c^2 \right),
\end{aligned}
\end{equation*}
with $\K_{\covparam_c \X} = \sigma^2 \covparam_c \X$ and $\K_{\covparam_c \covparam_c} = \sigma^2 \covparam_c \covparam_c^\top$.
By marginalising out the ``noiseless inducing outputs'' $\hat{\bu}_i$, we have the joint distribution $p(\f_i, \bu_i)$ as
\begin{equation*}
\begin{aligned}
    p(\bu^c_i) &= \gauss(\bm{0}, \sigma_r^2 \inducingcov_c), \ \inducingcov_c = \covparam_c \covparam_c^\top + \precparam_c^2, \\
    p(\f_i | \X, \bu^c_i) &= \gauss(\sigma_c \sigma^{-2} \K_{\X \covparam_c} \inducingcov_c^{-1} \bu^c_{i}, \K_{\X\X} - \sigma^{-2} \K_{\X \covparam_c} \inducingcov_c^{-1} \K_{ \covparam_c \X}).
\end{aligned}
\end{equation*}
Collecting all the random variables in matrix forms, this leads to
\begin{align}
    p(\inducing_c) &= \matgauss(\bm{0}, \sigma_r^2 \eye, \inducingcov_c), \nonumber \\
    p(\F | \X, \inducing_c) &= \matgauss(\sigma_c \sigma^{-2} \inducing_c \inducingcov_c^{-1} \K_{\covparam_c \X}, \sigma_r^2 \eye, \sigma_r^{-2}(\K_{\X\X} - \sigma^{-2} \K_{\X \covparam_c} \inducingcov_c^{-1} \K_{ \covparam_c \X} ) ) \\
    &= \matgauss(\inducing_c \inducingcov_c^{-1} \sigma_c \covparam_c \X, \sigma_r^2 \eye, \X^\top \sigma_c^{2} (\eye - \covparam_c^\top \inducingcov_c^{-1} \covparam_c ) \X ). \nonumber
\end{align}
Also recall from conditioning rules of matrix normal distributions, we have that 
\begin{equation*}
    p(\weight | \inducing_c) = \matgauss\left( \inducing_c \inducingcov_c^{-1} \sigma_c \covparam_c, \sigma_r^2 \eye, \sigma_c^{2} (\eye - \covparam_c^\top \inducingcov_c^{-1} \covparam_c ) \right).
\end{equation*}
Since for $\weight \sim \matgauss(\mathbf{M}, \fullcov_1, \fullcov_2)$ we have $\weight \X \overset{d}{\sim} \matgauss(\mathbf{M}\X, \fullcov_1, \X^\top \fullcov_2 \X)$, this immediately shows that $p(\F | \X, \inducing_c)$ is the push-forward distribution of $p(\weight | \inducing_c)$ for the operation $\F = \weight \X$. In other words:
\begin{equation*}
    \F \sim p(\F | \X, \inducing_c) \quad \Leftrightarrow \quad \weight \sim p(\weight | \inducing_c), \ \F = \weight \X.
\end{equation*}
This confirms the interpretation of $\inducing_c$ as "scaled noisy inducing outputs" that lie in the same space as $\F$. Notice that in the main text we provide a pictorial visualisation of $\inducing_c$ by selecting $\sigma_c = 1$. As the inducing weights $\inducing$ are the focus of our analysis here, we conclude that this specific choice of $\sigma_c$ is without loss of generality.

So far the $\inducing_c$ variables assist the posterior inference to capture correlations across functions values of different inputs. Up to now the function values remain independent across output dimensions, which is also reflected by the diagonal row covariance matrices in the above matrix normal distributions. As in neural networks the output dimension can be fairly large (e.g.~$d_{out} = 1000$), to further improve memory efficiency, we proceed to project the \emph{column vectors} of $\inducing_c$ to an $M_{out}$ dimensional space with $M_{out} << d_{out}$. This dimension reduction step is done with a generative approach, similar to probabilistic PCA \citep{tipping1999probabilistic}:
\begin{equation}
\begin{split}
    \inducing &\sim \matgauss(0, \inducingcov_r, \inducingcov_c), \\
    \inducing_c | \inducing &\sim \matgauss(\sigma_r \covparam_r^\top\inducingprec_r\inducing, \sigma_r^2(\eye - \covparam_r^\top\inducingprec_r\covparam_r), \inducingcov_c).
\end{split}
\end{equation}
Note that the column covariance matrices in the above two matrix normal distributions are the same, and the conditional sampling procedure is done by a linear transformation of the columns in $\inducing$ plus noise terms. Again from the marginalisation and conditioning rules of matrix normal distributions, we have that the full joint distribution \cref{eq:inducing_full_joint}, after proper marginalisation and conditioning, returns
\begin{align*}
    p(\inducing) &= \matgauss(0, \inducingcov_r, \inducingcov_c), \\
    p(\inducing_c | \inducing) &= \matgauss(\sigma_r \covparam_r^\top\inducingprec_r\inducing, \sigma_r^2(\eye - \covparam_r^\top\inducingprec_r\covparam_r), \inducingcov_c).
\end{align*}
This means $\inducing$ can be viewed as ``projected noisy inducing points'' for the \acrshortpl{gp} $p(\F)$, whose corresponding ``inducing inputs'' are row vectors in $\covparam_c$.
Similarly, column vectors in $\inducing_r \X$ can be viewed as the noisy projections of the column vectors in $\F$, in other words $\inducing_r$ can also be viewed as ``neural network weights'' connecting the data inputs $\X$ to the projected output space that $\inducing$ lives in.

As for the variational objective, since $q(\weight | \inducing)$ and $p(\weight | \inducing)$ only differ in the scale of the covariance matrices, it is straightforward to show that the push-forward distribution $q(\weight | \inducing) \rightarrow q(\F |\X, \inducing)$ has the same mean as $p(\F |\X, \inducing)$, but with a different covariance matrix that scales $p(\F | \X, \inducing)$'s covariance matrix by $\lambda^2$. As the operation $\F = \weight \X$ maps $\weight \in \mathbb{R}^{d_{out} \times d_{in}}$ to $\F \in \mathbb{R}^{d_{out} \times N}$, this means the conditional KL is scaled up/down, depending on whether $N \geq d_{in}$ or not:
\begin{align*}
    \kl[q(\F |\X, \inducing) || p(\F |\X, \inducing)] &= \frac{N}{d_{in}} R(\lambda), \\
    R(\lambda) &:= \kl[q(\weight |\inducing) || p(\weight | \inducing)].
\end{align*}
In summary, the push-forward distribution of $q(\allweights, \inducing_{1:L}) \rightarrow q(\F_{1:L}, \inducing_{1:L})$ is
\begin{equation*}
    q(\F_{1:L}, \inducing_{1:L}) = \textstyle\prod\nolimits_{l=1}^L q(\F_{l} | \F_{l-1}, \inducing_l) q(\inducing_l), \quad \F_0 := \X,
\end{equation*}
and the corresponding variational lower-bound for $q(\F_{1:L}, \inducing_{1:L})$ becomes (with $\data = (\X, \Y)$)
\begin{equation}\label{eq:inducing_elbo_function_space}
    \mathcal{L}(q(\F_{1:L}, \inducing_{1:L})) = \mathbb{E}_{q(\F_{1:L})}[\log p(\Y | \F_{L})] - \textstyle\sum\nolimits_{l=1}^L \left( \frac{N}{d^l_{in}} R(\lambda_l) + \kl[q(\inducing_l) || p(\inducing_l)] \right),
\end{equation}
with $d^l_{in}$ the input dimension of layer $l$. 

Note that 
\begin{equation}
    \mathbb{E}_{q(\F_{1:L})}[\log p(\Y | \F_{L})] = \mathbb{E}_{q(\allweights)}[\log p(\Y | \X, \allweights)]
    = \mathbb{E}_{q(\allweights)}[\log p(\data | \allweights)].
\end{equation}
Comparing equations \cref{eq:elbo_inducing_weight_reduced,eq:inducing_elbo_function_space}, we see that the only difference between weight-space and function-space variational objectives comes in the scale of the conditional KL term. Though not investigated in the experiments, we conjecture that it could bring potential advantage to optimise the following variational lower-bound:
\begin{align}
    \tilde{\mathcal{L}}(q(\F_{1:L}, \inducing_{1:L})) =& \mathbb{E}_{q(\F_{1:L})}[\log p(\Y | \F_{L})] - \textstyle\sum\nolimits_{l=1}^L \left( \beta_l R(\lambda_l) + \kl[q(\inducing_l) || p(\inducing_l)] \right), \\
\beta_l =& \min(1, \frac{N}{d^l_{in}}).\nonumber
\end{align}
The intuition is that, as uncertainty is expected to be lower when $N \geq d_{in}$, it makes sense to use $\beta = 1 \leq N / d_{in}$ to reduce the regularisation effect introduced by the KL term. In other words, this allows the variational posterior to focus more on fitting the data, and in this ``large-data'' regime over-fitting is less likely to appear. On the other hand, function-space inference approaches (such as \acrshortpl{gp}) often return better uncertainty estimates when trained on small data ($N < d_{in}$). So choosing $\beta = N / d_{in} < 1$ in this case would switch to function-space inference and thereby improving uncertainty quality potentially. In the CIFAR experiments, the usage of convolutional filters leads to the fact that $N \geq d_{in}^l$ for all ResNet layers. Therefore in those experiments $\beta_l = 1$ for all layers, which effectively falls back to the weight-space objective \cref{eq:elbo_inducing_weight_reduced}.

\section{Whitening and hierarchical inducing variables}
\label{app:hierarchical_inducing}

The inducing weights $\allinducing$ further allow for introducing a single inducing weight matrix $\inducing$ that is shared across the network. By doing so, correlations of weights between layers in the approximate posterior are introduced. The inducing weights are then sampled jointly conditioned on the global inducing weights. This requires that all inducing weight matrices are of the same size along at least one axis, such that they can be concatenated along the other one.

The easiest way of introducing a global inducing weight matrix is by proceeding similarly to the introduction of the per-layer inducing weights.
As a pre-requisite, we need to work with ``whitened'' inducing weights, i.e. set the covariance of the marginal $p(\inducing_l)$ to the identity and pre-multiply the covariance block between $\weight_l$ and $\inducing_l$ with the inverse Cholesky of $\inducingcov_l$.
In this whitened model, the full augmented prior per-layer is:
\begin{align}
    \pmat{\weight&\inducing_c\\\inducing_r&\inducing}
    &\sim p(\weight, \inducing_c, \inducing_r, \inducing) :=
    \matgauss(0, \whitenedfullcov_r, \whitenedfullcov_r), \\
    \text{with}\quad
    \whitenedchol_r &= \pmat{\sigma_r\eye&0\\\inducingchol_r^{-1}\covparam_r&\inducingchol_r^{-1}\precparam_r}
    \\
    \text{s.t.}\quad \whitenedfullcov_r &= \whitenedchol_r\whitenedchol_r^\top = \pmat{\sigma_r^2\eye & \sigma_r\covparam_r^\top\inducingchol_r^{-\top}\\\sigma_r\inducingchol_r^{-1}\covparam_r&\eye} \nonumber \\
    \text{and}\quad
    \whitenedchol_c &= \pmat{\sigma_c\eye&0\\\inducingchol_c^{-1}\covparam_c&\inducingchol_c^{-1}\precparam_c} \\
    \text{s.t.}\quad
    \whitenedfullcov_c &= \whitenedchol_c\whitenedchol_c^\top = \pmat{\sigma_c^2\eye & \sigma_c\covparam_c^\top\inducingchol_c^{-\top}\\\sigma_c\inducingchol_c^{-1}\covparam_c&\eye}.\nonumber
\end{align}
One can verify that this whitened model leads to the same conditional distribution $p(\weight | \inducing)$ as presented in the main text.
After whitening, for each $\inducing_l$ we have that $p(\vect(\inducing_l)) = \gauss(0, \eye)$, therefore we can also write their joint distribution as $p(\vect(\allinducing)) = \gauss(0, \eye)$.
In order to construct a matrix normal prior $p(\allinducing) = \matgauss(0, \eye, \eye)$, the inducing weight matrices $\allinducing$ needs to be stacked either along the rows or columns, requiring the other dimension to be matching across all layers.
Then, As the covariance is the identity with $\sigma = \sigma_r = \sigma_c = 1$, we can augment $p(\inducing_{1:L})$ in the exact same way as we previously augmented the prior $p(\weight_l)$ with $\inducing_l$.

\section{Open-source code}
\label{app:open_source_code}
We open-source our approach as a PyTorch wrapper \texttt{bayesianize}:\\ \url{https://github.com/microsoft/bayesianize}

\texttt{bayesianize} is a lightweight Bayesian neural network (BNN) wrapper, and the goal is to allow for easy conversion of neural networks in existing scripts to BNNs with minimal changes to the code.
Currently the wrapper supports the following uncertainty estimation methods for feed-forward neural networks and convolutional neural networks:
\begin{itemize}
    \item Mean-field variational inference (MFVI) with fully factorised Gaussian (FFG) approximation, i.e.~FFG-$\weight$ in abbreviation.
    \item Variational inference with full-covariance Gaussian approximation (FCG-$\weight$).
    \item Inducing weight approaches, including FFG-$\inducing$, FCG-$\inducing$ and Ensemble-$\inducing$.
\end{itemize}

The main workhorse of our library is the \texttt{bayesianize\_} function, which turns deterministic \texttt{nn.Linear} and \texttt{nn.Conv} layers into their Bayesian counterparts. For example, to construct a Bayesian ResNet-18 that uses the variational inducing weight method, run:
\begin{minted}{python}
import bnn
net = torchvision.models.resnet18()
bnn.bayesianize_(net, inference="inducing", inducing_rows=64, inducing_cols=64)
\end{minted}
In the above code, \texttt{inducing\_rows} corresponds to $M_{out}$ and \texttt{inducing\_cols} corresponds to $M_{in}$. In other words, they specify the dimensions of $\inducing \in \mathbb{R}^{M_{out} \times M_{in}}$.
Then the converted BNN can be trained in almost identical way as one would train a deterministic net:
\begin{minted}{python}
yhat = net(x_train)
nll = F.cross_entropy(yhat, y_train)
kl = sum(m.kl_divergence() for m in net.modules()
         if hasattr(m, "kl_divergence"))
loss = nll + kl / dataset_size
loss.backward()
optim.step()
\end{minted}
Note that while the call to the forward method of the \texttt{net} looks the same, it is no longer deterministic because the weights are sampled, so subsequent calls will lead to different predictions. Therefore, when testing, an average of multiple predictions is needed. For example, in BNN classification:
\begin{minted}{python}
net.eval()
with torch.no_grad():
    logits = torch.stack([net(x_test) for _ in range(num_samples)])
    probs = logits.softmax(-1).mean(0)
\end{minted}
In the above code, \texttt{probs} computes
$$ p(\bm{y} | \x, \mathcal{D}) \approx \frac{1}{K} \sum_{k=1}^K p(\bm{y} | \x, \allweights^{(k)}), \quad \allweights^{(k)} \sim q,$$
where $K$ is the number of Monte Carlo samples \texttt{num\_samples}.

\texttt{bayesianize} also supports using different methods or arguments for different layers, by passing in a dictionary for the inference argument. This way we can, for example, take a \emph{pre-trained} ResNet and only perform (approximate) Bayesian inference over the weights of the final, linear layer:
\begin{minted}{python}
net = ... # load pre-trained network with net.fc as the last layer
bnn.bayesianize_(net, inference={
    net.fc: {"inference": "fcg"}
    })
optim = torch.optim.Adam(net.fc.parameters(), 1e-3)
# then train the last Bayesian layer accordingly
...
\end{minted}
For more possible ways of configuring the BNN settings, see example config \texttt{json} files in the open-source repository.

\section{Experimental details}
\label{app:experiment_details}

\subsection{Regression experiments}
\label{app:regression_details}

Following \citep{foong2019between}, we sample $50$ inputs each from $\mathcal{U}\left[-1, -0.7\right]$ and $\mathcal{U}\left[0.5, 1\right]$ as inputs and targets $y \sim \gauss(\cos(0.4x + 0.8), 0.01)$. As a prior we use a zero-mean Gaussian with standard deviation $\frac{4}{\sqrt{d_{in}}}$ for the weights and biases of each layer. Our network architecture has a single hidden layer of $50$ units and uses a $\tanh$-nonlinearity.
All three variational methods are optimised using Adam \citep{kingma2014adam} for $20,000$ updates with an initial learning rate of $10^{-3}$.
We average over $32$ MC samples from the approximate posterior for every update.
For Ensemble-U and FCG-U we decay the learning rate by a factor of $0.1$ after $10,000$ updates and the size of the inducing weight matrix is $2\times 25$ for the input layer (accounting for the bias) and $25 \times 1$ for the output layer. Ensemble-U uses an ensemble size of $8$.

For NUTS we use the implementation provided in Pyro \citep{bingham2019pyro}.
We draw a total of $25,000$ samples, discarding the first $5000$ as burn-in and using $1000$ randomly selected ones for prediction.

\subsection{Classification experiments}

We base our implementation on the Resnet-18 class in torchvision \citep{paszke2017automatic}, replacing the input convolutional layer with a $3\times3$ kernel size and removing the max-pooling layer.
We train the deterministic network on CIFAR-10 using Adam with a learning rate of $3\times 10^{-4}$ for 200 epochs.
On CIFAR-100 we found SGD with a momentum of $0.9$ and initial learning rate of $0.1$ decayed by a factor of $0.1$ after $60$, $120$ and $160$ epochs to lead to better accuracies.
The ensemble is formed of the five deterministic networks trained with different random seeds.

For \acrshort{ffg}-$\weight$ we initialised the mean parameters using the default initialisation in pytorch for the corresponding deterministic layers.
The initial standard deviations are set to $10^{-4}$.
We train using Adam for $200$ epochs on CIFAR-10 with a learning rate of $3\times10^{-4}$, and $300$ epochs on CIFAR-100 with an initial learning rate of $10^{-3}$, decaying by a factor of $0.1$ after $200$ epochs.
On both datasets we only use the \acrfull{nll} part of the variational lower bound for the first $100$ epochs as initialisation to the maximum likelihood parameter and then anneal the weight of the kl term linearly over the following $50$ epochs.
For the prior we use a standard Gaussian on all weights and biases and restrict the standard deviation of the posterior to be at most $\sigma_{max} = 0.1$.
We also experimented with a larger upper limit of $\sigma_{max} = 0.3$, but found this to negatively affect both accuracy and calibration.

All the $\inducing$-space approaches use Gaussian priors $p(\vect(\weight_l)) = \gauss(0, 1 / \sqrt{d_{in}})$, motivated by the connection to \acrshortpl{gp}. Hyperparameter and optimisation details for the inducing weight methods on CIFAR-10 are discussed below in the details on the ablation study.
We train all methods using Adam for $300$ epochs with a learning rate of $10^{-3}$ for the first $200$ epochs and then decay by a factor of $10$.
For the initial $100$ epochs we train without the KL-term of the ELBO and then anneal its weight linearly over the following $50$ epochs.
For the tables and figures in the main text, we set $\lambda_{max} = 0.1$ for Ensemble-$\inducing$ on both datasets, and $\sigma_{max} = 0.1, \lambda_{max} = 0.03$ on CIFAR-10 for \acrshort{ffg}-$\inducing$.
We initialise the entries of the $Z$ matrices by sampling from a zero-mean Gaussian with variance $\frac{1}{M}$ and set the diagonal entries of the $D$ matrices to $10^{-3}$.
For \acrshort{ffg}-$\inducing$ we initialise the mean of the variational Gaussian posterior in $\inducing$-space by sampling from a standard Gaussian and set the initial variances to $10^{-3}$.
For Ensemble-$\inducing$ initialisation, we draw an $M\times M$ shaped sample from a standard Gaussian that is shared across ensemble members and add independent Gaussian noise with a standard deviation of $0.1$ for each member.
We use an ensemble size of $5$.
During optimisation, we draw $1$ MC samples per update step for both \acrshort{ffg}-$\inducing$ and Ensemble-$\inducing$ (such that each ensemble member is used once).
For testing we use $20$ MC samples for all variational methods.
We fit BatchNorm parameters by minimising the \acrshort{nll}.

\paragraph{The study of hyper-parameter selection on CIFAR-10}
We run the inducing weight method with the following options:
\begin{itemize}
    \item Row/column dimensions of $\inducing_l$ ($M$): $M \in \{ 16, 32, 64, 128 \}$. \\We set $M = M_{in} = M_{out}$ except for the last layer, where we use $M_{in} = M$ and $M_{out} = 10$.
    \item $\lambda_{max}$ values for \acrshort{ffg}-$\inducing$ and Ensemble-$\inducing$: $\lambda_{max} \in \{0, 0.03, 0.1, 0.3 \}$. \\When $\lambda_{max} = 0$ it means $q(\weight | \inducing)$ is a delta measure centered at the mean of $p(\weight | \inducing)$.
    \item $\sigma_{\max}$ values for \acrshort{ffg}-$\inducing$: $\sigma_{\max} = \{0, 0.1, 0.3 \}$. \\When $\sigma_{max} = 0$ we use a MAP estimate for $\inducing$.
\end{itemize}
Each experiment is repeated with 5 random seeds to collect the averaged results on a single NVIDIA RTX 2080TI. The models are trained with 100 epochs in total. We first run 50 epochs of maximum likelihood to initialise the model, then run 40 epochs training on the modified variational lower-bound with KL annealing (linear scaling schedule), finally we run 10 epochs of training with the variational lower-bound (i.e. no KL annealing). We use Adam optimiser with learning rate $3e-4$ and the default $\beta_1, \beta_2$ parameters in PyTorch's implementation.

\section{Additional Results}
\label{app:additional_results}

Below in \cref{tab:indistribution_full_resnet50_c10,tab:indistribution_full_resnet50_c100} we provide extended versions of \cref{tab:indistribution}.
This table contains standard errors across the random seeds for the corresponding metrics and we additionally report \acrshortpl{nll} and Brier scores. The error bar results are not available for Ensemble-$\weight$, as it is constructed from the 5 independently trained deterministic neural network with maximum likelihood.


\begin{table*}[ht]
\centering
\caption{Complete in-distribution results for Resnet-50 on CIFAR10}
\newcommand{\fulluncertaintytable}{\toprule
Method & Acc. $\uparrow$ & NLL $\downarrow$ & ECE $\downarrow$ & Brier $\downarrow$\\
\midrule
Deterministic & $94.72{\scriptstyle {\pm} 0.08}$ & $0.43{\scriptstyle {\pm} 0.01}$ & $4.46{\scriptstyle {\pm} 0.08}$ & $0.10{\scriptstyle {\pm} 0.00}$ \\
Ensemble-W & $95.90{\scriptstyle \phantom{{\pm} 0.00}}$ & $0.20{\scriptstyle \phantom{{\pm} 0.00}}$ & $1.08{\scriptstyle \phantom{{\pm} 0.00}}$ & $0.06{\scriptstyle \phantom{{\pm} 0.00}}$ \\
FFG-W & $94.13{\scriptstyle {\pm} 0.08}$ & $0.18{\scriptstyle {\pm} 0.00}$ & $0.50{\scriptstyle {\pm} 0.06}$ & $0.09{\scriptstyle {\pm} 0.00}$ \\
FFG-U (M=64) & $94.40{\scriptstyle {\pm} 0.05}$ & $0.17{\scriptstyle {\pm} 0.00}$ & $0.64{\scriptstyle {\pm} 0.06}$ & $0.08{\scriptstyle {\pm} 0.00}$ \\
FFG-U (M=128) & $94.66{\scriptstyle {\pm} 0.09}$ & $0.17{\scriptstyle {\pm} 0.00}$ & $1.59{\scriptstyle {\pm} 0.06}$ & $0.08{\scriptstyle {\pm} 0.00}$ \\
Ensemble-U (M=64) & $94.94{\scriptstyle {\pm} 0.07}$ & $0.16{\scriptstyle {\pm} 0.00}$ & $0.45{\scriptstyle {\pm} 0.06}$ & $0.08{\scriptstyle {\pm} 0.00}$ \\
Ensemble-U (M=128) & $95.34{\scriptstyle {\pm} 0.05}$ & $0.17{\scriptstyle {\pm} 0.00}$ & $1.29{\scriptstyle {\pm} 0.05}$ & $0.07{\scriptstyle {\pm} 0.00}$ \\
\bottomrule
}
\begin{tabular}{l|cccc}
\toprule
Method & Acc. $\uparrow$ & NLL $\downarrow$ & ECE $\downarrow$ & Brier $\downarrow$\\
\midrule
Deterministic & $75.73{\scriptstyle {\pm} 0.16}$ & $2.14{\scriptstyle {\pm} 0.01}$ & $19.69{\scriptstyle {\pm} 0.15}$ & $0.43{\scriptstyle {\pm} 0.00}$\\
Ensemble-W & $79.33{\scriptstyle \phantom{{\pm} 0.00}}$ & $1.23{\scriptstyle \phantom{{\pm} 0.00}}$ & $6.51{\scriptstyle \phantom{{\pm} 0.00}}$ & $0.31{\scriptstyle \phantom{{\pm} 0.00}}$\\
FFG-W & $74.44{\scriptstyle {\pm} 0.27}$ & $1.01{\scriptstyle {\pm} 0.01}$ & $4.24{\scriptstyle {\pm} 0.10}$ & $0.35{\scriptstyle {\pm} 0.00}$\\
FFG-U (M=64) & $75.37{\scriptstyle {\pm} 0.09}$ & $0.92{\scriptstyle {\pm} 0.01}$ & $2.29{\scriptstyle {\pm} 0.39}$ & $0.34{\scriptstyle {\pm} 0.00}$\\
FFG-U (M=128) & $75.88{\scriptstyle {\pm} 0.13}$ & $0.91{\scriptstyle {\pm} 0.00}$ & $6.66{\scriptstyle {\pm} 0.15}$ & $0.34{\scriptstyle {\pm} 0.00}$\\
Ensemble-U (M=64) & $75.97{\scriptstyle {\pm} 0.12}$ & $0.90{\scriptstyle {\pm} 0.00}$ & $1.12{\scriptstyle {\pm} 0.06}$ & $0.33{\scriptstyle {\pm} 0.00}$\\
Ensemble-U (M=128) & $77.61{\scriptstyle {\pm} 0.11}$ & $0.94{\scriptstyle {\pm} 0.00}$ & $6.00{\scriptstyle {\pm} 0.12}$ & $0.32{\scriptstyle {\pm} 0.00}$\\
\bottomrule

\end{tabular}
\label{tab:indistribution_full_resnet50_c10}
\end{table*}

\begin{table*}[ht]
\centering
\caption{Complete in-distribution results for Resnet-50 on CIFAR100}
\newcommand{\fulluncertaintytable}{\toprule
Method & Acc. $\uparrow$ & NLL $\downarrow$ & ECE $\downarrow$ & Brier $\downarrow$\\
\midrule
Deterministic & $75.73{\scriptstyle {\pm} 0.16}$ & $2.14{\scriptstyle {\pm} 0.01}$ & $19.69{\scriptstyle {\pm} 0.15}$ & $0.43{\scriptstyle {\pm} 0.00}$\\
Ensemble-W & $79.33{\scriptstyle \phantom{{\pm} 0.00}}$ & $1.23{\scriptstyle \phantom{{\pm} 0.00}}$ & $6.51{\scriptstyle \phantom{{\pm} 0.00}}$ & $0.31{\scriptstyle \phantom{{\pm} 0.00}}$\\
FFG-W & $74.44{\scriptstyle {\pm} 0.27}$ & $1.01{\scriptstyle {\pm} 0.01}$ & $4.24{\scriptstyle {\pm} 0.10}$ & $0.35{\scriptstyle {\pm} 0.00}$\\
FFG-U (M=64) & $75.37{\scriptstyle {\pm} 0.09}$ & $0.92{\scriptstyle {\pm} 0.01}$ & $2.29{\scriptstyle {\pm} 0.39}$ & $0.34{\scriptstyle {\pm} 0.00}$\\
FFG-U (M=128) & $75.88{\scriptstyle {\pm} 0.13}$ & $0.91{\scriptstyle {\pm} 0.00}$ & $6.66{\scriptstyle {\pm} 0.15}$ & $0.34{\scriptstyle {\pm} 0.00}$\\
Ensemble-U (M=64) & $75.97{\scriptstyle {\pm} 0.12}$ & $0.90{\scriptstyle {\pm} 0.00}$ & $1.12{\scriptstyle {\pm} 0.06}$ & $0.33{\scriptstyle {\pm} 0.00}$\\
Ensemble-U (M=128) & $77.61{\scriptstyle {\pm} 0.11}$ & $0.94{\scriptstyle {\pm} 0.00}$ & $6.00{\scriptstyle {\pm} 0.12}$ & $0.32{\scriptstyle {\pm} 0.00}$\\
\bottomrule
}
\begin{tabular}{l|cccc}
\toprule
Method & Acc. $\uparrow$ & NLL $\downarrow$ & ECE $\downarrow$ & Brier $\downarrow$\\
\midrule
Deterministic & $75.73{\scriptstyle {\pm} 0.16}$ & $2.14{\scriptstyle {\pm} 0.01}$ & $19.69{\scriptstyle {\pm} 0.15}$ & $0.43{\scriptstyle {\pm} 0.00}$\\
Ensemble-W & $79.33{\scriptstyle \phantom{{\pm} 0.00}}$ & $1.23{\scriptstyle \phantom{{\pm} 0.00}}$ & $6.51{\scriptstyle \phantom{{\pm} 0.00}}$ & $0.31{\scriptstyle \phantom{{\pm} 0.00}}$\\
FFG-W & $74.44{\scriptstyle {\pm} 0.27}$ & $1.01{\scriptstyle {\pm} 0.01}$ & $4.24{\scriptstyle {\pm} 0.10}$ & $0.35{\scriptstyle {\pm} 0.00}$\\
FFG-U (M=64) & $75.37{\scriptstyle {\pm} 0.09}$ & $0.92{\scriptstyle {\pm} 0.01}$ & $2.29{\scriptstyle {\pm} 0.39}$ & $0.34{\scriptstyle {\pm} 0.00}$\\
FFG-U (M=128) & $75.88{\scriptstyle {\pm} 0.13}$ & $0.91{\scriptstyle {\pm} 0.00}$ & $6.66{\scriptstyle {\pm} 0.15}$ & $0.34{\scriptstyle {\pm} 0.00}$\\
Ensemble-U (M=64) & $75.97{\scriptstyle {\pm} 0.12}$ & $0.90{\scriptstyle {\pm} 0.00}$ & $1.12{\scriptstyle {\pm} 0.06}$ & $0.33{\scriptstyle {\pm} 0.00}$\\
Ensemble-U (M=128) & $77.61{\scriptstyle {\pm} 0.11}$ & $0.94{\scriptstyle {\pm} 0.00}$ & $6.00{\scriptstyle {\pm} 0.12}$ & $0.32{\scriptstyle {\pm} 0.00}$\\
\bottomrule

\end{tabular}
\label{tab:indistribution_full_resnet50_c100}
\end{table*}

The results of pruning different fractions of the weights can be found in \cref{tab:indistribution_prune_resnet50} for the in-distribution uncertainty evaluation for Resnet-50 and the \acrshort{ood} detection in \cref{tab:ood_detection_prune_resnet50}.
For the pruning experiments, we take the parameters from the corresponding full runs, set a fixed percentage of the weights to be deterministically $0$ and fine-tune the remaining weights with a new optimizer for $50$ epochs.
We use Adam with a learning rate of $10^{-4}$.
For \acrshort{ffg}-$\weight$ we select the weights with the smallest ratio of absolute mean to standard deviation in the approximate posterior, and for \acrshort{ffg}-$\inducing$ the $Z$ parameters with the smallest absolute value.

For \acrshort{ffg}-$\weight$ we find that pruning up to $90\%$ of the weights only worsens \acrshort{ece} and \acrshort{nll} on the more difficult CIFAR100 datasets.
Pruning $99\%$ of the weights worsens accuracy and \acrshort{ood} detection, but interestingly improves \acrshort{ece} on CIFAR100, where accuracy is noticeably worse.

Pruning $25\%$ and $50\%$ of the $Z$ parameters in \acrshort{ffg}-$\inducing$ results in a total parameter count of $4,408,790$ and $3,106,678$, i.e. $18.7$ and $13.2\%$ of the deterministic parameters respectively on ResNet-50.
Up to pruning $50\%$ of the $Z$ parameters, we find that only \acrshort{ece} becomes slightly worse, although on CIFAR100 it is still better than the \acrshort{ece} for \acrshort{ffg}-$\weight$ at $100\%$ of the weights.
Other metrics are not affected neither on the in-distribution uncertainty or \acrshort{ood} detection, except for a minor drop in accuracy.

\begin{table*}[ht]
\centering
\caption{Pruning in-distribution uncertainty results for Resnet-50. The percentage refers to the weights left for \acrshort{ffg}-$\weight$ and the number of $Z$ parameters for \acrshort{ffg}-$\inducing$.}
\newcommand{\pruninguncertaintytable}{\toprule
& \multicolumn{4}{c}{CIFAR10}& \multicolumn{4}{c}{CIFAR100}\\
Method & Acc. $\uparrow$ & NLL $\downarrow$ & ECE $\downarrow$ & Brier $\downarrow$ & Acc. $\uparrow$ & NLL $\downarrow$ & ECE $\downarrow$ & Brier $\downarrow$\\
\midrule
FFG-W (100\%) & $94.13{\scriptstyle {\pm} 0.08}$ & $0.18{\scriptstyle {\pm} 0.00}$ & $0.50{\scriptstyle {\pm} 0.06}$ & $0.09{\scriptstyle {\pm} 0.00}$ & $74.44{\scriptstyle {\pm} 0.27}$ & $1.01{\scriptstyle {\pm} 0.01}$ & $4.24{\scriptstyle {\pm} 0.10}$ & $0.35{\scriptstyle {\pm} 0.00}$\\
FFG-W (50\%) & $94.07{\scriptstyle {\pm} 0.03}$ & $0.18{\scriptstyle {\pm} 0.00}$ & $0.40{\scriptstyle {\pm} 0.04}$ & $0.09{\scriptstyle {\pm} 0.00}$ & $74.38{\scriptstyle {\pm} 0.21}$ & $1.02{\scriptstyle {\pm} 0.01}$ & $4.15{\scriptstyle {\pm} 0.09}$ & $0.36{\scriptstyle {\pm} 0.00}$\\
FFG-W (10\%) & $94.17{\scriptstyle {\pm} 0.06}$ & $0.18{\scriptstyle {\pm} 0.00}$ & $0.58{\scriptstyle {\pm} 0.02}$ & $0.09{\scriptstyle {\pm} 0.00}$ & $74.42{\scriptstyle {\pm} 0.23}$ & $1.10{\scriptstyle {\pm} 0.01}$ & $6.86{\scriptstyle {\pm} 0.08}$ & $0.36{\scriptstyle {\pm} 0.00}$\\
FFG-W (1\%) & $93.60{\scriptstyle {\pm} 0.09}$ & $0.19{\scriptstyle {\pm} 0.00}$ & $0.80{\scriptstyle {\pm} 0.06}$ & $0.09{\scriptstyle {\pm} 0.00}$ & $67.08{\scriptstyle {\pm} 0.33}$ & $1.19{\scriptstyle {\pm} 0.01}$ & $1.36{\scriptstyle {\pm} 0.18}$ & $0.44{\scriptstyle {\pm} 0.00}$\\
FFG-W (0.1\%) & $58.59{\scriptstyle {\pm} 0.75}$ & $1.15{\scriptstyle {\pm} 0.02}$ & $7.33{\scriptstyle {\pm} 0.23}$ & $0.55{\scriptstyle {\pm} 0.01}$ & $10.66{\scriptstyle {\pm} 0.43}$ & $3.97{\scriptstyle {\pm} 0.03}$ & $2.80{\scriptstyle {\pm} 0.15}$ & $0.96{\scriptstyle {\pm} 0.00}$\\
FFG-U (100\%) & $94.40{\scriptstyle {\pm} 0.05}$ & $0.17{\scriptstyle {\pm} 0.00}$ & $0.64{\scriptstyle {\pm} 0.06}$ & $0.08{\scriptstyle {\pm} 0.00}$ & $75.37{\scriptstyle {\pm} 0.09}$ & $0.92{\scriptstyle {\pm} 0.01}$ & $2.29{\scriptstyle {\pm} 0.39}$ & $0.34{\scriptstyle {\pm} 0.00}$\\
FFG-U (75\%) & $94.45{\scriptstyle {\pm} 0.05}$ & $0.18{\scriptstyle {\pm} 0.00}$ & $2.19{\scriptstyle {\pm} 0.11}$ & $0.09{\scriptstyle {\pm} 0.00}$ & $75.26{\scriptstyle {\pm} 0.07}$ & $0.93{\scriptstyle {\pm} 0.00}$ & $3.74{\scriptstyle {\pm} 0.67}$ & $0.35{\scriptstyle {\pm} 0.00}$\\
FFG-U (50\%) & $94.31{\scriptstyle {\pm} 0.07}$ & $0.18{\scriptstyle {\pm} 0.00}$ & $2.31{\scriptstyle {\pm} 0.09}$ & $0.09{\scriptstyle {\pm} 0.00}$ & $74.89{\scriptstyle {\pm} 0.07}$ & $0.94{\scriptstyle {\pm} 0.00}$ & $5.04{\scriptstyle {\pm} 0.77}$ & $0.35{\scriptstyle {\pm} 0.00}$\\
FFG-U (25\%) & $93.34{\scriptstyle {\pm} 0.03}$ & $0.22{\scriptstyle {\pm} 0.00}$ & $4.83{\scriptstyle {\pm} 0.09}$ & $0.11{\scriptstyle {\pm} 0.00}$ & $71.32{\scriptstyle {\pm} 0.16}$ & $1.09{\scriptstyle {\pm} 0.01}$ & $12.44{\scriptstyle {\pm} 0.90}$ & $0.42{\scriptstyle {\pm} 0.00}$\\
\bottomrule
}
\scalebox{0.9}{
    \begin{tabular}{l|cccccccc}
    \toprule
& \multicolumn{4}{c}{CIFAR10}& \multicolumn{4}{c}{CIFAR100}\\
Method & Acc. $\uparrow$ & NLL $\downarrow$ & ECE $\downarrow$ & Brier $\downarrow$ & Acc. $\uparrow$ & NLL $\downarrow$ & ECE $\downarrow$ & Brier $\downarrow$\\
\midrule
FFG-W (100\%) & $94.13{\scriptstyle {\pm} 0.08}$ & $0.18{\scriptstyle {\pm} 0.00}$ & $0.50{\scriptstyle {\pm} 0.06}$ & $0.09{\scriptstyle {\pm} 0.00}$ & $74.44{\scriptstyle {\pm} 0.27}$ & $1.01{\scriptstyle {\pm} 0.01}$ & $4.24{\scriptstyle {\pm} 0.10}$ & $0.35{\scriptstyle {\pm} 0.00}$\\
FFG-W (50\%) & $94.07{\scriptstyle {\pm} 0.03}$ & $0.18{\scriptstyle {\pm} 0.00}$ & $0.40{\scriptstyle {\pm} 0.04}$ & $0.09{\scriptstyle {\pm} 0.00}$ & $74.38{\scriptstyle {\pm} 0.21}$ & $1.02{\scriptstyle {\pm} 0.01}$ & $4.15{\scriptstyle {\pm} 0.09}$ & $0.36{\scriptstyle {\pm} 0.00}$\\
FFG-W (10\%) & $94.17{\scriptstyle {\pm} 0.06}$ & $0.18{\scriptstyle {\pm} 0.00}$ & $0.58{\scriptstyle {\pm} 0.02}$ & $0.09{\scriptstyle {\pm} 0.00}$ & $74.42{\scriptstyle {\pm} 0.23}$ & $1.10{\scriptstyle {\pm} 0.01}$ & $6.86{\scriptstyle {\pm} 0.08}$ & $0.36{\scriptstyle {\pm} 0.00}$\\
FFG-W (1\%) & $93.60{\scriptstyle {\pm} 0.09}$ & $0.19{\scriptstyle {\pm} 0.00}$ & $0.80{\scriptstyle {\pm} 0.06}$ & $0.09{\scriptstyle {\pm} 0.00}$ & $67.08{\scriptstyle {\pm} 0.33}$ & $1.19{\scriptstyle {\pm} 0.01}$ & $1.36{\scriptstyle {\pm} 0.18}$ & $0.44{\scriptstyle {\pm} 0.00}$\\
FFG-W (0.1\%) & $58.59{\scriptstyle {\pm} 0.75}$ & $1.15{\scriptstyle {\pm} 0.02}$ & $7.33{\scriptstyle {\pm} 0.23}$ & $0.55{\scriptstyle {\pm} 0.01}$ & $10.66{\scriptstyle {\pm} 0.43}$ & $3.97{\scriptstyle {\pm} 0.03}$ & $2.80{\scriptstyle {\pm} 0.15}$ & $0.96{\scriptstyle {\pm} 0.00}$\\
FFG-U (100\%) & $94.40{\scriptstyle {\pm} 0.05}$ & $0.17{\scriptstyle {\pm} 0.00}$ & $0.64{\scriptstyle {\pm} 0.06}$ & $0.08{\scriptstyle {\pm} 0.00}$ & $75.37{\scriptstyle {\pm} 0.09}$ & $0.92{\scriptstyle {\pm} 0.01}$ & $2.29{\scriptstyle {\pm} 0.39}$ & $0.34{\scriptstyle {\pm} 0.00}$\\
FFG-U (75\%) & $94.45{\scriptstyle {\pm} 0.05}$ & $0.18{\scriptstyle {\pm} 0.00}$ & $2.19{\scriptstyle {\pm} 0.11}$ & $0.09{\scriptstyle {\pm} 0.00}$ & $75.26{\scriptstyle {\pm} 0.07}$ & $0.93{\scriptstyle {\pm} 0.00}$ & $3.74{\scriptstyle {\pm} 0.67}$ & $0.35{\scriptstyle {\pm} 0.00}$\\
FFG-U (50\%) & $94.31{\scriptstyle {\pm} 0.07}$ & $0.18{\scriptstyle {\pm} 0.00}$ & $2.31{\scriptstyle {\pm} 0.09}$ & $0.09{\scriptstyle {\pm} 0.00}$ & $74.89{\scriptstyle {\pm} 0.07}$ & $0.94{\scriptstyle {\pm} 0.00}$ & $5.04{\scriptstyle {\pm} 0.77}$ & $0.35{\scriptstyle {\pm} 0.00}$\\
FFG-U (25\%) & $93.34{\scriptstyle {\pm} 0.03}$ & $0.22{\scriptstyle {\pm} 0.00}$ & $4.83{\scriptstyle {\pm} 0.09}$ & $0.11{\scriptstyle {\pm} 0.00}$ & $71.32{\scriptstyle {\pm} 0.16}$ & $1.09{\scriptstyle {\pm} 0.01}$ & $12.44{\scriptstyle {\pm} 0.90}$ & $0.42{\scriptstyle {\pm} 0.00}$\\
\bottomrule

    \end{tabular}
}
\label{tab:indistribution_prune_resnet50}
\end{table*}

\begin{table*}[ht]
\centering
\caption{Pruning \acrshort{ood} detection metrics for Resnet-$50$ trained on CIFAR$10$/$100$.}
\newcommand{\pruningaucoodtable}{\toprule
In-dist $\rightarrow$ OOD & \multicolumn{2}{c}{C10 $\rightarrow$ C100} & \multicolumn{2}{c}{C10 $\rightarrow$ SVHN} & \multicolumn{2}{c}{C100 $\rightarrow$ C10} & \multicolumn{2}{c}{C100 $\rightarrow$ SVHN}\\
Method / Metric & AUROC & AUPR & AUROC & AUPR & AUROC & AUPR & AUROC & AUPR\\
\midrule
FFG-W (100\%) & $.88 {\scriptstyle {\pm} .00}$ & $.90 {\scriptstyle {\pm} .00}$ & $.90 {\scriptstyle {\pm} .01}$ & $.86 {\scriptstyle {\pm} .01}$ & $.76 {\scriptstyle {\pm} .00}$ & $.79 {\scriptstyle {\pm} .00}$ & $.80 {\scriptstyle {\pm} .01}$ & $.69 {\scriptstyle {\pm} .01}$ \\
FFG-W (50\%) & $.88 {\scriptstyle {\pm} .00}$ & $.90 {\scriptstyle {\pm} .00}$ & $.90 {\scriptstyle {\pm} .00}$ & $.86 {\scriptstyle {\pm} .00}$ & $.76 {\scriptstyle {\pm} .00}$ & $.79 {\scriptstyle {\pm} .00}$ & $.79 {\scriptstyle {\pm} .01}$ & $.69 {\scriptstyle {\pm} .01}$ \\
FFG-W (10\%) & $.88 {\scriptstyle {\pm} .00}$ & $.90 {\scriptstyle {\pm} .00}$ & $.90 {\scriptstyle {\pm} .01}$ & $.87 {\scriptstyle {\pm} .01}$ & $.76 {\scriptstyle {\pm} .00}$ & $.79 {\scriptstyle {\pm} .00}$ & $.79 {\scriptstyle {\pm} .01}$ & $.69 {\scriptstyle {\pm} .01}$ \\
FFG-W (1\%) & $.88 {\scriptstyle {\pm} .00}$ & $.89 {\scriptstyle {\pm} .00}$ & $.90 {\scriptstyle {\pm} .01}$ & $.86 {\scriptstyle {\pm} .01}$ & $.72 {\scriptstyle {\pm} .00}$ & $.75 {\scriptstyle {\pm} .00}$ & $.71 {\scriptstyle {\pm} .02}$ & $.57 {\scriptstyle {\pm} .02}$ \\
FFG-W (0.1\%) & $.65 {\scriptstyle {\pm} .01}$ & $.65 {\scriptstyle {\pm} .01}$ & $.38 {\scriptstyle {\pm} .03}$ & $.24 {\scriptstyle {\pm} .03}$ & $.56 {\scriptstyle {\pm} .01}$ & $.59 {\scriptstyle {\pm} .01}$ & $.31 {\scriptstyle {\pm} .04}$ & $.21 {\scriptstyle {\pm} .02}$ \\
FFG-U (100\%) & $.89 {\scriptstyle {\pm} .00}$ & $.91 {\scriptstyle {\pm} .00}$ & $.94 {\scriptstyle {\pm} .01}$ & $.91 {\scriptstyle {\pm} .01}$ & $.77 {\scriptstyle {\pm} .00}$ & $.79 {\scriptstyle {\pm} .00}$ & $.83 {\scriptstyle {\pm} .01}$ & $.74 {\scriptstyle {\pm} .01}$ \\
FFG-U (75\%) & $.89 {\scriptstyle {\pm} .00}$ & $.91 {\scriptstyle {\pm} .00}$ & $.94 {\scriptstyle {\pm} .00}$ & $.91 {\scriptstyle {\pm} .00}$ & $.77 {\scriptstyle {\pm} .00}$ & $.79 {\scriptstyle {\pm} .00}$ & $.82 {\scriptstyle {\pm} .01}$ & $.72 {\scriptstyle {\pm} .02}$ \\
FFG-U (50\%) & $.89 {\scriptstyle {\pm} .00}$ & $.91 {\scriptstyle {\pm} .00}$ & $.93 {\scriptstyle {\pm} .00}$ & $.91 {\scriptstyle {\pm} .00}$ & $.77 {\scriptstyle {\pm} .00}$ & $.79 {\scriptstyle {\pm} .00}$ & $.82 {\scriptstyle {\pm} .01}$ & $.72 {\scriptstyle {\pm} .02}$ \\
FFG-U (25\%) & $.88 {\scriptstyle {\pm} .00}$ & $.90 {\scriptstyle {\pm} .00}$ & $.92 {\scriptstyle {\pm} .01}$ & $.88 {\scriptstyle {\pm} .01}$ & $.75 {\scriptstyle {\pm} .00}$ & $.77 {\scriptstyle {\pm} .00}$ & $.82 {\scriptstyle {\pm} .02}$ & $.72 {\scriptstyle {\pm} .03}$ \\
\bottomrule
}
\begin{tabular}{l|cccccccc}
\toprule
In-dist $\rightarrow$ OOD & \multicolumn{2}{c}{C10 $\rightarrow$ C100} & \multicolumn{2}{c}{C10 $\rightarrow$ SVHN} & \multicolumn{2}{c}{C100 $\rightarrow$ C10} & \multicolumn{2}{c}{C100 $\rightarrow$ SVHN}\\
Method / Metric & AUROC & AUPR & AUROC & AUPR & AUROC & AUPR & AUROC & AUPR\\
\midrule
FFG-W (100\%) & $.88 {\scriptstyle {\pm} .00}$ & $.90 {\scriptstyle {\pm} .00}$ & $.90 {\scriptstyle {\pm} .01}$ & $.86 {\scriptstyle {\pm} .01}$ & $.76 {\scriptstyle {\pm} .00}$ & $.79 {\scriptstyle {\pm} .00}$ & $.80 {\scriptstyle {\pm} .01}$ & $.69 {\scriptstyle {\pm} .01}$ \\
FFG-W (50\%) & $.88 {\scriptstyle {\pm} .00}$ & $.90 {\scriptstyle {\pm} .00}$ & $.90 {\scriptstyle {\pm} .00}$ & $.86 {\scriptstyle {\pm} .00}$ & $.76 {\scriptstyle {\pm} .00}$ & $.79 {\scriptstyle {\pm} .00}$ & $.79 {\scriptstyle {\pm} .01}$ & $.69 {\scriptstyle {\pm} .01}$ \\
FFG-W (10\%) & $.88 {\scriptstyle {\pm} .00}$ & $.90 {\scriptstyle {\pm} .00}$ & $.90 {\scriptstyle {\pm} .01}$ & $.87 {\scriptstyle {\pm} .01}$ & $.76 {\scriptstyle {\pm} .00}$ & $.79 {\scriptstyle {\pm} .00}$ & $.79 {\scriptstyle {\pm} .01}$ & $.69 {\scriptstyle {\pm} .01}$ \\
FFG-W (1\%) & $.88 {\scriptstyle {\pm} .00}$ & $.89 {\scriptstyle {\pm} .00}$ & $.90 {\scriptstyle {\pm} .01}$ & $.86 {\scriptstyle {\pm} .01}$ & $.72 {\scriptstyle {\pm} .00}$ & $.75 {\scriptstyle {\pm} .00}$ & $.71 {\scriptstyle {\pm} .02}$ & $.57 {\scriptstyle {\pm} .02}$ \\
FFG-W (0.1\%) & $.65 {\scriptstyle {\pm} .01}$ & $.65 {\scriptstyle {\pm} .01}$ & $.38 {\scriptstyle {\pm} .03}$ & $.24 {\scriptstyle {\pm} .03}$ & $.56 {\scriptstyle {\pm} .01}$ & $.59 {\scriptstyle {\pm} .01}$ & $.31 {\scriptstyle {\pm} .04}$ & $.21 {\scriptstyle {\pm} .02}$ \\
FFG-U (100\%) & $.89 {\scriptstyle {\pm} .00}$ & $.91 {\scriptstyle {\pm} .00}$ & $.94 {\scriptstyle {\pm} .01}$ & $.91 {\scriptstyle {\pm} .01}$ & $.77 {\scriptstyle {\pm} .00}$ & $.79 {\scriptstyle {\pm} .00}$ & $.83 {\scriptstyle {\pm} .01}$ & $.74 {\scriptstyle {\pm} .01}$ \\
FFG-U (75\%) & $.89 {\scriptstyle {\pm} .00}$ & $.91 {\scriptstyle {\pm} .00}$ & $.94 {\scriptstyle {\pm} .00}$ & $.91 {\scriptstyle {\pm} .00}$ & $.77 {\scriptstyle {\pm} .00}$ & $.79 {\scriptstyle {\pm} .00}$ & $.82 {\scriptstyle {\pm} .01}$ & $.72 {\scriptstyle {\pm} .02}$ \\
FFG-U (50\%) & $.89 {\scriptstyle {\pm} .00}$ & $.91 {\scriptstyle {\pm} .00}$ & $.93 {\scriptstyle {\pm} .00}$ & $.91 {\scriptstyle {\pm} .00}$ & $.77 {\scriptstyle {\pm} .00}$ & $.79 {\scriptstyle {\pm} .00}$ & $.82 {\scriptstyle {\pm} .01}$ & $.72 {\scriptstyle {\pm} .02}$ \\
FFG-U (25\%) & $.88 {\scriptstyle {\pm} .00}$ & $.90 {\scriptstyle {\pm} .00}$ & $.92 {\scriptstyle {\pm} .01}$ & $.88 {\scriptstyle {\pm} .01}$ & $.75 {\scriptstyle {\pm} .00}$ & $.77 {\scriptstyle {\pm} .00}$ & $.82 {\scriptstyle {\pm} .02}$ & $.72 {\scriptstyle {\pm} .03}$ \\
\bottomrule

\end{tabular}
\label{tab:ood_detection_prune_resnet50}
\end{table*}

The number of parameters for \acrshort{ffg}-U and Ensemble-U with an ensemble size of 5 in the ResNet-50 experiments are reported in \cref{tab:parameter_count}.
The corresponding parameter counts for pruning \acrshort{ffg}-$\weight$ and \acrshort{ffg}-$\inducing$ (M=64) are in \cref{tab:parameter_count_pruning}.

\begin{table*}[ht]
    \centering
    \caption{Parameter counts for the inducing models with varying $\inducing$ size $M$.}
    \scalebox{0.9}{
        \begin{tabular}{ll|ccccc|c}
            \toprule
            $M$ & Method & $M=16$ & $M=32$ & $M=64$ & $M=128$ & $M=256$ & Deterministic \\
            \midrule
            Abs. value & \acrshort{ffg}-U & $1,384,662$ & $2,771,446$ & $5,710,902$ & $12,253,366$ & $27,992,502$ & $23,520,842$ \\
            & Ensemble-U & $1,426,134$ & $2,937,334$ & $6,374,454$ & $14,907,574$ & $38,609,334$ & \\
            rel. size ($\%$) & \acrshort{ffg}-U & $5.89$ &  $11.78$ &  $24.28$ &  $52.10$ & $119.01$ & $100$ \\
            & Ensemble-U & $6.06$ & $12.49$ & $27.10$ & $63.38$ & $164.15$ &\\
            \bottomrule
        \end{tabular}
    }
    \label{tab:parameter_count}
\end{table*}

\begin{table*}[ht]
    \centering
    \caption{Pruning parameter counts for keeping fractions of the weights in \acrshort{ffg}-$\weight$ and the $Z$ parameters in \acrshort{ffg}-$\inducing$.}
    \begin{tabular}{l|cc}
        \toprule
        Method & Abs. param. count & rel. size (\%) \\
        \midrule 
        \acrshort{ffg}-$\weight$ ($100\%$) & $46,988,564$ & $199.8$ \\
        \acrshort{ffg}-$\weight$ ($50\%$) & $23,520,852$ & $100$ \\
        \acrshort{ffg}-$\weight$ ($10\%$) & $4,746,682$ & $20.2$ \\
        \acrshort{ffg}-$\weight$ ($1\%$) & $522,494$ & $2.2$ \\
        \acrshort{ffg}-$\weight$ ($0.1\%$) & $100,075$ & $0.4$ \\ 
        \acrshort{ffg}-$\inducing$ ($100\%$) & $5,710,902$ & $24.28$ \\
        \acrshort{ffg}-$\inducing$ ($75\%$) & $4,408,790$ & $18.7$ \\
        \acrshort{ffg}-$\inducing$ ($50\%$) & $3,106,678$ & $13.2$ \\
        \acrshort{ffg}-$\inducing$ ($25\%$) & $1,804,566$ & $7.7$ \\
        \bottomrule
    \end{tabular}
    \label{tab:parameter_count_pruning}
\end{table*}

In \cref{tab:corrupted_c10_accuracy,tab:corrupted_c10_ece,tab:corrupted_c100_accuracy,tab:corrupted_c100_ece} we report the numerical results for \cref{fig:cifar_corrupted}.

\begin{table*}[ht]
\centering
\caption{Corrupted CIFAR-10 accuracy ($\uparrow$) values (in $\%$).}
\newcommand{\corruptedcifartenacctable}{\toprule
& \multicolumn{5}{c}{Skew Intensity}\\
Method & 1 & 2 & 3 & 4 & 5\\
\midrule
Deterministic & $87.90 {\scriptstyle \pm 2.31}$ & $82.02 {\scriptstyle \pm 2.84}$ & $76.31 {\scriptstyle \pm 3.80}$ & $68.91 {\scriptstyle \pm 4.81}$ & $57.94 {\scriptstyle \pm 5.10}$\\
Ensemble-W & $89.45 {\scriptstyle \pm 2.27}$ & $83.94 {\scriptstyle \pm 2.71}$ & $78.40 {\scriptstyle \pm 3.61}$ & $71.18 {\scriptstyle \pm 4.53}$ & $60.15 {\scriptstyle \pm 4.82}$\\
FFG-W & $83.80 {\scriptstyle \pm 2.43}$ & $76.22 {\scriptstyle \pm 3.10}$ & $69.30 {\scriptstyle \pm 4.11}$ & $61.82 {\scriptstyle \pm 4.66}$ & $50.72 {\scriptstyle \pm 4.68}$\\
FFG-U & $86.90 {\scriptstyle \pm 2.47}$ & $80.33 {\scriptstyle \pm 3.14}$ & $74.34 {\scriptstyle \pm 4.06}$ & $67.23 {\scriptstyle \pm 4.75}$ & $57.00 {\scriptstyle \pm 4.83}$\\
Ensemble-U & $87.35 {\scriptstyle \pm 2.39}$ & $80.45 {\scriptstyle \pm 3.19}$ & $73.89 {\scriptstyle \pm 4.23}$ & $66.52 {\scriptstyle \pm 4.96}$ & $54.89 {\scriptstyle \pm 5.26}$\\
\bottomrule
}
\begin{tabular}{l|ccccc}
\toprule
& \multicolumn{5}{c}{Skew Intensity}\\
Method & 1 & 2 & 3 & 4 & 5\\
\midrule
Deterministic & $87.90 {\scriptstyle \pm 2.31}$ & $82.02 {\scriptstyle \pm 2.84}$ & $76.31 {\scriptstyle \pm 3.80}$ & $68.91 {\scriptstyle \pm 4.81}$ & $57.94 {\scriptstyle \pm 5.10}$\\
Ensemble-W & $89.45 {\scriptstyle \pm 2.27}$ & $83.94 {\scriptstyle \pm 2.71}$ & $78.40 {\scriptstyle \pm 3.61}$ & $71.18 {\scriptstyle \pm 4.53}$ & $60.15 {\scriptstyle \pm 4.82}$\\
FFG-W & $83.80 {\scriptstyle \pm 2.43}$ & $76.22 {\scriptstyle \pm 3.10}$ & $69.30 {\scriptstyle \pm 4.11}$ & $61.82 {\scriptstyle \pm 4.66}$ & $50.72 {\scriptstyle \pm 4.68}$\\
FFG-U & $86.90 {\scriptstyle \pm 2.47}$ & $80.33 {\scriptstyle \pm 3.14}$ & $74.34 {\scriptstyle \pm 4.06}$ & $67.23 {\scriptstyle \pm 4.75}$ & $57.00 {\scriptstyle \pm 4.83}$\\
Ensemble-U & $87.35 {\scriptstyle \pm 2.39}$ & $80.45 {\scriptstyle \pm 3.19}$ & $73.89 {\scriptstyle \pm 4.23}$ & $66.52 {\scriptstyle \pm 4.96}$ & $54.89 {\scriptstyle \pm 5.26}$\\
\bottomrule

\end{tabular}
\label{tab:corrupted_c10_accuracy}
\end{table*}

\begin{table*}[ht]
\centering
\caption{Corrupted CIFAR-10 \acrshort{ece} ($\downarrow$) values (in $\%$).}
\newcommand{\corruptedcifartenecetable}{\toprule
& \multicolumn{5}{c}{Skew Intensity}\\
Method & 1 & 2 & 3 & 4 & 5\\
\midrule
Deterministic & $10.41 {\scriptstyle \pm 2.03}$ & $15.58 {\scriptstyle \pm 2.52}$ & $20.56 {\scriptstyle \pm 3.37}$ & $27.06 {\scriptstyle \pm 4.23}$ & $37.26 {\scriptstyle \pm 4.65}$\\
Ensemble-W & $4.12 {\scriptstyle \pm 1.31}$ & $7.01 {\scriptstyle \pm 1.64}$ & $10.10 {\scriptstyle \pm 2.33}$ & $14.12 {\scriptstyle \pm 2.84}$ & $20.48 {\scriptstyle \pm 2.95}$\\
FFG-W & $13.05 {\scriptstyle \pm 0.64}$ & $12.14 {\scriptstyle \pm 0.89}$ & $11.56 {\scriptstyle \pm 0.93}$ & $10.77 {\scriptstyle \pm 1.05}$ & $10.86 {\scriptstyle \pm 1.31}$\\
FFG-U & $2.47 {\scriptstyle \pm 1.04}$ & $4.93 {\scriptstyle \pm 1.66}$ & $7.77 {\scriptstyle \pm 2.44}$ & $11.27 {\scriptstyle \pm 2.81}$ & $16.16 {\scriptstyle \pm 2.92}$\\
Ensemble-U & $2.77 {\scriptstyle \pm 1.04}$ & $5.86 {\scriptstyle \pm 1.69}$ & $9.06 {\scriptstyle \pm 2.40}$ & $12.54 {\scriptstyle \pm 2.66}$ & $19.66 {\scriptstyle \pm 3.12}$\\
\bottomrule
}
\begin{tabular}{l|ccccc}
\toprule
& \multicolumn{5}{c}{Skew Intensity}\\
Method & 1 & 2 & 3 & 4 & 5\\
\midrule
Deterministic & $10.41 {\scriptstyle \pm 2.03}$ & $15.58 {\scriptstyle \pm 2.52}$ & $20.56 {\scriptstyle \pm 3.37}$ & $27.06 {\scriptstyle \pm 4.23}$ & $37.26 {\scriptstyle \pm 4.65}$\\
Ensemble-W & $4.12 {\scriptstyle \pm 1.31}$ & $7.01 {\scriptstyle \pm 1.64}$ & $10.10 {\scriptstyle \pm 2.33}$ & $14.12 {\scriptstyle \pm 2.84}$ & $20.48 {\scriptstyle \pm 2.95}$\\
FFG-W & $13.05 {\scriptstyle \pm 0.64}$ & $12.14 {\scriptstyle \pm 0.89}$ & $11.56 {\scriptstyle \pm 0.93}$ & $10.77 {\scriptstyle \pm 1.05}$ & $10.86 {\scriptstyle \pm 1.31}$\\
FFG-U & $2.47 {\scriptstyle \pm 1.04}$ & $4.93 {\scriptstyle \pm 1.66}$ & $7.77 {\scriptstyle \pm 2.44}$ & $11.27 {\scriptstyle \pm 2.81}$ & $16.16 {\scriptstyle \pm 2.92}$\\
Ensemble-U & $2.77 {\scriptstyle \pm 1.04}$ & $5.86 {\scriptstyle \pm 1.69}$ & $9.06 {\scriptstyle \pm 2.40}$ & $12.54 {\scriptstyle \pm 2.66}$ & $19.66 {\scriptstyle \pm 3.12}$\\
\bottomrule

\end{tabular}
\label{tab:corrupted_c10_ece}
\end{table*}

\begin{table*}[ht]
\centering
\caption{Corrupted CIFAR-100 accuracy ($\uparrow$) values (in $\%$).}
\newcommand{\corruptedcifarhundredacctable}{\toprule
& \multicolumn{5}{c}{Skew Intensity}\\
Method & 1 & 2 & 3 & 4 & 5\\
\midrule
Deterministic & $63.18 {\scriptstyle \pm 3.06}$ & $54.23 {\scriptstyle \pm 3.67}$ & $48.47 {\scriptstyle \pm 4.27}$ & $41.84 {\scriptstyle \pm 4.59}$ & $31.96 {\scriptstyle \pm 3.96}$\\
Ensemble-W & $67.10 {\scriptstyle \pm 3.19}$ & $57.92 {\scriptstyle \pm 3.82}$ & $51.83 {\scriptstyle \pm 4.50}$ & $45.16 {\scriptstyle \pm 4.91}$ & $34.93 {\scriptstyle \pm 4.27}$\\
FFG-W & $57.49 {\scriptstyle \pm 3.17}$ & $47.62 {\scriptstyle \pm 3.64}$ & $41.99 {\scriptstyle \pm 4.15}$ & $35.61 {\scriptstyle \pm 4.24}$ & $26.59 {\scriptstyle \pm 3.66}$\\
FFG-U & $61.71 {\scriptstyle \pm 3.35}$ & $52.61 {\scriptstyle \pm 3.84}$ & $47.08 {\scriptstyle \pm 4.36}$ & $40.56 {\scriptstyle \pm 4.54}$ & $30.88 {\scriptstyle \pm 3.93}$\\
Ensemble-U & $61.87 {\scriptstyle \pm 3.36}$ & $52.69 {\scriptstyle \pm 3.97}$ & $46.96 {\scriptstyle \pm 4.52}$ & $40.59 {\scriptstyle \pm 4.72}$ & $30.85 {\scriptstyle \pm 3.98}$\\
\bottomrule
}
\begin{tabular}{l|ccccc}
\toprule
& \multicolumn{5}{c}{Skew Intensity}\\
Method & 1 & 2 & 3 & 4 & 5\\
\midrule
Deterministic & $63.18 {\scriptstyle \pm 3.06}$ & $54.23 {\scriptstyle \pm 3.67}$ & $48.47 {\scriptstyle \pm 4.27}$ & $41.84 {\scriptstyle \pm 4.59}$ & $31.96 {\scriptstyle \pm 3.96}$\\
Ensemble-W & $67.10 {\scriptstyle \pm 3.19}$ & $57.92 {\scriptstyle \pm 3.82}$ & $51.83 {\scriptstyle \pm 4.50}$ & $45.16 {\scriptstyle \pm 4.91}$ & $34.93 {\scriptstyle \pm 4.27}$\\
FFG-W & $57.49 {\scriptstyle \pm 3.17}$ & $47.62 {\scriptstyle \pm 3.64}$ & $41.99 {\scriptstyle \pm 4.15}$ & $35.61 {\scriptstyle \pm 4.24}$ & $26.59 {\scriptstyle \pm 3.66}$\\
FFG-U & $61.71 {\scriptstyle \pm 3.35}$ & $52.61 {\scriptstyle \pm 3.84}$ & $47.08 {\scriptstyle \pm 4.36}$ & $40.56 {\scriptstyle \pm 4.54}$ & $30.88 {\scriptstyle \pm 3.93}$\\
Ensemble-U & $61.87 {\scriptstyle \pm 3.36}$ & $52.69 {\scriptstyle \pm 3.97}$ & $46.96 {\scriptstyle \pm 4.52}$ & $40.59 {\scriptstyle \pm 4.72}$ & $30.85 {\scriptstyle \pm 3.98}$\\
\bottomrule

\end{tabular}
\label{tab:corrupted_c100_accuracy}
\end{table*}

\begin{table*}[ht]
\centering
\caption{Corrupted CIFAR-100 \acrshort{ece} ($\downarrow$) values (in $\%$).}
\newcommand{\corruptedcifarhundredecetable}{\toprule
& \multicolumn{5}{c}{Skew Intensity}\\
Method & 1 & 2 & 3 & 4 & 5\\
\midrule
Deterministic & $30.06 {\scriptstyle \pm 2.70}$ & $37.50 {\scriptstyle \pm 3.17}$ & $42.48 {\scriptstyle \pm 3.66}$ & $48.41 {\scriptstyle \pm 4.06}$ & $57.19 {\scriptstyle \pm 3.60}$\\
Ensemble-W & $12.31 {\scriptstyle \pm 2.04}$ & $17.03 {\scriptstyle \pm 2.31}$ & $20.36 {\scriptstyle \pm 2.61}$ & $24.16 {\scriptstyle \pm 2.98}$ & $29.72 {\scriptstyle \pm 2.49}$\\
FFG-W & $14.28 {\scriptstyle \pm 0.78}$ & $11.07 {\scriptstyle \pm 1.11}$ & $11.13 {\scriptstyle \pm 0.85}$ & $11.21 {\scriptstyle \pm 1.29}$ & $11.96 {\scriptstyle \pm 1.68}$\\
FFG-U & $4.64 {\scriptstyle \pm 1.66}$ & $7.80 {\scriptstyle \pm 2.03}$ & $10.42 {\scriptstyle \pm 2.39}$ & $13.98 {\scriptstyle \pm 2.83}$ & $19.17 {\scriptstyle \pm 2.72}$\\
Ensemble-U & $5.84 {\scriptstyle \pm 1.91}$ & $10.28 {\scriptstyle \pm 2.43}$ & $13.60 {\scriptstyle \pm 2.96}$ & $17.54 {\scriptstyle \pm 3.42}$ & $23.56 {\scriptstyle \pm 3.16}$\\
\bottomrule
}
\begin{tabular}{l|ccccc}
\toprule
& \multicolumn{5}{c}{Skew Intensity}\\
Method & 1 & 2 & 3 & 4 & 5\\
\midrule
Deterministic & $30.06 {\scriptstyle \pm 2.70}$ & $37.50 {\scriptstyle \pm 3.17}$ & $42.48 {\scriptstyle \pm 3.66}$ & $48.41 {\scriptstyle \pm 4.06}$ & $57.19 {\scriptstyle \pm 3.60}$\\
Ensemble-W & $12.31 {\scriptstyle \pm 2.04}$ & $17.03 {\scriptstyle \pm 2.31}$ & $20.36 {\scriptstyle \pm 2.61}$ & $24.16 {\scriptstyle \pm 2.98}$ & $29.72 {\scriptstyle \pm 2.49}$\\
FFG-W & $14.28 {\scriptstyle \pm 0.78}$ & $11.07 {\scriptstyle \pm 1.11}$ & $11.13 {\scriptstyle \pm 0.85}$ & $11.21 {\scriptstyle \pm 1.29}$ & $11.96 {\scriptstyle \pm 1.68}$\\
FFG-U & $4.64 {\scriptstyle \pm 1.66}$ & $7.80 {\scriptstyle \pm 2.03}$ & $10.42 {\scriptstyle \pm 2.39}$ & $13.98 {\scriptstyle \pm 2.83}$ & $19.17 {\scriptstyle \pm 2.72}$\\
Ensemble-U & $5.84 {\scriptstyle \pm 1.91}$ & $10.28 {\scriptstyle \pm 2.43}$ & $13.60 {\scriptstyle \pm 2.96}$ & $17.54 {\scriptstyle \pm 3.42}$ & $23.56 {\scriptstyle \pm 3.16}$\\
\bottomrule

\end{tabular}
\label{tab:corrupted_c100_ece}
\end{table*}

\begin{figure*}[t]
\centering
\includegraphics[width=\linewidth]{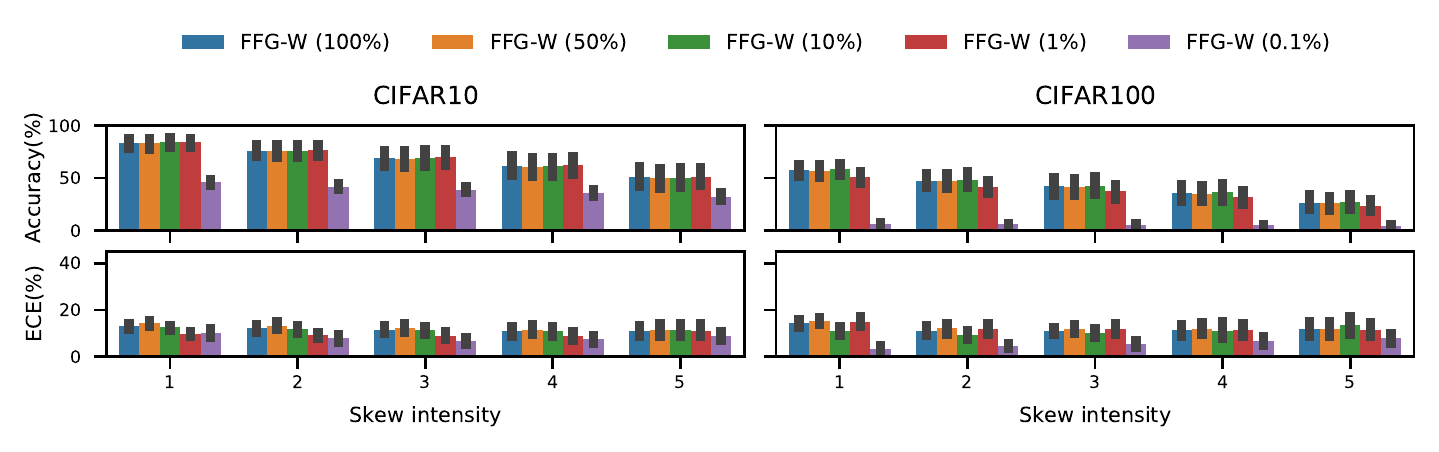}
\caption{Accuracy ($\uparrow$) and \acrshort{ece} ($\downarrow$) on corrupted CIFAR for pruning \acrshort{ffg}-W. We show the mean and two standard errors for each metric on the $19$ perturbations provided in \citep{hendrycks2019benchmarking}.}
\label{fig:cifar_corrupted_pruning_w}
\end{figure*}

In \cref{tab:corrupted_c10_accuracy_pruning,tab:corrupted_c10_ece_pruning,tab:corrupted_c100_accuracy_pruning,tab:corrupted_c100_ece_pruning} we report the corresponding results for pruning \acrshort{ffg}-W and \acrshort{ffg}-U.
See \cref{fig:cifar_corrupted_pruning_w,fig:cifar_corrupted_pruning_w} for visualisation.

\begin{table*}[ht]
\centering
\caption{Corrupted CIFAR-10 accuracy ($\uparrow$) values (in $\%$) for pruning \acrshort{ffg}-W and \acrshort{ffg}-U.}
\newcommand{\corruptedcifartenaccpruningtable}{\toprule
& \multicolumn{5}{c}{Skew Intensity}\\
Method & 1 & 2 & 3 & 4 & 5\\
\midrule
FFG-W (100\%) & $83.80 {\scriptstyle \pm 2.43}$ & $76.22 {\scriptstyle \pm 3.10}$ & $69.30 {\scriptstyle \pm 4.11}$ & $61.82 {\scriptstyle \pm 4.66}$ & $50.72 {\scriptstyle \pm 4.68}$\\
FFG-W (50\%) & $83.39 {\scriptstyle \pm 2.66}$ & $75.58 {\scriptstyle \pm 3.23}$ & $68.43 {\scriptstyle \pm 4.21}$ & $60.69 {\scriptstyle \pm 4.75}$ & $49.75 {\scriptstyle \pm 4.73}$\\
FFG-W (10\%) & $84.04 {\scriptstyle \pm 2.61}$ & $76.21 {\scriptstyle \pm 3.24}$ & $69.35 {\scriptstyle \pm 4.18}$ & $61.65 {\scriptstyle \pm 4.72}$ & $50.51 {\scriptstyle \pm 4.74}$\\
FFG-W (1\%) & $84.16 {\scriptstyle \pm 2.31}$ & $76.48 {\scriptstyle \pm 3.02}$ & $69.72 {\scriptstyle \pm 3.92}$ & $62.72 {\scriptstyle \pm 4.47}$ & $51.26 {\scriptstyle \pm 4.58}$\\
FFG-W (0.1\%) & $46.33 {\scriptstyle \pm 1.30}$ & $41.92 {\scriptstyle \pm 1.44}$ & $38.91 {\scriptstyle \pm 1.59}$ & $36.03 {\scriptstyle \pm 1.76}$ & $32.43 {\scriptstyle \pm 1.88}$\\
FFG-U (100\%) & $86.90 {\scriptstyle \pm 2.47}$ & $80.33 {\scriptstyle \pm 3.14}$ & $74.34 {\scriptstyle \pm 4.06}$ & $67.23 {\scriptstyle \pm 4.75}$ & $57.00 {\scriptstyle \pm 4.83}$\\
FFG-U (75\%) & $86.99 {\scriptstyle \pm 2.37}$ & $80.64 {\scriptstyle \pm 2.95}$ & $74.99 {\scriptstyle \pm 3.81}$ & $67.87 {\scriptstyle \pm 4.53}$ & $57.33 {\scriptstyle \pm 4.65}$\\
FFG-U (50\%) & $86.93 {\scriptstyle \pm 2.36}$ & $80.70 {\scriptstyle \pm 2.93}$ & $75.10 {\scriptstyle \pm 3.76}$ & $68.14 {\scriptstyle \pm 4.44}$ & $57.66 {\scriptstyle \pm 4.49}$\\
FFG-U (25\%) & $85.93 {\scriptstyle \pm 2.39}$ & $79.41 {\scriptstyle \pm 2.96}$ & $73.48 {\scriptstyle \pm 3.82}$ & $66.52 {\scriptstyle \pm 4.52}$ & $55.57 {\scriptstyle \pm 4.59}$\\
\bottomrule
}
\begin{tabular}{l|ccccc}
\toprule
& \multicolumn{5}{c}{Skew Intensity}\\
Method & 1 & 2 & 3 & 4 & 5\\
\midrule
FFG-W (100\%) & $83.80 {\scriptstyle \pm 2.43}$ & $76.22 {\scriptstyle \pm 3.10}$ & $69.30 {\scriptstyle \pm 4.11}$ & $61.82 {\scriptstyle \pm 4.66}$ & $50.72 {\scriptstyle \pm 4.68}$\\
FFG-W (50\%) & $83.39 {\scriptstyle \pm 2.66}$ & $75.58 {\scriptstyle \pm 3.23}$ & $68.43 {\scriptstyle \pm 4.21}$ & $60.69 {\scriptstyle \pm 4.75}$ & $49.75 {\scriptstyle \pm 4.73}$\\
FFG-W (10\%) & $84.04 {\scriptstyle \pm 2.61}$ & $76.21 {\scriptstyle \pm 3.24}$ & $69.35 {\scriptstyle \pm 4.18}$ & $61.65 {\scriptstyle \pm 4.72}$ & $50.51 {\scriptstyle \pm 4.74}$\\
FFG-W (1\%) & $84.16 {\scriptstyle \pm 2.31}$ & $76.48 {\scriptstyle \pm 3.02}$ & $69.72 {\scriptstyle \pm 3.92}$ & $62.72 {\scriptstyle \pm 4.47}$ & $51.26 {\scriptstyle \pm 4.58}$\\
FFG-W (0.1\%) & $46.33 {\scriptstyle \pm 1.30}$ & $41.92 {\scriptstyle \pm 1.44}$ & $38.91 {\scriptstyle \pm 1.59}$ & $36.03 {\scriptstyle \pm 1.76}$ & $32.43 {\scriptstyle \pm 1.88}$\\
FFG-U (100\%) & $86.90 {\scriptstyle \pm 2.47}$ & $80.33 {\scriptstyle \pm 3.14}$ & $74.34 {\scriptstyle \pm 4.06}$ & $67.23 {\scriptstyle \pm 4.75}$ & $57.00 {\scriptstyle \pm 4.83}$\\
FFG-U (75\%) & $86.99 {\scriptstyle \pm 2.37}$ & $80.64 {\scriptstyle \pm 2.95}$ & $74.99 {\scriptstyle \pm 3.81}$ & $67.87 {\scriptstyle \pm 4.53}$ & $57.33 {\scriptstyle \pm 4.65}$\\
FFG-U (50\%) & $86.93 {\scriptstyle \pm 2.36}$ & $80.70 {\scriptstyle \pm 2.93}$ & $75.10 {\scriptstyle \pm 3.76}$ & $68.14 {\scriptstyle \pm 4.44}$ & $57.66 {\scriptstyle \pm 4.49}$\\
FFG-U (25\%) & $85.93 {\scriptstyle \pm 2.39}$ & $79.41 {\scriptstyle \pm 2.96}$ & $73.48 {\scriptstyle \pm 3.82}$ & $66.52 {\scriptstyle \pm 4.52}$ & $55.57 {\scriptstyle \pm 4.59}$\\
\bottomrule

\end{tabular}
\label{tab:corrupted_c10_accuracy_pruning}
\end{table*}

\begin{table*}[ht]
\centering
\caption{Corrupted CIFAR-10 \acrshort{ece} ($\downarrow$) values (in $\%$) for pruning \acrshort{ffg}-W and \acrshort{ffg}-U.}
\newcommand{\corruptedcifartenecepruningtable}{\toprule
& \multicolumn{5}{c}{Skew Intensity}\\
Method & 1 & 2 & 3 & 4 & 5\\
\midrule
FFG-W (100\%) & $13.05 {\scriptstyle \pm 0.64}$ & $12.14 {\scriptstyle \pm 0.89}$ & $11.56 {\scriptstyle \pm 0.93}$ & $10.77 {\scriptstyle \pm 1.05}$ & $10.86 {\scriptstyle \pm 1.31}$\\
FFG-W (50\%) & $14.27 {\scriptstyle \pm 0.59}$ & $13.19 {\scriptstyle \pm 0.88}$ & $12.36 {\scriptstyle \pm 0.94}$ & $11.59 {\scriptstyle \pm 1.07}$ & $11.43 {\scriptstyle \pm 1.33}$\\
FFG-W (10\%) & $12.50 {\scriptstyle \pm 0.52}$ & $11.66 {\scriptstyle \pm 0.76}$ & $11.33 {\scriptstyle \pm 0.90}$ & $10.86 {\scriptstyle \pm 1.04}$ & $11.28 {\scriptstyle \pm 1.38}$\\
FFG-W (1\%) & $9.86 {\scriptstyle \pm 0.49}$ & $9.17 {\scriptstyle \pm 0.62}$ & $8.85 {\scriptstyle \pm 0.87}$ & $8.87 {\scriptstyle \pm 0.93}$ & $11.10 {\scriptstyle \pm 1.45}$\\
FFG-W (0.1\%) & $10.08 {\scriptstyle \pm 0.87}$ & $7.82 {\scriptstyle \pm 0.96}$ & $6.74 {\scriptstyle \pm 0.81}$ & $7.34 {\scriptstyle \pm 0.79}$ & $8.77 {\scriptstyle \pm 0.93}$\\
FFG-U (100\%) & $2.47 {\scriptstyle \pm 1.04}$ & $4.93 {\scriptstyle \pm 1.66}$ & $7.77 {\scriptstyle \pm 2.44}$ & $11.27 {\scriptstyle \pm 2.81}$ & $16.16 {\scriptstyle \pm 2.92}$\\
FFG-U (75\%) & $2.79 {\scriptstyle \pm 0.60}$ & $3.92 {\scriptstyle \pm 0.92}$ & $5.27 {\scriptstyle \pm 1.59}$ & $7.63 {\scriptstyle \pm 2.00}$ & $11.80 {\scriptstyle \pm 2.39}$\\
FFG-U (50\%) & $3.11 {\scriptstyle \pm 0.59}$ & $4.26 {\scriptstyle \pm 0.92}$ & $5.44 {\scriptstyle \pm 1.58}$ & $7.58 {\scriptstyle \pm 1.94}$ & $11.38 {\scriptstyle \pm 2.26}$\\
FFG-U (25\%) & $5.27 {\scriptstyle \pm 0.35}$ & $5.33 {\scriptstyle \pm 0.57}$ & $6.20 {\scriptstyle \pm 1.19}$ & $7.95 {\scriptstyle \pm 1.55}$ & $11.19 {\scriptstyle \pm 2.21}$\\
\bottomrule
}
\begin{tabular}{l|ccccc}
\toprule
& \multicolumn{5}{c}{Skew Intensity}\\
Method & 1 & 2 & 3 & 4 & 5\\
\midrule
FFG-W (100\%) & $13.05 {\scriptstyle \pm 0.64}$ & $12.14 {\scriptstyle \pm 0.89}$ & $11.56 {\scriptstyle \pm 0.93}$ & $10.77 {\scriptstyle \pm 1.05}$ & $10.86 {\scriptstyle \pm 1.31}$\\
FFG-W (50\%) & $14.27 {\scriptstyle \pm 0.59}$ & $13.19 {\scriptstyle \pm 0.88}$ & $12.36 {\scriptstyle \pm 0.94}$ & $11.59 {\scriptstyle \pm 1.07}$ & $11.43 {\scriptstyle \pm 1.33}$\\
FFG-W (10\%) & $12.50 {\scriptstyle \pm 0.52}$ & $11.66 {\scriptstyle \pm 0.76}$ & $11.33 {\scriptstyle \pm 0.90}$ & $10.86 {\scriptstyle \pm 1.04}$ & $11.28 {\scriptstyle \pm 1.38}$\\
FFG-W (1\%) & $9.86 {\scriptstyle \pm 0.49}$ & $9.17 {\scriptstyle \pm 0.62}$ & $8.85 {\scriptstyle \pm 0.87}$ & $8.87 {\scriptstyle \pm 0.93}$ & $11.10 {\scriptstyle \pm 1.45}$\\
FFG-W (0.1\%) & $10.08 {\scriptstyle \pm 0.87}$ & $7.82 {\scriptstyle \pm 0.96}$ & $6.74 {\scriptstyle \pm 0.81}$ & $7.34 {\scriptstyle \pm 0.79}$ & $8.77 {\scriptstyle \pm 0.93}$\\
FFG-U (100\%) & $2.47 {\scriptstyle \pm 1.04}$ & $4.93 {\scriptstyle \pm 1.66}$ & $7.77 {\scriptstyle \pm 2.44}$ & $11.27 {\scriptstyle \pm 2.81}$ & $16.16 {\scriptstyle \pm 2.92}$\\
FFG-U (75\%) & $2.79 {\scriptstyle \pm 0.60}$ & $3.92 {\scriptstyle \pm 0.92}$ & $5.27 {\scriptstyle \pm 1.59}$ & $7.63 {\scriptstyle \pm 2.00}$ & $11.80 {\scriptstyle \pm 2.39}$\\
FFG-U (50\%) & $3.11 {\scriptstyle \pm 0.59}$ & $4.26 {\scriptstyle \pm 0.92}$ & $5.44 {\scriptstyle \pm 1.58}$ & $7.58 {\scriptstyle \pm 1.94}$ & $11.38 {\scriptstyle \pm 2.26}$\\
FFG-U (25\%) & $5.27 {\scriptstyle \pm 0.35}$ & $5.33 {\scriptstyle \pm 0.57}$ & $6.20 {\scriptstyle \pm 1.19}$ & $7.95 {\scriptstyle \pm 1.55}$ & $11.19 {\scriptstyle \pm 2.21}$\\
\bottomrule

\end{tabular}
\label{tab:corrupted_c10_ece_pruning}
\end{table*}

\begin{table*}[ht]
\centering
\caption{Corrupted CIFAR-100 accuracy ($\uparrow$) values (in $\%$) for pruning \acrshort{ffg}-W and \acrshort{ffg}-U.}
\newcommand{\corruptedcifarhundredaccpruningtable}{\toprule
& \multicolumn{5}{c}{Skew Intensity}\\
Method & 1 & 2 & 3 & 4 & 5\\
\midrule
FFG-W (100\%) & $57.49 {\scriptstyle \pm 3.17}$ & $47.62 {\scriptstyle \pm 3.64}$ & $41.99 {\scriptstyle \pm 4.15}$ & $35.61 {\scriptstyle \pm 4.24}$ & $26.59 {\scriptstyle \pm 3.66}$\\
FFG-W (50\%) & $57.16 {\scriptstyle \pm 3.16}$ & $47.33 {\scriptstyle \pm 3.65}$ & $41.43 {\scriptstyle \pm 4.18}$ & $35.02 {\scriptstyle \pm 4.25}$ & $25.89 {\scriptstyle \pm 3.58}$\\
FFG-W (10\%) & $58.77 {\scriptstyle \pm 3.18}$ & $48.61 {\scriptstyle \pm 3.69}$ & $42.89 {\scriptstyle \pm 4.25}$ & $36.33 {\scriptstyle \pm 4.35}$ & $26.97 {\scriptstyle \pm 3.70}$\\
FFG-W (1\%) & $50.64 {\scriptstyle \pm 2.89}$ & $41.70 {\scriptstyle \pm 3.35}$ & $37.35 {\scriptstyle \pm 3.69}$ & $31.56 {\scriptstyle \pm 3.65}$ & $23.84 {\scriptstyle \pm 3.05}$\\
FFG-W (0.1\%) & $6.72 {\scriptstyle \pm 0.23}$ & $6.00 {\scriptstyle \pm 0.24}$ & $5.77 {\scriptstyle \pm 0.28}$ & $5.43 {\scriptstyle \pm 0.33}$ & $4.82 {\scriptstyle \pm 0.33}$\\
FFG-U (100\%) & $61.71 {\scriptstyle \pm 3.35}$ & $52.61 {\scriptstyle \pm 3.84}$ & $47.08 {\scriptstyle \pm 4.36}$ & $40.56 {\scriptstyle \pm 4.54}$ & $30.88 {\scriptstyle \pm 3.93}$\\
FFG-U (75\%) & $61.47 {\scriptstyle \pm 3.41}$ & $52.25 {\scriptstyle \pm 3.95}$ & $46.84 {\scriptstyle \pm 4.46}$ & $40.37 {\scriptstyle \pm 4.65}$ & $30.48 {\scriptstyle \pm 3.97}$\\
FFG-U (50\%) & $60.76 {\scriptstyle \pm 3.46}$ & $51.68 {\scriptstyle \pm 3.96}$ & $46.28 {\scriptstyle \pm 4.43}$ & $39.84 {\scriptstyle \pm 4.61}$ & $29.85 {\scriptstyle \pm 3.92}$\\
FFG-U (25\%) & $58.11 {\scriptstyle \pm 3.20}$ & $49.06 {\scriptstyle \pm 3.64}$ & $43.62 {\scriptstyle \pm 4.09}$ & $37.20 {\scriptstyle \pm 4.23}$ & $27.93 {\scriptstyle \pm 3.60}$\\
\bottomrule
}
\begin{tabular}{l|ccccc}
\toprule
& \multicolumn{5}{c}{Skew Intensity}\\
Method & 1 & 2 & 3 & 4 & 5\\
\midrule
FFG-W (100\%) & $57.49 {\scriptstyle \pm 3.17}$ & $47.62 {\scriptstyle \pm 3.64}$ & $41.99 {\scriptstyle \pm 4.15}$ & $35.61 {\scriptstyle \pm 4.24}$ & $26.59 {\scriptstyle \pm 3.66}$\\
FFG-W (50\%) & $57.16 {\scriptstyle \pm 3.16}$ & $47.33 {\scriptstyle \pm 3.65}$ & $41.43 {\scriptstyle \pm 4.18}$ & $35.02 {\scriptstyle \pm 4.25}$ & $25.89 {\scriptstyle \pm 3.58}$\\
FFG-W (10\%) & $58.77 {\scriptstyle \pm 3.18}$ & $48.61 {\scriptstyle \pm 3.69}$ & $42.89 {\scriptstyle \pm 4.25}$ & $36.33 {\scriptstyle \pm 4.35}$ & $26.97 {\scriptstyle \pm 3.70}$\\
FFG-W (1\%) & $50.64 {\scriptstyle \pm 2.89}$ & $41.70 {\scriptstyle \pm 3.35}$ & $37.35 {\scriptstyle \pm 3.69}$ & $31.56 {\scriptstyle \pm 3.65}$ & $23.84 {\scriptstyle \pm 3.05}$\\
FFG-W (0.1\%) & $6.72 {\scriptstyle \pm 0.23}$ & $6.00 {\scriptstyle \pm 0.24}$ & $5.77 {\scriptstyle \pm 0.28}$ & $5.43 {\scriptstyle \pm 0.33}$ & $4.82 {\scriptstyle \pm 0.33}$\\
FFG-U (100\%) & $61.71 {\scriptstyle \pm 3.35}$ & $52.61 {\scriptstyle \pm 3.84}$ & $47.08 {\scriptstyle \pm 4.36}$ & $40.56 {\scriptstyle \pm 4.54}$ & $30.88 {\scriptstyle \pm 3.93}$\\
FFG-U (75\%) & $61.47 {\scriptstyle \pm 3.41}$ & $52.25 {\scriptstyle \pm 3.95}$ & $46.84 {\scriptstyle \pm 4.46}$ & $40.37 {\scriptstyle \pm 4.65}$ & $30.48 {\scriptstyle \pm 3.97}$\\
FFG-U (50\%) & $60.76 {\scriptstyle \pm 3.46}$ & $51.68 {\scriptstyle \pm 3.96}$ & $46.28 {\scriptstyle \pm 4.43}$ & $39.84 {\scriptstyle \pm 4.61}$ & $29.85 {\scriptstyle \pm 3.92}$\\
FFG-U (25\%) & $58.11 {\scriptstyle \pm 3.20}$ & $49.06 {\scriptstyle \pm 3.64}$ & $43.62 {\scriptstyle \pm 4.09}$ & $37.20 {\scriptstyle \pm 4.23}$ & $27.93 {\scriptstyle \pm 3.60}$\\
\bottomrule

\end{tabular}
\label{tab:corrupted_c100_accuracy_pruning}
\end{table*}

\begin{table*}[ht]
\centering
\caption{Corrupted CIFAR-100 \acrshort{ece} ($\downarrow$) values (in $\%$) for pruning \acrshort{ffg}-W and \acrshort{ffg}-U.}
\newcommand{\corruptedcifarhundredecepruningtable}{\toprule
& \multicolumn{5}{c}{Skew Intensity}\\
Method & 1 & 2 & 3 & 4 & 5\\
\midrule
FFG-W (100\%) & $14.28 {\scriptstyle \pm 0.78}$ & $11.07 {\scriptstyle \pm 1.11}$ & $11.13 {\scriptstyle \pm 0.85}$ & $11.21 {\scriptstyle \pm 1.29}$ & $11.96 {\scriptstyle \pm 1.68}$\\
FFG-W (50\%) & $15.42 {\scriptstyle \pm 0.84}$ & $12.04 {\scriptstyle \pm 1.19}$ & $11.85 {\scriptstyle \pm 0.91}$ & $11.77 {\scriptstyle \pm 1.31}$ & $11.71 {\scriptstyle \pm 1.61}$\\
FFG-W (10\%) & $10.85 {\scriptstyle \pm 0.84}$ & $9.08 {\scriptstyle \pm 0.95}$ & $9.96 {\scriptstyle \pm 1.10}$ & $10.97 {\scriptstyle \pm 1.74}$ & $13.35 {\scriptstyle \pm 1.98}$\\
FFG-W (1\%) & $14.86 {\scriptstyle \pm 1.10}$ & $11.99 {\scriptstyle \pm 1.13}$ & $11.78 {\scriptstyle \pm 1.20}$ & $11.48 {\scriptstyle \pm 1.47}$ & $11.37 {\scriptstyle \pm 1.64}$\\
FFG-W (0.1\%) & $3.21 {\scriptstyle \pm 0.57}$ & $4.47 {\scriptstyle \pm 0.55}$ & $5.28 {\scriptstyle \pm 0.70}$ & $6.53 {\scriptstyle \pm 0.96}$ & $7.79 {\scriptstyle \pm 1.01}$\\
FFG-U (100\%) & $4.64 {\scriptstyle \pm 1.66}$ & $7.80 {\scriptstyle \pm 2.03}$ & $10.42 {\scriptstyle \pm 2.39}$ & $13.98 {\scriptstyle \pm 2.83}$ & $19.17 {\scriptstyle \pm 2.72}$\\
FFG-U (75\%) & $4.78 {\scriptstyle \pm 1.58}$ & $7.29 {\scriptstyle \pm 1.91}$ & $9.59 {\scriptstyle \pm 2.33}$ & $13.05 {\scriptstyle \pm 2.91}$ & $18.39 {\scriptstyle \pm 2.87}$\\
FFG-U (50\%) & $5.31 {\scriptstyle \pm 1.55}$ & $7.10 {\scriptstyle \pm 1.81}$ & $9.06 {\scriptstyle \pm 2.18}$ & $12.35 {\scriptstyle \pm 2.82}$ & $17.43 {\scriptstyle \pm 2.83}$\\
FFG-U (25\%) & $9.74 {\scriptstyle \pm 0.99}$ & $8.17 {\scriptstyle \pm 1.01}$ & $8.67 {\scriptstyle \pm 1.21}$ & $10.53 {\scriptstyle \pm 1.85}$ & $13.22 {\scriptstyle \pm 2.15}$\\
\bottomrule}
\begin{tabular}{l|ccccc}
\toprule
& \multicolumn{5}{c}{Skew Intensity}\\
Method & 1 & 2 & 3 & 4 & 5\\
\midrule
FFG-W (100\%) & $14.28 {\scriptstyle \pm 0.78}$ & $11.07 {\scriptstyle \pm 1.11}$ & $11.13 {\scriptstyle \pm 0.85}$ & $11.21 {\scriptstyle \pm 1.29}$ & $11.96 {\scriptstyle \pm 1.68}$\\
FFG-W (50\%) & $15.42 {\scriptstyle \pm 0.84}$ & $12.04 {\scriptstyle \pm 1.19}$ & $11.85 {\scriptstyle \pm 0.91}$ & $11.77 {\scriptstyle \pm 1.31}$ & $11.71 {\scriptstyle \pm 1.61}$\\
FFG-W (10\%) & $10.85 {\scriptstyle \pm 0.84}$ & $9.08 {\scriptstyle \pm 0.95}$ & $9.96 {\scriptstyle \pm 1.10}$ & $10.97 {\scriptstyle \pm 1.74}$ & $13.35 {\scriptstyle \pm 1.98}$\\
FFG-W (1\%) & $14.86 {\scriptstyle \pm 1.10}$ & $11.99 {\scriptstyle \pm 1.13}$ & $11.78 {\scriptstyle \pm 1.20}$ & $11.48 {\scriptstyle \pm 1.47}$ & $11.37 {\scriptstyle \pm 1.64}$\\
FFG-W (0.1\%) & $3.21 {\scriptstyle \pm 0.57}$ & $4.47 {\scriptstyle \pm 0.55}$ & $5.28 {\scriptstyle \pm 0.70}$ & $6.53 {\scriptstyle \pm 0.96}$ & $7.79 {\scriptstyle \pm 1.01}$\\
FFG-U (100\%) & $4.64 {\scriptstyle \pm 1.66}$ & $7.80 {\scriptstyle \pm 2.03}$ & $10.42 {\scriptstyle \pm 2.39}$ & $13.98 {\scriptstyle \pm 2.83}$ & $19.17 {\scriptstyle \pm 2.72}$\\
FFG-U (75\%) & $4.78 {\scriptstyle \pm 1.58}$ & $7.29 {\scriptstyle \pm 1.91}$ & $9.59 {\scriptstyle \pm 2.33}$ & $13.05 {\scriptstyle \pm 2.91}$ & $18.39 {\scriptstyle \pm 2.87}$\\
FFG-U (50\%) & $5.31 {\scriptstyle \pm 1.55}$ & $7.10 {\scriptstyle \pm 1.81}$ & $9.06 {\scriptstyle \pm 2.18}$ & $12.35 {\scriptstyle \pm 2.82}$ & $17.43 {\scriptstyle \pm 2.83}$\\
FFG-U (25\%) & $9.74 {\scriptstyle \pm 0.99}$ & $8.17 {\scriptstyle \pm 1.01}$ & $8.67 {\scriptstyle \pm 1.21}$ & $10.53 {\scriptstyle \pm 1.85}$ & $13.22 {\scriptstyle \pm 2.15}$\\
\bottomrule
\end{tabular}
\label{tab:corrupted_c100_ece_pruning}
\end{table*}

\begin{figure*}[t]
\centering
\includegraphics[width=\linewidth]{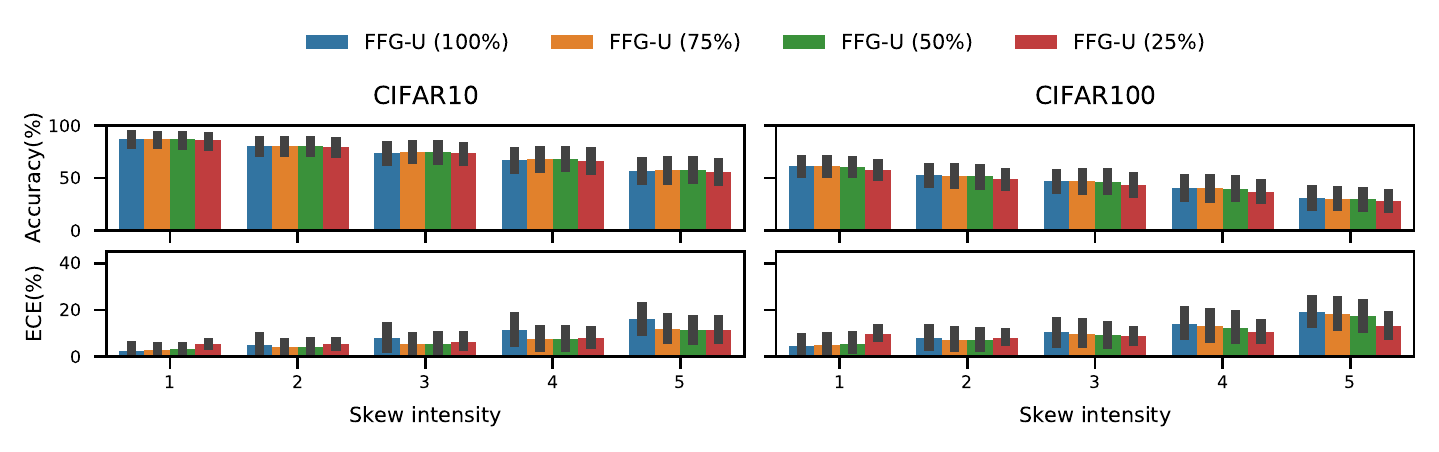}
\caption{Accuracy ($\uparrow$) and \acrshort{ece} ($\downarrow$) on corrupted CIFAR for pruning \acrshort{ffg}-U. We show the mean and two standard errors for each metric on the $19$ perturbations provided in \citep{hendrycks2019benchmarking}.}
\label{fig:cifar_corrupted_pruning_u}
\end{figure*}

\end{document}